\documentclass{article}

\usepackage{microtype}
\usepackage{graphicx}
\usepackage{booktabs}
\usepackage[dvipsnames,table]{xcolor}
\usepackage{hyperref}

\usepackage[accepted]{icml2023}

\usepackage{amsmath,amssymb,mathtools}
\usepackage{bm}
\usepackage{subcaption}
\usepackage{tabularx}
\newcolumntype{Y}{>{\centering\arraybackslash}X}
\newcolumntype{Z}{>{\raggedleft\arraybackslash}X}
\usepackage{nicefrac}
\usepackage{tikz}
\usepackage[inline,shortlabels]{enumitem}
\usepackage{url}
\usepackage[capitalize,noabbrev]{cleveref}

\graphicspath{{./figs/}}
\newcommand{\xhdr}[1]{\textbf{#1.}\;}

\newcommand{\bR}{\ensuremath \mathbb{R}}
\newcommand{\bN}{\ensuremath \mathbb{N}}

\DeclareMathOperator*{\argmax}{arg\,max}

\newcommand{\nbinops}{\ensuremath N_{\mathrm{bin}}}
\newcommand{\nunops}{\ensuremath N_{\mathrm{una}}}
\newcommand{\psymb}{\ensuremath p_{\mathrm{sym}}}
\newcommand{\pconst}{\ensuremath p_{\mathrm{c}}}
\newcommand{\nconsts}{\ensuremath N_{\mathrm{const}}}
\newcommand{\nivs}{\ensuremath N_{\mathrm{iv}}}
\newcommand{\ngrid}{\ensuremath N_{\mathrm{grid}}}
\newcommand{\maxtime}{\ensuremath T}
\newcommand{\ivrange}{\ensuremath (y_0^{\min}, y_0^{\max})}

\newcommand{\nnodes}{\ensuremath K}
\newcommand{\pint}{\ensuremath p_{\mathrm{int}}}
\newcommand{\preal}{\ensuremath p_{\mathrm{real}}}

\icmltitlerunning{\hfill Predicting Ordinary Differential Equations with Transformers\hfill \thepage}

\begin{document}

\twocolumn[
\icmltitle{Predicting Ordinary Differential Equations with Transformers}

\icmlsetsymbol{MPI}{$\dagger$}
\icmlsetsymbol{helmholtzwork}{*}

\begin{icmlauthorlist}
\icmlauthor{Sören Becker}{HelmholtzAI}
\icmlauthor{Michal Klein}{Apple,helmholtzwork}
\icmlauthor{Alexander Neitz}{DeepMind,MPI}
\icmlauthor{Giambattista Parascandolo}{OpenAI,MPI}
\icmlauthor{Niki Kilbertus}{HelmholtzAI,TUM}

\end{icmlauthorlist}

\icmlaffiliation{HelmholtzAI}{Helmholtz AI, Helmholtz Center Munich, Munich, Germany.}
\icmlaffiliation{Apple}{Apple, Paris, France.}
\icmlaffiliation{DeepMind}{DeepMind, London, United Kingdom.}
\icmlaffiliation{OpenAI}{OpenAI, San Francisco, United States.}
\icmlaffiliation{TUM}{Technical University of Munich, Germany}

\icmlcorrespondingauthor{Sören Becker}{soeren.becker@helmholtz-munich.de}
\icmlcorrespondingauthor{Niki Kilbertus}{niki.kilbertus@helmholtz-munich.de}

\vskip 0.3in
]

\printAffiliationsAndNotice{\helmholtzwork{}\MPI{}}

\begin{abstract}
We develop a transformer-based sequence-to-sequence model that recovers scalar ordinary differential equations (ODEs) in symbolic form from irregularly sampled and noisy observations of a single solution trajectory. We demonstrate in extensive empirical evaluations that our model performs better or on par with existing methods in terms of accurate recovery across various settings. Moreover, our method is efficiently scalable: after one-time pretraining on a large set of ODEs, we can infer the governing law of a new observed solution in a few forward passes of the model.
\end{abstract}

\section{Introduction}
\label{sec:introduction}
Researchers in the natural sciences increasingly turn to machine learning (ML) to aid the discovery of natural laws from observational data alone, which is often abundantly available, hoping to bypass expensive and cumbersome targeted experimentation.
While there may be fundamental limitations to what can be extracted from observations alone, recent successes of ML provide ample reason for excitement.
Partially fueled by these promises, interest in symbolic regression (SR) has received renewed attention \citep{la2021contemporary,makke2022interpretable}.
A symbolic representation of a law has several advantages over black-box representations in that they are typically parsimonious and directly interpretable as well as amenable to analytic analysis.

Numerous symbolic regression methods have been proposed recently to infer functional relationships, i.e., to infer a function $f$ symbolically given (noisy) examples $(x_i , f(x_i))_{i =1}^n$ \citep{la2021contemporary}.
Arguably, a more interesting---but also more challenging---task is to infer dynamical laws.
We represent a dynamical law as an ODE $\dot{y} := \nicefrac{dy}{dt} = f(y)$, which is fully determined by $f$.
In this setting, the goal is to infer $f$ from (noisy, irregular) samples $(t_i, y_i)_{i=1}^n$, where $y$ is a solution of the ODE (and $y_i$ denotes the observed value for $y(t_i)$).
Compared to symbolic regression for functional relationships, relatively little work exists on directly inferring dynamical laws (see \cref{sec:related_work} for details).
In principle, any functional symbolic regression method may be applied to $(y_i, \hat{\dot{y}}_i)_{i=1}^n$, where $\hat{\dot{y}}_i$ are estimated derivatives at the observed time points $t_i$, to obtain $f$.
This naturally raises the question whether methods tailored to directly inferring dynamical laws yields better results.

In this work, we develop \textbf{N}eural \textbf{S}ymbolic \textbf{O}rdinary \textbf{D}ifferential \textbf{E}quation (\textbf{NSODE}), a transformer based sequence-to-sequence model specifically tailored to inferring dynamics directly in an end-to-end fashion from $(t_i, y_i)$ samples of a single solution trajectory.
NSODE leverages large scale pre-training for efficient inference at test time.
We first (randomly) generate a total of $>$3M scalar, autonomous, non-linear, first-order ODEs, together with a total of $>$63M numerical solutions from various (random) initial conditions.
All solutions are carefully checked for convergence of the numerical integration.
Code and data are publicly available at \url{https://github.com/soerenab/nsode23}.

NSODE, an encoder-decoder transformer, is then trained in a supervised fashion to map observed trajectories, i.e., numeric sequences of the form $(t_i, y_i)_{i=1}^n$, directly to symbolic equations as strings, e.g., \texttt{"y**2+1.64*cos(y)"}, which is the prediction for $f$.
This example directly highlights the benefit of symbolic representations in that the $y^2$ and $\cos(y)$ terms tell us something about the fundamental dynamics of the observed system; the constant \texttt{1.64} will have semantic meaning in a given context.
NSODE combines and innovates on technical advances regarding input representations and an efficiently optimizable loss formulation.
Our model outperforms most existing methods and is more efficient at inference time than models with competitive performance.

\begin{figure*}
  \centering
  \includegraphics[width=1.0\textwidth]{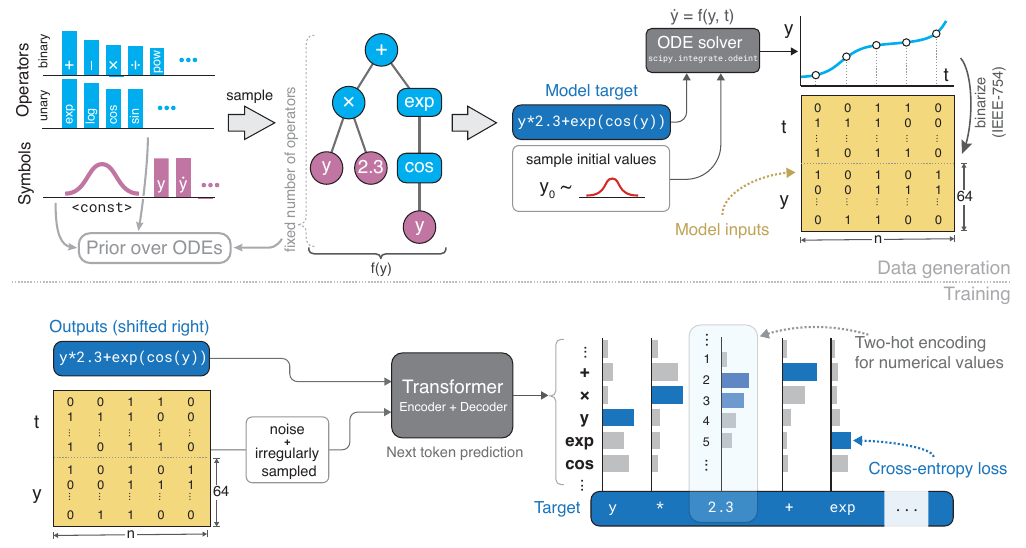}
  \caption{An overview illustration of the data generation (top) and training pipeline (bottom). In the lower right, blue bars correspond to one-hot and two-hot target encodings whereas gray bars correspond to model predictions. Our dataset stores solutions in numerical (non-binarized) form on the entire regular solution time grid.}
  \label{fig:overview}
\end{figure*}

\section{Background and Related Work}
\label{sec:related_work}
Modeling dynamics and forecasting their behavior has a long history in machine learning.
While Neural Ordinary Differential Equations (NODE) \citep{chen2018neural} (with a large body of follow up work) are perhaps the most prominent modern approach from the recent years, their inherent black-box character complicates interpretability or scientific understanding of the observed phenomena.
Recent alternatives such as tractable dendritic RNNs \citep{brenner2022tractable} and Neural Operators \citep{kovachki2021neural, LearningDissipativeDynamics} set out to facilitate scientific discovery by combining accurate predictions with improved interpretability.
A considerable advantage of these and similar approaches is their scalability to high dimensional systems as well as their relative robustness to noise, missing or irregularly sampled data \citep{iakovlev2021learning} and challenging properties such as multiscale dynamics \citep{vlachas2022multiscale} or chaos \citep{park2022recurrent, patel2022using}. 
The recent benchmark study by \citet{gilpin2021} provides an excellent overview of dynamics forecasting models including deep learning-based blackbox approaches as well as symbolic models.

However, in this work we focus on models that explicitly predict a mathematical expression in symbolic form because of the advantages of their interpretable and parsimonious representations.
We now provide an overview of the most prominent types of symbolic regression techniques.
Our overview also includes methods that have been primarily designed to infer functional relationships only, i.e., that are not specifically tailored to inferring dynamical laws.

\xhdr{Evolutionary algorithms}
Traditionally, many approaches to symbolic regression broadly fall into the category of evolutionary algorithms such as genetic programming (GP) \citep{koza}.
Genetic programming randomly evolves a population of prospective mathematical expressions over multiple iterations and mimics natural selection by keeping only the best contenders across iterations, where superiority is measured by user-defined fitness functions \citep{schmidt2009distilling}.
Process-based modeling follows a similar approach but includes domain knowledge-informed constraints on particular components of the system in order to reduce the search space to reasonable candidates \citep{todorovski1997declarative,bridewell2008inductive,simidjievski2020equation}.
A large body of work examines and improves upon all aspects of GP to overcome typical shortcomings such as premature convergence, overly complex output expressions, scalability (both in terms of time as well as memory), or the difficulty of incorporating prior knowledge \citep{schmidt2010age,arnaldo2014multiple,la2016epsilon,virgolin2017scalable,la2018learning,virgolin2019linear,burlacu2020operon,de2021interaction,tohme2022gsr,mundhenk2021symbolic}.

\xhdr{Regression (gradient-based methods)}
More recently, symbolic regression has been approached via machine learning methods which exploit gradient information to optimize within the space of (finite) compositions of pre-defined basis functions.
\citet{sindy} builds on sparse linear regression to identify a linear combination of basis functions.
This approach has inspired a large body of follow-up work generalizing the idea to partial observations \citep{bakarji2022discovering}, parameterized functions \citep{lejarza2022data}, simultaneous discovery of coordinates \citep{champion2019data}, coordinate transformations that linearize the dynamics \citep{lusch2018deep}, and partial differential equations \citep{rudy2017data} among others.
Similarly, \citet{mcconaghy2011ffx} use path-wise regularized learning with ElasticNets on a large body of pre-generated non-linear basis functions for symbolic regression.
These techniques often deploy sparsity-promoting regularizers and train one or multiple models for each set of observations.
Once trained, the model itself represents the predicted symbolic expression, which can be read off the non-zero coefficients. 
This modeling principle is also employed by many other approaches that replace linear regression by neural networks with diverse sets of activation functions, both for differential equations \citep{pdenet2,rodenet} and non-differential algebraic equations \citep{eql2}.

\xhdr{Hybrid models}
Supervised learning with gradient-based optimization to directly output the symbolic expression (e.g., as a string) is challenged by the formulation of a differentiable loss between the predicted symbolic expression (string) and the observed data (numerical).
Thus, prior work on functional symbolic regression resorted to reinforcement learning \citep{deepsymres,landajuela2021discovering} or combinations of neural networks and evolutionary algorithms \citep{atkinson2019data, costa2021fast}.
A hybrid approach combining gradient-free, human intuition-guided heuristic search via genetic programming with neural network-based optimization has been presented for non-differential equations by \citet{aifeynman2}.
This method proceeds by a divide-and-conquer strategy in applying the hand-tuned heuristics.
It has recently been extended to dynamical systems by \citet{weilbach2021inferring}.

\xhdr{Monte Carlo methods}
Finally, \citet{jin2019bayesian,brence2021probabilistic} and extensions such as \citet{gec2022discovery} incorporate prior knowledge by flexibly specifying distributions over the allowed function space.
This also allows them to perform Bayesian updates and ultimately equation discovery via Monte Carlo sampling from the (posterior) distribution.

\xhdr{Sequence-to-sequence models}
The closest works to ours use pre-trained, attention-based sequence-to-sequence models for symbolic regression \emph{of functional relationships} \citep{nesymres, SymbolicGPT2021,kamienny2022end,vastl2022symformer} or (discrete) recurrence relations \citep{pmlr-v162-d-ascoli22a}.
They exploit the fact that symbolic expressions for (multi-variate) scalar functions can be both generated and evaluated on random inputs cheaply, resulting in essentially unlimited labeled training data that allows for gradient-based optimization using the cross-entropy loss on the symbol level (instead of numerical proximity between evaluations of the true and predicted functions).
Our model differs in a number of key innovations described in \cref{sec:model} overcoming limitations of existing methods and rendering it suitable for inferring dynamical laws directly.

\section{Method}
\label{sec:method}
Many previous symbolic regression methods have been described as ``discovering natural laws''.
However, most of them learn \emph{fixed functional relationships} from input-output pairs, whereas we seek to actually infer \emph{the underlying dynamic law} that governs the behavior of the observed solution trajectory directly.
One way to still apply functional SR to dynamics is to approximate derivatives $\hat{\dot{y}}_i$ from the observed data $(y_i)_{i=1}^n$ and use $(y_i, \hat{\dot{y}})_{i=1}^n$ to infer $f$ (as in $\dot{y} = f(y)$).
Since temporal derivatives are usually not measured directly, this approach crucially depends on the derivative estimation, typically via finite difference approximations.
Even though higher-order finite difference methods can be extended to irregularly sampled and noisy observations, naturally the question arises whether methods specifically tailored to inferring dynamics directly may be superior.
\citet{dcode} have also recently identified alternative loss formulations that bypass unobserved time derivatives as an open challenge in symbolic regression for dynamical systems.
With NSODE, we propose a solution to this challenge.

\xhdr{Problem setting}
Given noisy observations $\{(t_i, y_i)\}_{i=1}^n$ of a trajectory $y: [t_1, t_n] \to \bR$ that is a solution of the scalar ODE $\dot{y} = f(y)$, we aim to recover the function~$f$ in symbolic form.
In this formulation, we explicitly assume that the observed system actually evolves according to an ODE in canonical form~$\dot{y} = f(y)$ such that~$f$ can be expressed in closed form using the mathematical operators seen during training (see \cref{sec:data} for details).

While solutions to the class of ODEs considered in this work are known to have relatively simple limiting behaviors (essentially either blowing up or reaching a constant equilibrium), within finite time they still exhibit rich and varied behavior.
We show some examples in \cref{app:example_trajectories}.

\subsection{Data Generation}
\label{sec:data}

\xhdr{Sampling symbolic expressions}
\label{xhdr:sampling}
To exploit large-scale supervised pretraining, we randomly generate a dataset of $\sim$63M ODEs in symbolic form along with their numerical solutions for multiple randomly sampled initial values.
Since we assume ODEs to be in canonical form $\dot{y} = f(y)$, generating an ODE is equivalent to generating a symbolic expression $f(y)$.
We follow \citet{lample2019deep}, who sample such an expression $f(y)$ as a unary-binary tree, where each internal node corresponds to an operator and each leaf node corresponds to a constant or variable.
The algorithm consists of two phases: (1) A unary-binary tree is sampled uniformly from the distribution of unary-binary trees with up to $k\in \bN$ internal nodes, which crucially does not overrepresent small trees corresponding to short expressions. Here the maximum number of internal nodes $\nnodes$ is a hyperparameter of the algorithm.
(2) The sampled tree is ``decorated'', that is, each binary internal node is assigned a binary operator, each unary internal node is assigned a unary operator, and each leaf is assigned a variable or constant. 
Hence, we have to specify a distribution over the $\nbinops$ binary operators, a distribution over the $\nunops$ unary operators, a probability $\psymb \in (0,1)$ to decide between symbols and constants for leaf nodes, as well as a distribution $\pconst$ over constants.
For constants, in NSODE we further distinguish explicitly between sampling an integer or non-integer value.
Together with $\nnodes$, these choices uniquely determine a distribution over equations $f$ and are described in detail in \cref{app:modeltraining}.
The top part of \cref{fig:overview} depicts an overview of the data generation procedure.

The pre-order traversal of a sampled tree results in the symbolic expression for~$f$ in prefix notation.
After conversion to the more common mathematical infix notation, we simplify each expression using the computer algebra system SymPy \citep{sympy}, and filter out constant equations~$f(y) = c$ as well as expressions that contain operators or symbols that were not in the support of the original distribution.\footnote{With the exception of a unary $-$, which we do not discard.}
We call the structure modulo the value of the constants of such an expression (i.e., replacing all actual constant values by a generic \texttt{<const>} token) a \textbf{skeleton}.

Many skeletons can be represented by different unary-binary trees and
hence many of the generated trees will be simplified to the same skeleton.
To ensure diversity and to mitigate potential dataset bias towards particular expressions, we discard duplicates on the skeleton level.
To further cheaply increase the variability of ODEs we sample $\nconsts$ unique sets of constants per skeleton.
When sampling constants we take care not to modify the canonical expression by adhering to the rules listed in \cref{app:constant_rules}.
Our final dataset contains linear and non-linear as well as homogeneous and inhomogeneous ODEs and we provide summary statistics about the distribution over equations in \cref{app:datastats}.
Besides the number of internal nodes, a simple yet common measure of \textbf{complexity} for each symbolic equation is the overall count of symbols (e.g., $y$, or constants) as well as operators.
We follow previous works on symbolic regression in using this complexity measure in our empirical evaluation and refer to \citet{la2021contemporary} for an overview of proposed alternatives.

\xhdr{Computing numerical solutions}
We obtain numerical solutions for all generated initial value problems via SciPy's interface \citep{scipy} to the LSODA software package \citep{lsoda} with both relative and absolute tolerances set to $10^{-9}$.
LSODA consists of a collection of ODE solvers and implements a strategy to automatically choose an appropriate solver for the problem at hand (e.g., recognizing stiff problems).
We solve each equation on a fixed time interval $t \in [0, \maxtime]$ and store solutions on a regular grid of $\ngrid$ points.
For each ODE, we sample up to $\nivs$ initial values $y(0) = y_0$ uniformly from $\ivrange$.\footnote{Due to a timeout per ODE, fewer solutions may remain if the solver fails for repeated initial value samples.}
While LSODA attempts to select an appropriate solver, numerical solutions still cannot be trusted in all cases.
Therefore, we check the validity of solutions via the following quality control check: we use 9th order central finite differences to approximate the temporal derivative of the solution trajectory (on the same equidistant temporal grid as the proposed solution and without adding noise), denoted by $\dot{y}_{\mathrm{fd}}$, and filter out any solution for which $\|\dot{y}_{\mathrm{fd}} - \dot{y} \|_{\infty} > \epsilon$, where we use $\epsilon = 1$.

\begin{table}
\rowcolors{2}{white}{gray!15}
\centering
\caption{Overview of our model architecture.}\label{tab:architecture}
\begin{tabularx}{\columnwidth}{l@{\hspace{6pt}}Y@{\hspace{0pt}}Y}
\toprule \rowcolor{white}
    & \textbf{Encoder} & \textbf{Decoder} \\ \midrule
    \textbf{layers} & 6 & 6 \\
    \textbf{heads} & 16 & 16  \\
    \textbf{embed. dim.} & 512 & 512 \\
    \textbf{forward dim.} & 2048 & 2048 \\
    \textbf{activation} & gelu & gelu  \\
    \textbf{vocab. size} & - &  43 \\
    \textbf{position enc.} & learned & learned \\
    \textbf{parameters} & 23.3M & 23.3M \\ \bottomrule
\end{tabularx}
\end{table}

\subsection{Model Design Choices and Training}
\label{sec:model}

NSODE consists of an encoder-decoder transformer with architecture choices listed in \cref{tab:architecture}.
A visual overview of the training is depicted in \cref{fig:overview}.

\xhdr{Representing input trajectories}
A key difficulty in feeding numerical observations $(y_i)_{i=1}^n$ as input sequence to a transformer is that their range may differ greatly both within a single solution as well as across ODEs.
For example, the linear ODE $\dot{y} = c \cdot y$ for a constant~$c$ is solved by an exponential $y(t) = y_0 \exp(c t)$ for initial value $y(0) = y_0$, which may span many orders of magnitude on a fixed time interval.
To prevent numerical errors and vanishing or exploding gradients caused by the large range of values, we assume each representable 64-bit float value is a token and use its IEEE-754 encoding as the token representation \citep{nesymres}.
We thus convert all pairs $(t_i, y_i)$ to their IEEE-754 64 bit representations, channel them through a linear layer, and then feed them to the encoder.
The linearly transformed bit pattern hence replaces the explicit embedding layer that commonly preceeds the encoder.

\xhdr{Representing symbolic expressions}
The target sequence (i.e., the string for the symbolic expression of~$f$) is tokenized on the (mathematical) symbol-level.
For all operators and variables we include separate unique tokens in the vocabulary. These tokens are one-hot encoded and passed through a learnable embedding layer before their embedded representations are fed to the decoder.

Constants (as in fixed numerical values) play a special role in sequence-to-sequence approaches to symbolic regression.
While the cross-entropy loss works well for discrete, one-hot encoded operators and symbols (e.g. \texttt{+,exp,sin,x,y}), one cannot directly add all possible constant values such as \texttt{1.452} to the vocabulary as separate tokens.
Naively tokenizing on the digit level, i.e., representing real values literally as the sequence of characters (e.g., \texttt{"1,.,4,5,2"}), not only significantly expands the length of target sequences and thus the computational cost, but also requires a variable number of prediction steps for every single constant.
As a workaround previous works on \emph{functional SR} resort to one of two strategies: 
(1) represent all constants with a special \texttt{<const>} token and optimize their actual values in a separate fine-tuning step.
(2) round constants to a finite number of possible values, which can then all be represented as individual tokens.

The second optimization of strategy (1) comes at a substantial computational cost as constants have to be fit per inferred expression.
For efficiency and scalability, we would like the sequence-to-sequence model propose a complete equation, including values for the involved constants.
Even more detrimental to our problem setting, this approach does not transfer to inferring ODEs: to optimize constants via a regression loss, one would first have to solve the predicted ODE $\dot{y} = \hat{f}(y)$ to obtain predicted $\{\hat{y}_i\}_{i=1}^n$ values that can be compared to the set of observations $\{y_i\}_{i=1}^n$.
That is, the objective function to be optimized when fine-tuning constants involves solving an ODE.
While differentiable ODE solvers exist, optimizing constants per inferred expression this way is prohibitively expensive and typically highly unstable.
Even though strategy (2) avoids a separate optimization step and can leverage clever encoding schemes with improved token efficiency, it comes with an inherent loss of precision.

Therefore, we propose the following representation of constant values.
Taking inspiration from \citet{schrittwieser2020mastering}, we encode constants in a \emph{two-hot} fashion.
We fix a finite homogeneous grid on the real numbers $x_1 < x_2 < \ldots < x_m$ for some~$m \in \bN$ and add those values as tokens to the vocabulary.
The range of integers, the grid range $(x_1, x_m)$, and number of grid points $m$ are hyperparameters that can be tuned for performance.
Our choices are described in \cref{app:modeldesign}.
For any constant $c$ in the target sequence we then find $i \in \{1, \ldots, m-1\}$ such that $x_i \le c < x_{i+1}$ and encode~$c$ as a distribution supported on $x_i, x_{i+1}$ with weights $\alpha, \beta$ such that $\alpha x_i + \beta x_{i+1} = c$.
That is, the target in the cross-entropy loss for a constant token is not a strict one-hot encoding, but a distribution supported on two (neighboring) vocabulary tokens resulting in a lossless encoding of continuous values in $[x_1, x_m]$ which does not require rounding.
While this two-hot representation can be used directly in the cross-entropy loss function and thus greatly facilitates training, it can not be passed directly through an embedding layer.
For a generic constant in the target sequence represented as $\alpha x_i + \beta x_{i+1}$, we thus instead provide the linear combination of the two embeddings $\alpha\; \text{\texttt{embed(x\textsubscript{i})}} + \beta\; \text{\texttt{embed(x\textsubscript{i+1})}}$ as decoder input.

\xhdr{Decoding constants}
When decoding a predicted sequence, we check at each prediction step whether the $\argmax$ of the logits corresponds to one of the~$m$ constant tokens $\{x_1, \ldots, x_m\}$.
If not, we proceed by conventional one-hot decoding to obtain predicted operators and variables.
If instead the argmax corresponds to, for example, $x_i$, we also pick its largest-logit neighbor ($x_{i-1}$ or $x_{i+1}$; suppose $x_{i+1}$), renormalize their probabilities by applying a softmax to all logits and use the resulting two probability estimates as weights $\alpha, \beta$.
Constants are then ultimately decoded as $\alpha x_i + \beta x_{i+1}$.
We depict our decoding scheme in \cref{fig:overview}.

\xhdr{Sampling solutions}
To infer a symbolic expression for the governing ODE of a new observed solution trajectory $\{(t_i, y_i)\}_{i=1}^n$, all the typical policies such as greedy, sampling, or beam search are available.
In our evaluation, we leverage computationally cheap forward passes to perform a beam search with 1536 beams.
We provide details about how NSODE is evaluated in \cref{sec:metrics}.

\label{sec:experiments}
\begin{table*}
\rowcolors{2}{white}{gray!15}
\centering
\caption{Overview of baselines (f.d.: finite differences, ode: proposed for ODEs, MC: Monte Carlo, reg.: regression).}\label{tab:baselines}
{\small
\begin{tabularx}{\linewidth}{ccccYc}
  \toprule \rowcolor{white}
    \textbf{name} & \textbf{type} & \textbf{ode} & \textbf{f.d.} & \textbf{description} & \textbf{reference} \\ \midrule
    AFP        & GP & no & yes & age-fitness Pareto optimization & \citep{schmidt2010age} \\
    FE-AFP     & GP & no & yes & AFP with co-evolved fitness estimates & \citep{schmidt2010age} \\
    EHC        & GP & no & yes & AFP with epigenetic hillclimbing  & \citep{la2016thesis} \\
    EPLEX      & GP & no & yes & epsilon-lexicase selection & \citep{la2016epsilon} \\
    GPGOMEA    & GP & no & yes & gene-pool optimal mixing & \citep{virgolin2017scalable} \\
    FEAT       & GP & no & yes & learned differentiable features & \citep{la2018learning} \\
    PySR       & GP & no & yes & AutoML-Zero + simulated annealing + const.\ optim. & \citep{pysr} \\
    SINDy      & reg    & yes & yes & sparse linear regression & \citep{sindy} \\
    FFX        & reg    & no & yes & pathwise regularized ElasticNet regression & \citep{mcconaghy2011ffx} \\
    BSR        & MC   & no & yes & MCMC on linearly mixed tree-representations  & \citep{jin2019bayesian} \\
    ProGED     & MC   & yes & no & MC on probabilistic context free grammars+const.\ optim. & \citep{brence2021probabilistic} \\
    \bottomrule
\end{tabularx}}%
\end{table*}

\xhdr{Training}
We train two versions of our model. \textbf{NSODE} is trained on $n=256$ time-points per trajectory, which are sampled on an equidistant grid over the training interval $[0, T]$.
To increase the robustness of the model with respect to noisy, potentially incomplete observations, we train a second model, \textbf{NSODE-eps}, for which we add multiplicative Gaussian noise centered on 1 with a standard deviation of $\sigma = 0.01$ to the observed input trajectory.
Furthermore we do not feed the full solution trajectory generated as described in \cref{sec:data} but keep only $n=128$ time-points which are selected uniformly at random from the interval $[0, T]$.
All details about model training such as hyperparameter choices and the used hardware are in \cref{app:modeldesign}.
\section{Experiments}
\subsection{Benchmark Datasets}\label{sec:datasets}
We evaluate model performance and generalization on several test sets, which are summarized in \cref{fig:datastats}.
\begin{itemize}[leftmargin=*,topsep=0pt,itemsep=0pt]
    \item \textbf{Classic}:
    To validate our approach on existing benchmarks, we turn to the functional symbolic regression literature and simply interpret functions as ODEs.
    In particular, we include all scalar functions listed in the overview in \cite{mcdermott2012genetic}, which includes equations from multiple established benchmarks \cite{keijzer2003improving, koza, koza1994genetic, uy2011semantically, vladislavleva2008order}.
    For example, we interpret the function $f(y) = y^3 + y^2 + y$ from \citet{uy2011semantically} as an autonomous ODE $\dot{y}(t) = f(y(t)) = y(t)^3 + y(t)^2 + y(t)$, which we solve numerically for randomly sampled initial values (as detailed in \cref{sec:data}).
    This test set consists of 26 distinct equations.
    \item \textbf{Textbook}: To assess how NSODE performs on ``real problems'', we manually curated 12 non-linear ODEs from Wikipedia, physics textbooks, and lecture notes from university courses on ODEs.
    These equations are listed in \cref{tab:textbook_equations} in \cref{app:textbook_equations}.
    We note that they are all relatively simple compared to the expressions in our generated training set, consisting mostly of low order polynomials, some of which with one fractional exponent.
    \item \textbf{Large}: The \textbf{Classic} and \textbf{Textbook} datasets are relatively small and simple in terms of the complexity and operator diversity of the expressions (cf.\ \cref{fig:datastats}).
    Hence, we generate a larger and more diverse dataset by resampling equations from the training distribution described in \cref{sec:data} and solving them for new initial conditions to generate new, unseen trajectory.
    To further reduce bias towards our training distribution, we employ rejection sampling to ensure that no skeleton is included more than once and that we include at most 10 equations per complexity.
    The final dataset consists of 162 ODEs, which is comparable in size to the datasets used in the recent extensive functional SR benchmark study by \citet{la2021contemporary}.
    We refrained from including even more equations, because most SR methods require a separate optimization per expression, quickly rendering the evaluation of baselines computationally infeasible.
\end{itemize}

\subsection{Baselines}\label{sec:baselines}
We compare our method to 11 popular baselines choosing strong contenders from different categories described in \cref{sec:related_work}.
We provide a brief overview in \cref{tab:baselines} and defer details on hyperparameter choices to \cref{app:baselines}.
All baselines explicitly fit a separate regression function for each individual observed trajectory.
Moreover, except for ProGED, which has a specific mode for ODE discovery, all models use functional SR and require finite difference approximations for the derivatives as inputs.
We use smoothed finite differences as provided by the PySindy \citep{pysindy} implementation \texttt{SmoothedFiniteDifference} with a smoothing window length of 15. Notebly, this implementation also provides methods to approximate temporal derivatives under unevenly spaced sampling intervals.
Beyond the baselines in \cref{tab:baselines}, we attempted to compare to AI Feynman \citep{aifeynman2, weilbach2021inferring}, deep symbolic regression \citep{deepsymres}, multiple regression genetic programming \citep{arnaldo2014multiple}, and semantic backpropagation-based genetic programming \citep{virgolin2019linear}.
However, due to their large inference time per equation we could not obtain sufficient results for a reasonable comparison to other models.
The relatively long inference times of these methods are confirmed in \citet[Fig.\ 1]{la2021contemporary}, where they all average at above 2.5 hours per expression.

\subsection{Metrics and Evaluation}
\label{sec:metrics}

\begin{figure*}[t!]
\centering
\begin{subfigure}{.33\textwidth}
    \centering
    \includegraphics[width=1\linewidth]
    {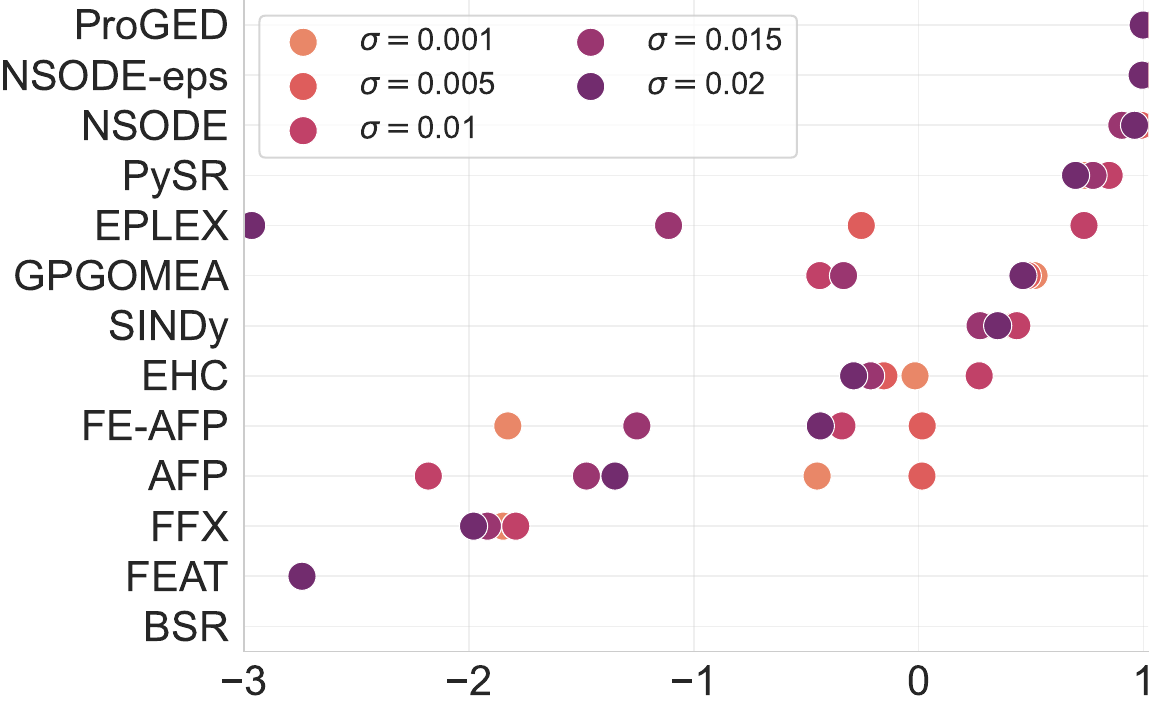}
    \caption{R$^2$}
    \label{fig:results_r2_complexity_time:r2}
\end{subfigure}%
\begin{subfigure}{.33\textwidth}
    \centering
    \includegraphics[width=1\linewidth]{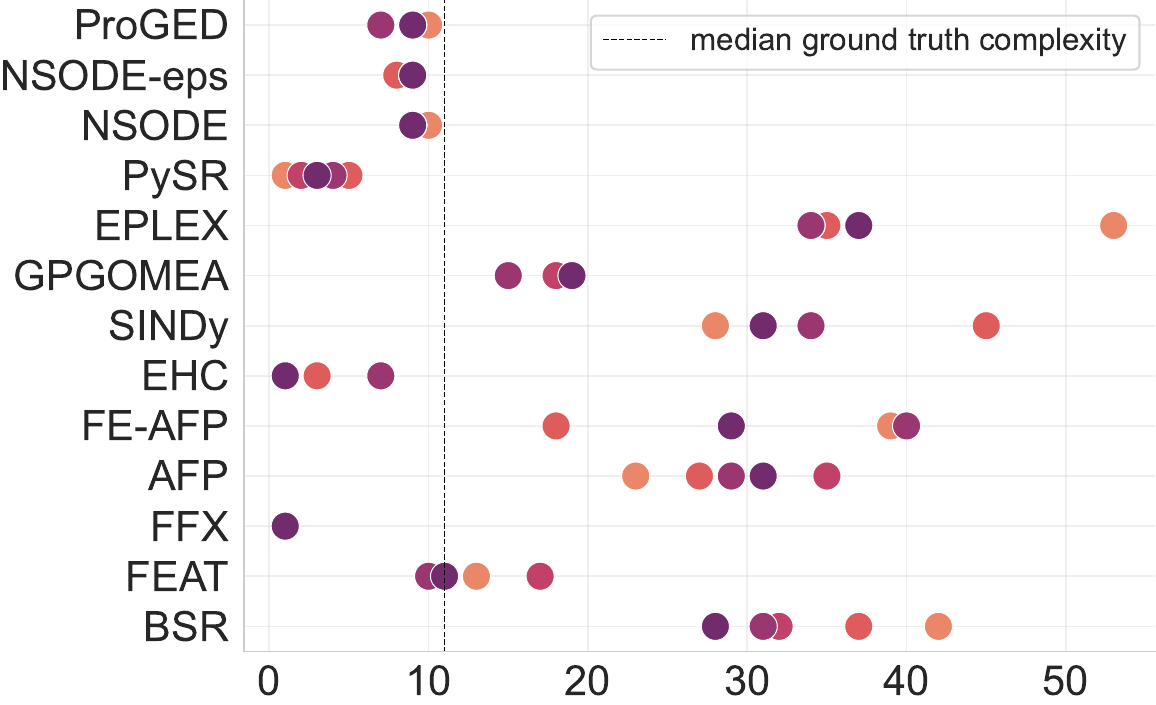}
    \caption{complexity}
\end{subfigure}%
\begin{subfigure}{.33\textwidth}
    \centering
    \includegraphics[width=1\linewidth]{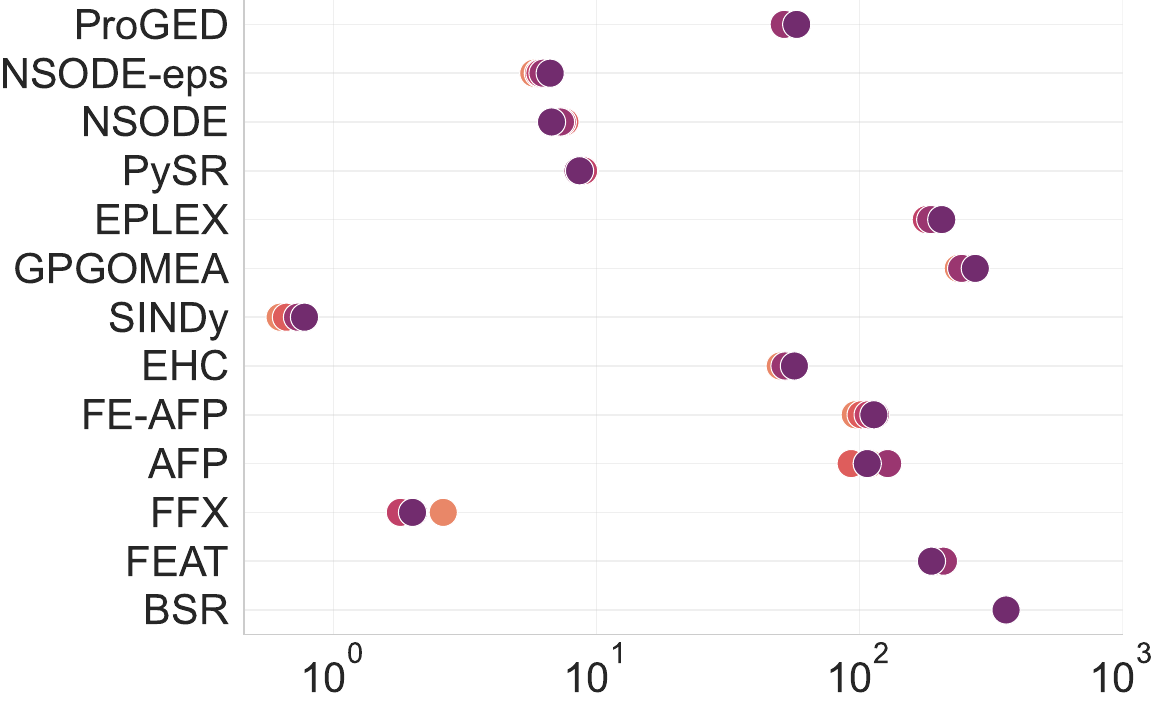}
    \caption{inference time (sec.)}
\end{subfigure}%
\caption{\textbf{Performance metrics on Classic.} Median R$^2$, complexity, and inference times on \textbf{Classic} with 192 irregularly spaced points and different noise levels $\sigma$ in the interpolation regime $[0, T]$.
Rows in all plots are ordered according to best R$^2$ scores.
In (a) the x-axis is restricted to the interval $[-3, 1]$; missing performances (e.g., for BSR) fall below this threshold. The black dashed line in (b) denotes the median complexity across all samples in the testset.}
\label{fig:results_r2_complexity_time}
\end{figure*}

\begin{figure*}[t!]
\centering
\begin{subfigure}{.33\textwidth}
    \centering
    \includegraphics[width=1\linewidth]{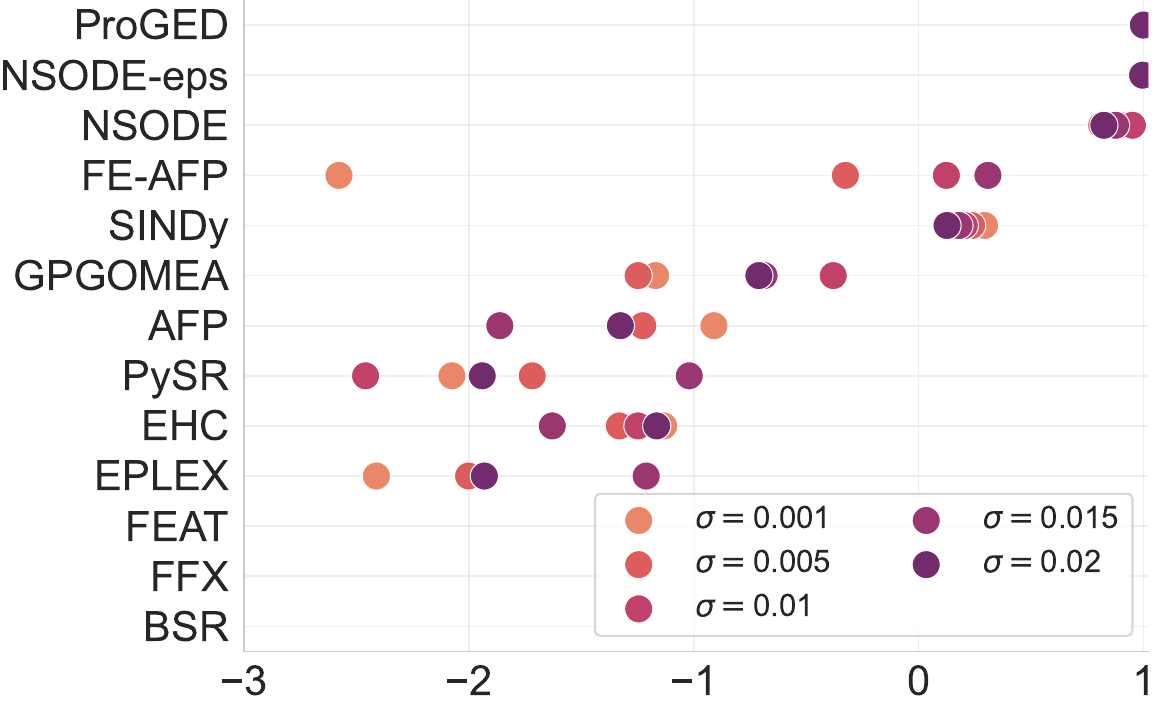}
    \caption{R$^2$ on \textbf{Classic}}
\end{subfigure}%
\begin{subfigure}{.33\textwidth}
    \centering
    \includegraphics[width=1\linewidth]{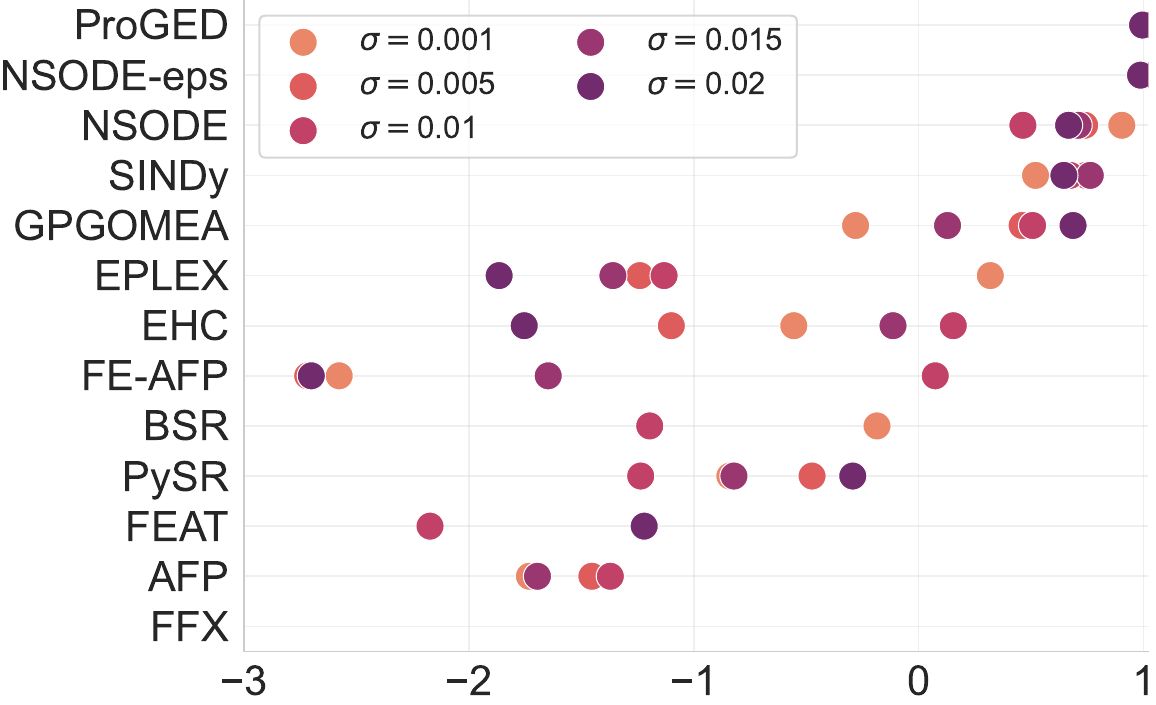}
    \caption{R$^2$ on \textbf{Textbook}}
\end{subfigure}%
\begin{subfigure}{.33\textwidth}
    \centering
    \includegraphics[width=1\linewidth]
    {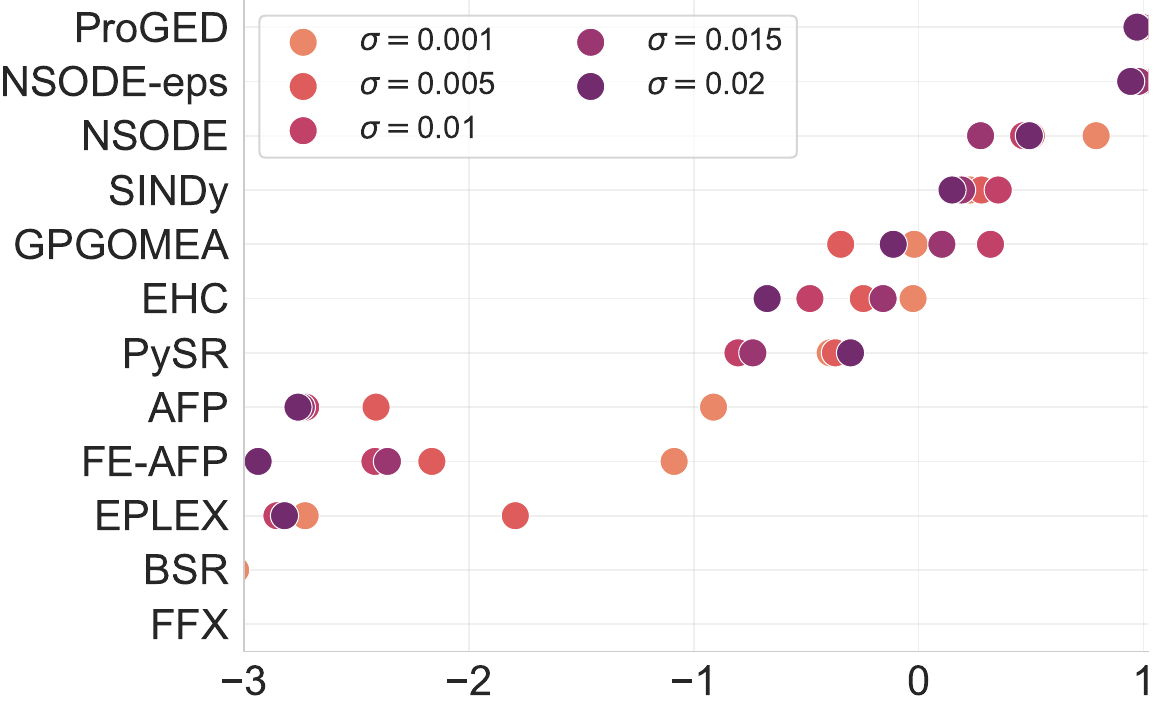}
    \caption{R$^2$ on \textbf{Large}}
\end{subfigure}%
\caption{\textbf{Performance across testsets.} Median R$^2$ scores across predictions for different models using 128 irregularly spaced time points for different noise levels $\sigma$ in the interpolation regime $[0, T]$. The x-axes are restricted to the interval $[-3, 1]$; missing performances fall below this threshold.}
\label{fig:results_r2_datasets}
\end{figure*}

\xhdr{Metrics}
For performance evaluations we follow standard procedures within the symbolic regression literature (see, e.g., \cite{la2021contemporary}) and assess accuracy, expression parsimony and inference time per equation. To compute accuracy metrics we first integrate the predicted ODE expression over the interval $[0, T]$ using the same initial value as in the ground truth trajectory. The integrated predicted trajectory is then compared to the ground truth trajectory in terms of the well-established coefficient of determination (R$^2$), the $L_1$ and $L_{\infty}$ norm of their difference as well as the average \texttt{numpy.isclose} as a possible alternative as suggested by \cite{nesymres}\footnote{For two arrays \texttt{a} and \texttt{b} we define the average isclose as \texttt{numpy.isclose(a, b).sum()/len(b)} with \texttt{atol=1e-10} and \texttt{rtol=0.05}; \texttt{a} corresponds to predictions, \texttt{b} corresponds to ground truth.}.
Expression parsimony is measured in terms of equation complexity as introduced in \cref{sec:data}.
In short, the complexity is the total number of operators, variables, and constants in an expression.

\xhdr{Model selection}
Many of the baseline methods as well as NSODE predict a list of candidate ODE equations. 
To select one final equation, we numerically integrate all predicted candidates over the training interval $[0, T]$ and select the candidate with the best R$^2$ score between the trajectory of the predicted ODE and the actually observed trajectory.

\begin{figure*}[h!]
\centering
\begin{subfigure}{.33\textwidth}
    \centering
    \includegraphics[width=1\linewidth]{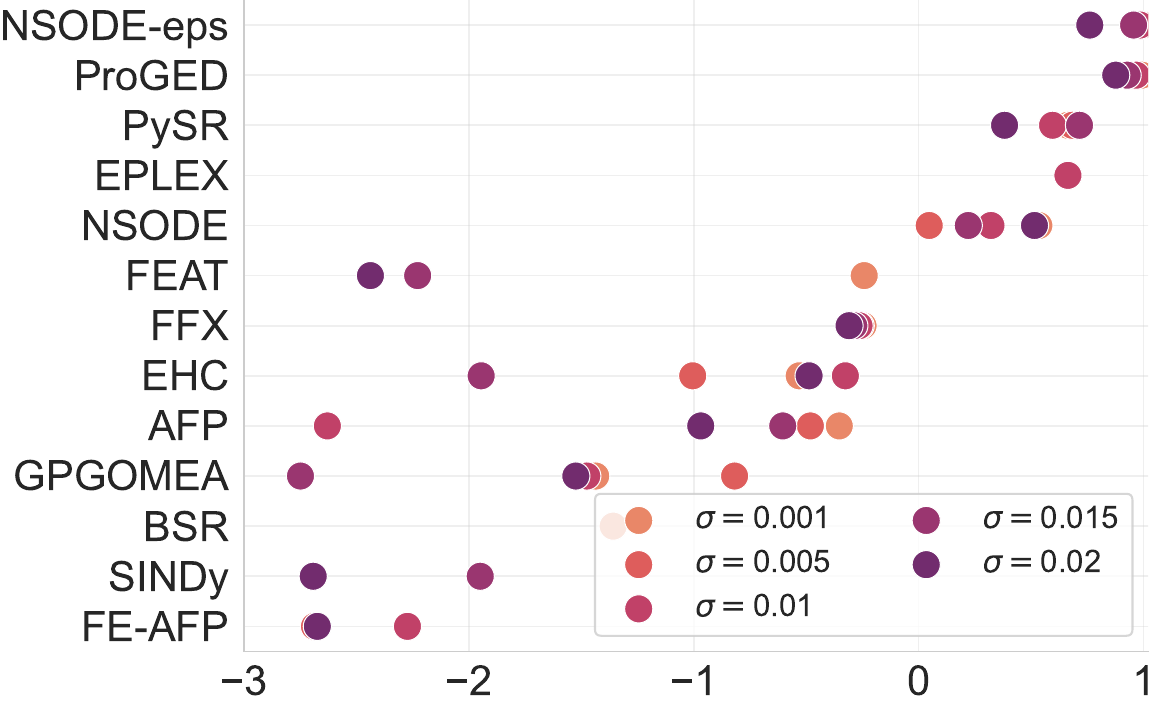}
    \caption{R$^2$ on \textbf{Classic}}
\end{subfigure}%
\begin{subfigure}{.33\textwidth}
    \centering
    \includegraphics[width=1\linewidth]{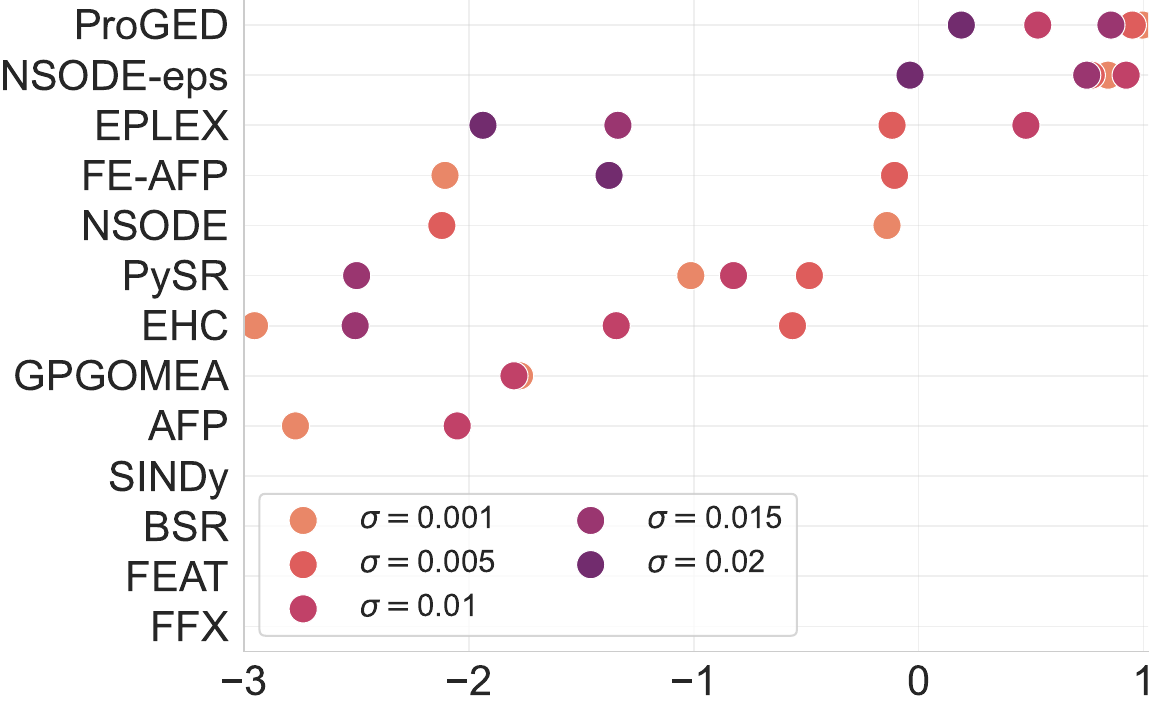}
    \caption{R$^2$ on \textbf{Textbook}}
\end{subfigure}%
\begin{subfigure}{.33\textwidth}
    \centering
    \includegraphics[width=1\linewidth]{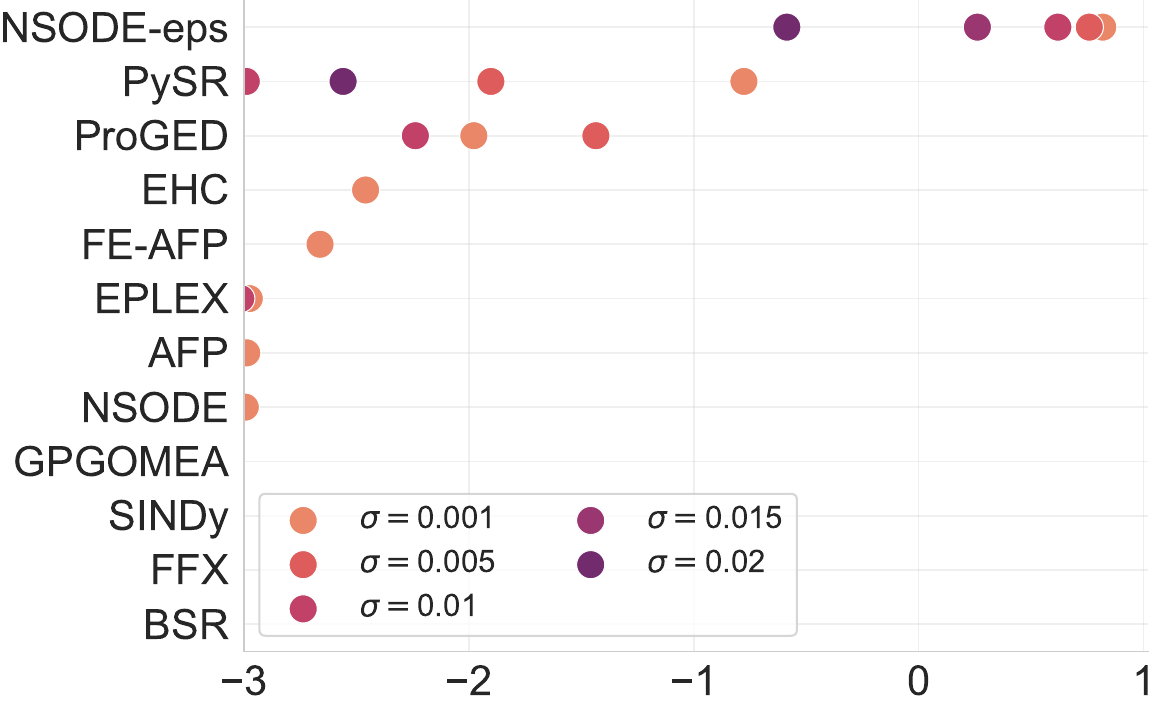}
    \caption{R$^2$ on \textbf{Large}}
\end{subfigure}%
\caption{\textbf{Extrapolation performance.} Median R$^2$ scores across predictions for different datasets with 128 randomly spaced time points for different noise levels $\sigma$ in the extrapolation regime $[T, T_{\mathrm{extra}}]$. The x-axes are restricted to the interval $[-3, 1]$; missing performances fall below this threshold.}
\label{fig:results_extrapolation}
\end{figure*}

\xhdr{Noise and irregular sampling}
To assess resilience to different signal-to-noise ratios we adopt the multiplicative noise model introduced in \citet{d2022deep}.
This noise model recognizes that zero-centered, fixed variance additive noise fails to take into account the magnitude of $y_i$ and may affect the signal-to-noise ratio too much or too little depending on the scale of the observations.
Instead we scale the standard deviation of additive, zero-centered Gaussian noise for~$y_i$ by $|y_i|$.
This can equivalently be modeled as multiplicative noise from a Gaussian distribution centered on $1$ where the choice of standard deviation $\sigma$ determines the signal-to-noise ratio with $\nicefrac{1}{\sigma}$.
We evaluate performances for $\sigma \in \{0.001, 0.005, 0.01, 0.015, 0.02\}$. 
We remark that these noise levels go beyond the noise level used for training NSODE-eps.
Additionally, we imitate irregularly spaced sampling intervals by uniformly randomly sub-sampling the original solution trajectory to $n \in \{128, 192, 256\}$ number of time points within the training interval $[0, T]$.

\xhdr{Extrapolation}
One motivation for symbolic regression is to find an expression that not only describes the observed data well, but that can also be used for extrapolation.
To assess the extrapolation capabilities of the predicted equations, we integrate the predicted ODEs on the adjacent interval $[T, T_{\mathrm{extra}}]$ and evaluate the accuracy metrics between the dense, noise-free ground truth trajectory and the predicted trajectory over this extrapolation interval.

\section{Results}\label{sec:results}

For a meaningful comparison, R$^2$ scores are always reported on the noiseless and equidistantly sampled trajectories after solving the ground truth and predicted ODEs on $[0,T]$ for interpolation and on $[T,T_{\mathrm{extra}}]$ for extrapolation, respectively.

\xhdr{Interpolation performance}
\cref{fig:results_r2_complexity_time} summarizes the performance of all models for different noise levels in terms of median accuracy (R$^2$), complexity, and inference time on \textbf{Classic} with $n=192$ time points.
Methods are ordered by the best R$^2$ across all noise levels for all three plots.
Our NSODE(-eps) and ProGED, the only other method not relying on finite difference approximations, are substantially more robust towards observation noise.
NSODE-eps and ProGED achieve essentially equal R$^2$ values.

In terms of parsimony, several models predict expressions of lower complexity than NSODE(-eps) and ProGED.
However, the median complexities of NSODE(-eps) and ProGED are in good agreement with the median complexity of the ground truth expressions (see \cref{fig:datastats} for full distributions).
Lower complexity is not necessarily better.

Finally, NSODE(-eps) is approximately one order of magnitude faster than ProGED, faring in the upper-mid range in terms of inference time overall.
While inference time is implementation and hardware dependent, we emphasize the conceptual difference that with NSODE(-eps) a single model can be used for all predictions, whereas all other models need to be fit separately per equation.

These results qualitatively generalize across our datasets as well as to different numbers of training points: \cref{fig:results_r2_datasets} shows that NSODE(-eps) again take 2nd and 3rd on \textbf{Classic}, \textbf{Textbook}, and \textbf{Large} using $n=128$ time points.
Again, NSODE-eps and ProGED perform similarly.
Additional results for all combinations of datasets and used time points $n$, as well as evaluations of the alternative accuracy metrics ($L_1$, $L_{\infty}$ norm, average \texttt{numpy.isclose}) can be found in \cref{app:full_results} and corroborate this overall trend.
We thus answer our initial question in the affirmative: \emph{Models tailored specifically to inferring dynamics are superior over (even highly optimized) re-purposed functional SR methods.}
NSODE-eps additionally scales efficiently.

\xhdr{Extrapolation}
Results for all testsets for $n=128$ time points in the extrapolation regime are shown in \cref{fig:results_extrapolation}.
Even in this challenging setting, NSODE-eps performs comparably to ProGED on the small \textbf{Classic} and \textbf{Textbook} testsets and is the only method producing reasonable results on \textbf{Large} for noise levels up to $\sigma=0.01$, which has been used during training.
\Cref{app:full_results} shows that these results also generalize to $n \in \{192, 256\}$.
While NSODE-eps and ProGED again outperform other models, their accuracy degrades on extrapolation tasks as the noise level increases.

\xhdr{Robustness}
\Cref{fig:results_r2_datasets,fig:results_extrapolation} show that NSODE-eps is considerably more robust to noisy observations than NSODE and works well even for noise levels two times higher than what has been used for training.
It also generalizes well to different numbers of (irregularly spaced) observations, and even manages extrapolation beyond the observed time range in a range of settings.

\section{Limitations}
Although NSODE comes with conceptual advantages over classical symbolic regression approaches for the task of ODE prediction, notably in that it does not require estimates of temporal derivatives, the presented approach also comes with a number of limitations. Perhaps most severely, we restrict ourself to the arguably most simple class of differential equations: explicit autonomous scalar first-order ODEs. These equations serve as a good initial benchmark but are limiting for applications in scientific discovery in practice. 
This restriction represents a design choice for the scope of this paper and does not necessarily imply a fundamental limitation of the approach: On the one hand, the model architecture can readily be extended to systems of equations, on the other hand we want to emphasize that systems of equations showcase a much richer set of qualitative behaviors, including oscillation and chaos, making model extensions towards them potentially non-trivial. As such it currently remains an open question and important challenge for future work to explore the scalability of the presented model paradigm for ODE prediction.

A second limitation of the presented model is that in its current form it cannot profit from multiple observations of the same process. In other words, even if we have multiple observed trajectories of the same process available, the model can only be applied to each trajectory individually. Allowing the model to profit from multiple observations appears to be a promising step to further increase its robustness to noise and irregularly sampled data.

\label{sec:limitations}

\section{Conclusion}
\label{sec:conclusion}

We presented NSODE, an efficiently scalable method to infer ordinary differential equations $\dot{y} = f(y)$ from a single observed solution trajectory.
NSODE follows the successful paradigm of large-scale pretraining of attention-based sequence-to-sequence models on essentially unlimited amounts of simulated data, where the inputs are the observed solution $\{(t_i,y_i)\}_{i=1}^n$ and the output is a symbolic expression for~$f$.
Once trained, our method performs on par or better than existing baselines and is an order of magnitude faster than similarly accurate symbolic regression techniques, which require a separate optimization for each expression.
NSODE is robust to different noise levels, the number of irregularly spaced samples and recovers dynamics that extrapolate beyond the observed time range.
While we have demonstrated the advantages of tailoring symbolic regression techniques specifically to recovering dynamics, interesting directions for future work include incorporating domain knowledge, and extending the framework to partial differential equations or high-dimensional systems of coupled differential equations.
Despite the huge potential of automated dynamical law learning for scientific discovery and hypothesis generation in the sciences, we caution against blindly trusting model outputs to represent generalizable real-world natural laws without rigorous experimental validation.

\section{Acknowledgements}
SB is supported by the Helmholtz Association under the joint research school ``Munich School for Data Science 
 - MUDS". This work was supported by the Helmholtz Association’s Initiative and Networking Fund on the HAICORE@FZJ partition.

\bibliography{references}
\bibliographystyle{icml2023}

\newpage
\appendix
\onecolumn

\section{Implementation Details}
\label{app:modeltraining}

\subsection{Rules to Resample Constants}
\label{app:constant_rules}
As described in \cref{xhdr:sampling}, we generate ODEs as unary-binary trees, convert them to infix notation and parse them into a canonical form using \texttt{sympy}\footnote{While \texttt{sympy} greatly helps with parsing functions into a canonical form, we remark that this is a pragmatic, best effort approach.}. For each skeleton we then create up to 25 ODEs by sampling different values for the constants. When resampling constants we want to ensure that we do not accidentally modify the skeleton as this would additionally burden our model with resolving potential ambiguities in the grammar of ODE expressions. Furthermore, we do not want to reintroduce duplicate samples on the skeleton level after carefully filtering them out previously. We therefore introduce the following sampling rules for constants:
\begin{enumerate}
    \item Do not sample constants of value 0.
    \item When the original constant in the skeleton is negative, sample a negative constant, otherwise sample a positive constant.
    \item Avoid base of 1 in power operations as $1^x = 1$.
    \item Avoid exponent of 1 and -1 in power operations as $x^1 = x$ and $x^{-1} = \nicefrac{1}{x}$.
    \item Avoid coefficients of value 1 and -1 as $1 \cdot x = x$ and $-1 \cdot x = -x$
    \item Avoid divisions by 1 and -1 as $x/1 = x$ and $x/-1 = -x$
\end{enumerate}

\subsection{Data generation}
\label{app:datageneration}

As discussed in the main text, the choices of the maximum number of internal nodes per tree $\nnodes$, the choice and distribution over $\nbinops$ binary operators, the choice and distribution over $\nunops$ unary operators, the probability with which to decorate a leaf with a symbol $\psymb$ (versus a constant with $1-\psymb$), and the distribution $\pconst$ over constants uniquely determine the training distribution over ODEs~$f$.
These choices can be viewed as flexible and semantically interpretable tuning knobs to choose a prior over ODEs.
For example, it may be known in a given context, that the system follows a ``simple'' law (small $\nnodes$) and does not contain exponential rates of change (do not include $\exp$ in the unary operators), and so on.
The choice of the maximum number of operators per tree, how to sample the operators, and how to fill in the leaf nodes define the training distribution, providing us with flexible and semantically meaningful tuning knobs to choose a prior over ODE systems for our model.
We summarize our choices in \cref{tab:dataparams,tab:binary_operators,tab:unary_operators}, where $\mathcal{U}$ denotes the uniform distribution.
Whenever a leaf node is decorated with a constant, the distribution over constants is determined by first choosing with equal probability whether to use an integer or a real value.
In case of an integer, we sample it from $\pint$, and in case of a real-valued constant we sample it from $\preal$ shown in \cref{tab:dataparams}. Finally, when it comes to the numerical solutions of the sampled ODEs, we fixed the parameters in \cref{tab:solutionparams} for our experiments.

We highlight that there is no such thing as ``a natural distribution over equations'' when it comes to ODEs.
Hence, ad-hoc choices have to be made in one way or another.
However, it is important to note that neither our chosen range of integers nor the range of real values for constants are in any way restrictive as they can be achieved by appropriate rescaling.
In particular, the model itself represents these constant values merely be non-numeric tokens and interpolates between those anchor tokens (our two-hot encoding) to represent continuous values.
Hence, the model is entirely agnostic to the actual numerical range spanned by these fixed grid tokens, but the relative accuracy in recovering interpolated values will be constant and thus scale with the absolute chosen range.
Therefore, scaling $\pint$ and $\preal$ by essentially any power of 10 does not affect our findings.
Similarly, the chosen range of initial values $\ivrange$ is non-restrictive as one could simply scale each observed trajectory to have its starting value lie within this range.

\begin{table}[b!]
    \caption{Parameter settings for the data generation.}\label{tab:dataparams}
    \centering
    \begin{tabularx}{0.7\columnwidth}{@{}lccccYc@{}}
    \rowcolor{white}
    \toprule
    parameter  & $\nnodes$ & $\nbinops$ & $\nunops$ & $\psymb$ & $\pint$ & $\preal$ \\ \midrule
    value      & 5 & 5 & 5 & 0.5 & $\mathcal{U}(\{-10, \ldots, 10\} \setminus \{0\})$ & $\mathcal{U}((-10, 10))$ \\
    \bottomrule
    \end{tabularx}%
\end{table}

\begin{table}[t!]
    \caption{Binary operators with their relative sampling frequencies}\label{tab:binary_operators}
    \centering
    \begin{tabularx}{0.5\columnwidth}{l*5{Y}}
    \rowcolor{white}
    \toprule
    operator  & $+$ & $-$ & $\cdot$ & $\div$ & pow \\ \midrule
    \rowcolor{white}
    probability & 0.2 & 0.2 & 0.2     & 0.2    & 0.2\\
    \bottomrule
    \end{tabularx}%
\end{table}

\begin{table}[t!]
    \caption{Unary operators with their relative sampling frequencies.}\label{tab:unary_operators}
    \centering
    \begin{tabularx}{0.5\columnwidth}{l*5{Y}}
    \rowcolor{white}
    \toprule
    operator  & $\sin$ & $\cos$ & $\exp$ & $\sqrt{\phantom{x}}$ & $\log$ \\ \midrule
    probability & 0.2    & 0.2    & 0.2    & 0.2     & 0.2 \\
    \bottomrule
    \end{tabularx}%
\end{table}

\begin{table}[t!]
    \caption{Parameters for numerical solutions of sampled ODEs.}\label{tab:solutionparams}
    \centering
    \begin{tabularx}{0.6\columnwidth}{l*5{c}Y}
    \rowcolor{white}
    \toprule
    parameter  & $\nconsts$ & $\nivs$ & $\maxtime$ & $T_{\mathrm{extra}}$ & $\ngrid$ & $\ivrange$  \\ \midrule
    \rowcolor{white}
    value & 25    & 25    & 2  &  4  & 1024     & $(-5,5)$ \\
    \bottomrule
    \end{tabularx}%
\end{table}

\subsection{Model}
\label{app:modeldesign}

For our Transformer model we choose the implementation of BigBird \citep{zaheer2020big} available in HuggingFace.
The model is trained on an internal academic compute cluster using 4 Nvidia A100 GPUs for 25 epochs after which we evaluate the best model based on the validation loss. We choose a batchsize of 600 and use a linear learning rate warm-up over 10,000 optimization step after which we keep the learning rate constant at $10^{-4}$.
For the fixed tokens that are used to decode constants, we choose an equidistant grid $-10=x_1 < x_2 < \ldots < x_m = 10$ with $m=21$.
This worked well empirically and using fewer or more tokens did not seem to improve model performance substantially. We note that architecture and hyperparameter choices correspond to ad-hoc decision and were not systematically optimized. We use the same choices in all experiments.

While not relevant for our dataset as we check for convergence of the ODE solvers, we remark that the input-encoding via IEEE-754 binary representations also graciously represents special values such as \texttt{nan} or \texttt{inf} without causing errors.
Those are thus valid inputs that may still provide useful training signal, e.g., ``the solution of the ODE of interest goes to \texttt{inf} quickly''.

\section{Textbook equations dataset}
\label{app:textbook_equations}
\Cref{tab:textbook_equations} list the equations we collected from wikipedia, textbooks and lecture notes together with the initial values that we solved them for. We can also see that almost all of these equations simplify to low-order polynomials.

\begin{table}[h!]
\centering
\caption{Equations of the \textbf{Textbook} testset.\label{tab:textbook_equations}}

\begin{tabularx}{\columnwidth}{lYYr}
\toprule \rowcolor{white}
\textbf{Name} & \textbf{Equation} $f(x)$ & \textbf{simplified} &  $\mathbf{y_0}$ \\ \midrule
autonomous Riccati                                                                     & $0.6\cdot y^2+2\cdot y+0.1$                                                         & $0.6\cdot y^2+2\cdot y+0.1$      & $-0.2$          \\
autonomous Stuart-Landau                                                               & $-2.2/2\cdot y^3 + 1.31\cdot y$                                                      & $-1.1\cdot y^3 + 1.31\cdot y$          & $0.1$           \\ 
autonomous Bernoulli                                                                   & $-1.3\cdot y+2.1\cdot y^{2.2}$                                                       & $-1.3\cdot y + 2.1\cdot y^{2.2}$         & $0.6$           \\ 
compound interest                                                                      & $0.1\cdot y$                                                                         & $0.1\cdot y$                       & $4.9$             \\ 
Newton's law of cooling                                                                & $-0.1\cdot(y-3)$                                                                     & $0.3-0.1\cdot y$                 & $4.9$             \\ 
Logistic equation                                                                      & $0.23\cdot y\cdot (1-y)$                                                             & $0.23\cdot(y-y^2)$            & $4.9$             \\ 
\begin{tabular}[c]{@{}l@{}}Logistic equation \\ with harvesting\end{tabular}           & $0.23\cdot y\cdot (1-0.33\cdot y) - 0.5$                                             & $0.23\cdot y-0.76\cdot y^2-0.5$ & $3.5$             \\ 
\begin{tabular}[c]{@{}l@{}}Logistic equation \\ with harvesting 2\end{tabular}         & $2\cdot y\cdot(1-y/3) - 0.5$                                                         & $2\cdot y-0.66\cdot y^2-0.5$ & $0.7$           \\ 
Solow-Swan                                                                             & \begin{tabular}[c]{@{}l@{}}$y^0.5\cdot (0.9\cdot  8 - (3 + 2.5)\cdot$\\$y^{1- 0.5})$ \end{tabular}                              & $7.2\cdot y^{0.5}-5.5\cdot y$   & $0.1$           \\ 
Tank draining                                                                          & $-\sqrt{2\cdot 9.81}\cdot (2/9)^2\cdot\sqrt{y}$                                       & $-0.21\cdot y^{0.5}$                & $1$             \\ 
\begin{tabular}[c]{@{}l@{}}Draining water \\ through a funnel\end{tabular}             & \begin{tabular}[c]{@{}l@{}}$-(0.5^2/4)\cdot\sqrt{2\cdot 9.81}\cdot$\\ $(\sin{1}/\cos{1})^2\cdot y^{-1.5}$\end{tabular} & $-0.67/y^{1.5}$                & $3$             \\ 
\begin{tabular}[c]{@{}l@{}}velocity of a body\\ thrown vertically upwards\end{tabular} & $-9.81 - 0.9\cdot  y/8.2$                                             & $-0.10\cdot y - 9.81$              & $0.1$           \\ 
\bottomrule
\end{tabularx}
\end{table}

\newpage
\section{Dataset statistics}\label{app:datastats}

\def\widthfrac{0.8}
\def\spacing{2mm}
\def\hspacing{2mm}
\begin{figure}[th!]
  \centering
  \textbf{Training data}\\[\hspacing]
  \includegraphics[width=\widthfrac\columnwidth]{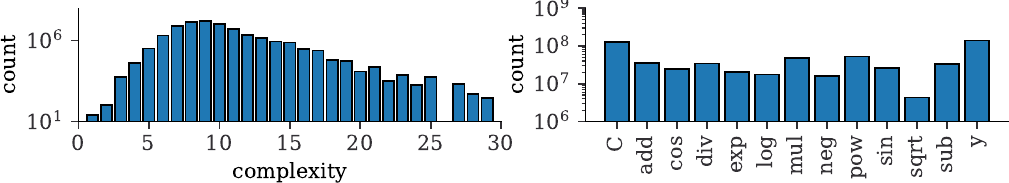}\\[\spacing]
  \textbf{Testset Large}\\[\hspacing]
  \includegraphics[width=\widthfrac\columnwidth]{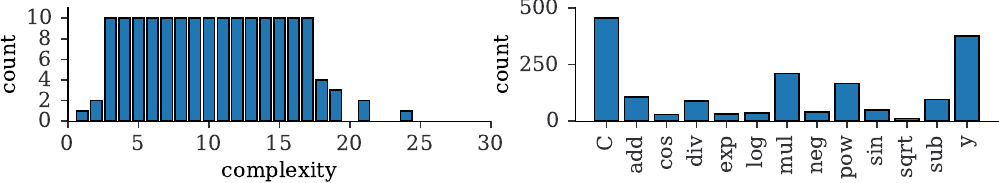}\\[\spacing]
  \textbf{Testset Classic}\\[\hspacing]
  \includegraphics[width=\widthfrac\columnwidth]{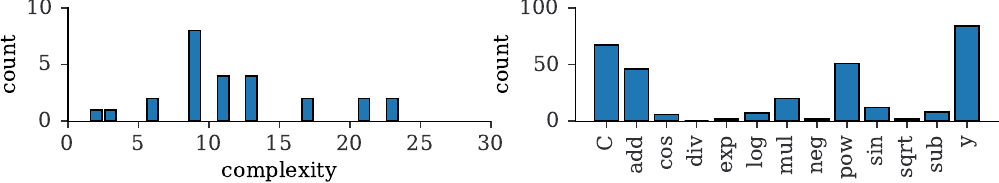}\\[\spacing]
  \textbf{Testset Textbook}\\[\hspacing]
  \includegraphics[width=\widthfrac\columnwidth]{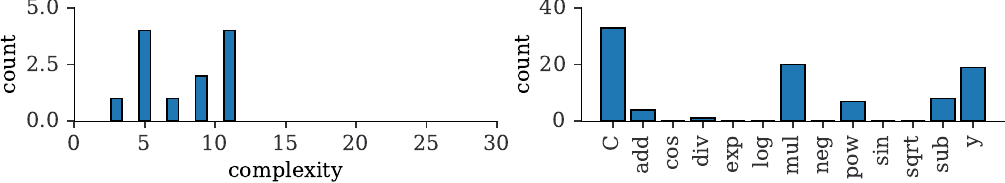}
  \caption{Distribution of complexity and operators for all datasets.\label{tab:datastats}}
  \label{fig:datastats}
\end{figure}

We provide an overview over the complexity distribution and the absolute frequency of all operators (after simplification) for all datasets in \cref{tab:datastats}. We can see that our self-generated dataset covers by far the larges complexity whereas both complexities and operator diversity are much lower for equations in the \textbf{Classic} and \textbf{Textbook} ODEs.

\section{Baselines}
\label{app:baselines}
We here describe more detail on the optimization of the baseline comparison models. Most models have a number of hyperparameters which need to be optimized per equation. Unless specified below, we use the default hyperparameters values and hyperparameter grid search settings specified in the implementation alongside the benchmark study by \cite{la2021contemporary}. Whenever supported by a model's implementation we use \texttt{GridSearchCV} from \texttt{scikit-learn} \citep{pedregosa2011scikit} for hyperparameter optimization. For this optimization we split the observed trajectory into a training interval $[0, 1]$ and a validation interval $[1, 2]$. In order to obtain results in reasonable time, we set a runtime limit of 3 minutes per hyperparameter optimization run.

\xhdr{AFP, EHC, EPLEX \& FE\_AFP}\\
op\_list=[`n',`v',`+',`-',`*',`/',`exp',`log',`2',`3', `sqrt',`sin',`cos']

\xhdr{FEAT}\\
functions= `+, -, *, /, \^{}2, \^{}3, sqrt, sin, cos, exp, log, \^{}'

\xhdr{PySR}\\
niterations=40\\ 
binary\_operators=[`plus', `sub', `mult', `pow', `div']\\
unary\_operators=[`cos', `exp', `sin', `neg', `log', `sqrt']

\xhdr{ProGED}\\
sample\_size=64\\
task\_type = `differential'

\xhdr{SINDy} We use the implementation available in PySINDy \citep{pysindy} and instantiate the basis functions with polynomials up to degree 10 as well as all unary operators listed in \cref{tab:unary_operators}. When fitting SINDy to data we often encountered numerical issues especially when using high-degree polynomial or the exponential function. To attenuate such issues discard the particular basis function that raised a numerical error and restart the fitting process. We remark that removing basis functions and restarting the optimization is practically feasible for SINDy due to its extremely fast runtime. At the same time, being a regression based model, SINDy can (in contrast to genetic programming based models) not easily recover from a numerical issue caused by a particular basis function. We run a separate full grid search per (sample x noise level x number of time points) over the following hyperparameters and respective values (these all include the default values): 
\begin{itemize}[leftmargin=*,topsep=0pt,itemsep=0pt]
    \item optimizer-threshold (\texttt{np.logspace(-5, 0, 10)}): Minimum magnitude for a coefficient in the weight vector to not be zeroed out.
    \item optimizer-alpha ($[0.001, 0.0025, 0.005, 0.01, 0.025, 0.05, 0.1, 0.2]$): L2 regularizer on parameters.
    \item finite differences order ($[2, 3, 5, 7, 9]$): Order of finite difference approximation.
    \item maximum number of optimization iterations ([20, 100]): Maximum number of optimization steps.
\end{itemize}

\xhdr{Model selection}
Most models provide a list of candidate solutions for each sample, e.g. the pareto-front (accuracy vs complexity tradeoff) in genetic processing based methods. To obtain a single predicted equation per model we use the model selection procedure outlined in \ref{sec:experiments}.

\section{Detailed results}
\label{app:full_results}
Here we provide detailed results on all experimental conditions across all datasets. We start with results on the interpolation interval $[0, \maxtime]$ in \cref{app:full_results_intra} before showing results on the extrapolation interval $[\maxtime, T_{\mathrm{extra}}]$ in  \cref{app:full_results_intra}. To facilitate navigation we provide an overview in \cref{tab:result_overview_table}.

\begin{table}[h!]
    \caption{Result overview.}\label{tab:result_overview_table}
    \centering
    \begin{tabularx}{0.5\columnwidth}{l*2{Y}}
    \rowcolor{white}
    \toprule
    Dataset  & interval & fig. number \\ \midrule
    \rowcolor{white}
    Classic  & interpolation & \cref{app:results_classic_intra} \\
    Textbook & interpolation & \cref{app:results_textbook_intra} \\
    Large    & interpolation & \cref{app:results_large_intra} \\
    \bottomrule
    Classic  & extrapolation & \cref{app:results_classic_extra}  \\
    Textbook & extrapolation & \cref{app:results_textbook_extra} \\
    Large    & extrapolation & \cref{app:results_large_extra} \\
    \bottomrule
    \end{tabularx}%
\end{table}

\subsection{Interpolation results}
\label{app:full_results_intra}
\begin{figure*}[h!]
\centering
\begin{subfigure}{.3\textwidth}
    \centering
    \includegraphics[width=1\linewidth]{figs/performance/classic_r2_128.pdf}
    \caption{R$^2$, n=128}
\end{subfigure}%
\begin{subfigure}{.3\textwidth}
    \centering
    \includegraphics[width=1\linewidth]{figs/performance/classic_r2_192.pdf}
    \caption{R$^2$, n=192}
\end{subfigure}%
\begin{subfigure}{.3\textwidth}
    \centering
    \includegraphics[width=1\linewidth]{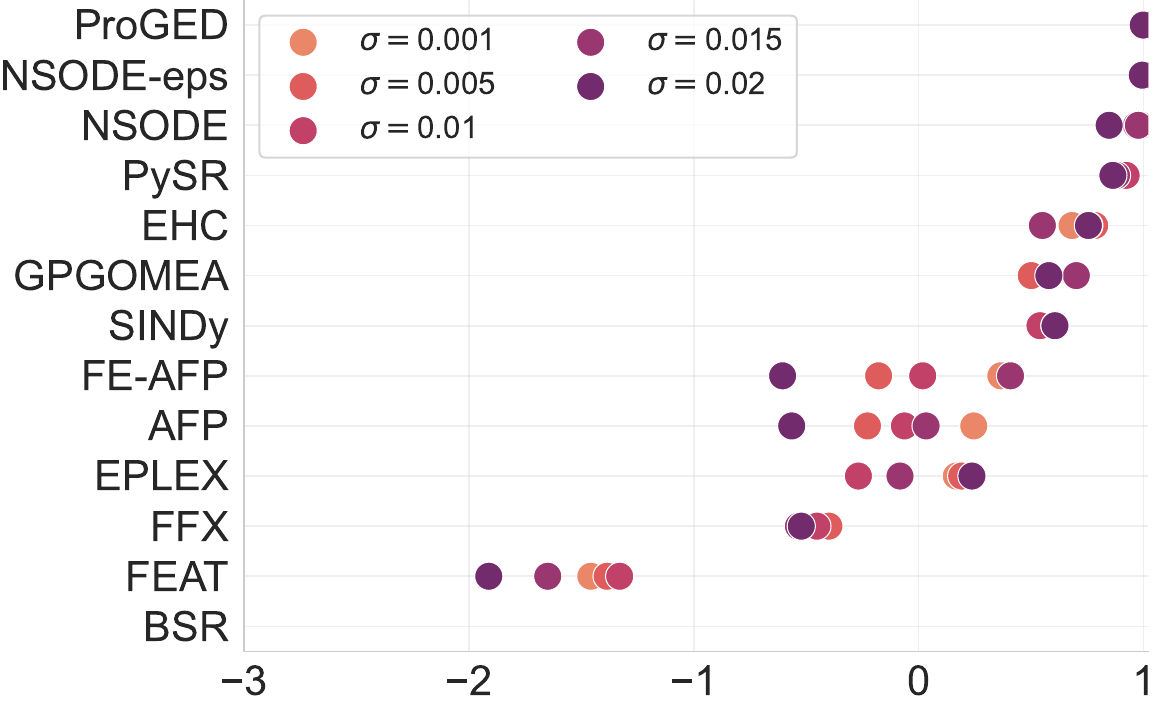}
    \caption{R$^2$, n=256}
\end{subfigure}
\begin{subfigure}{.3\textwidth}
    \centering
    \includegraphics[width=1\linewidth]{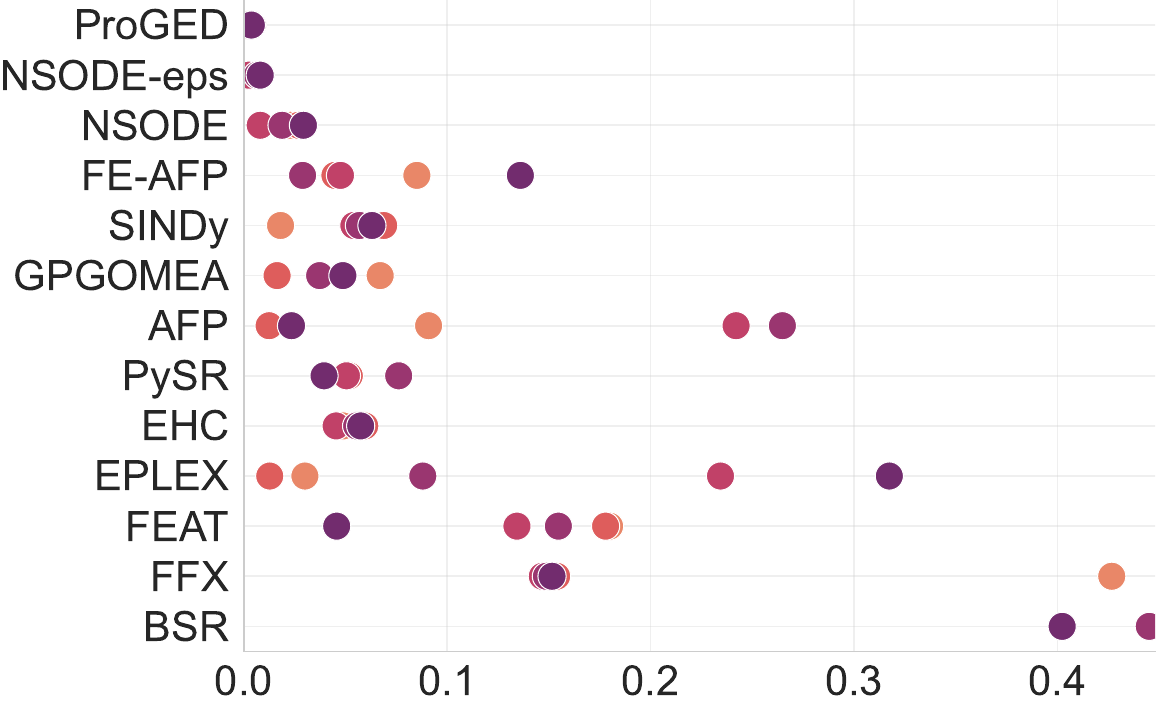}
    \caption{$L_1$, n=128}
\end{subfigure}%
\begin{subfigure}{.3\textwidth}
    \centering
    \includegraphics[width=1\linewidth]{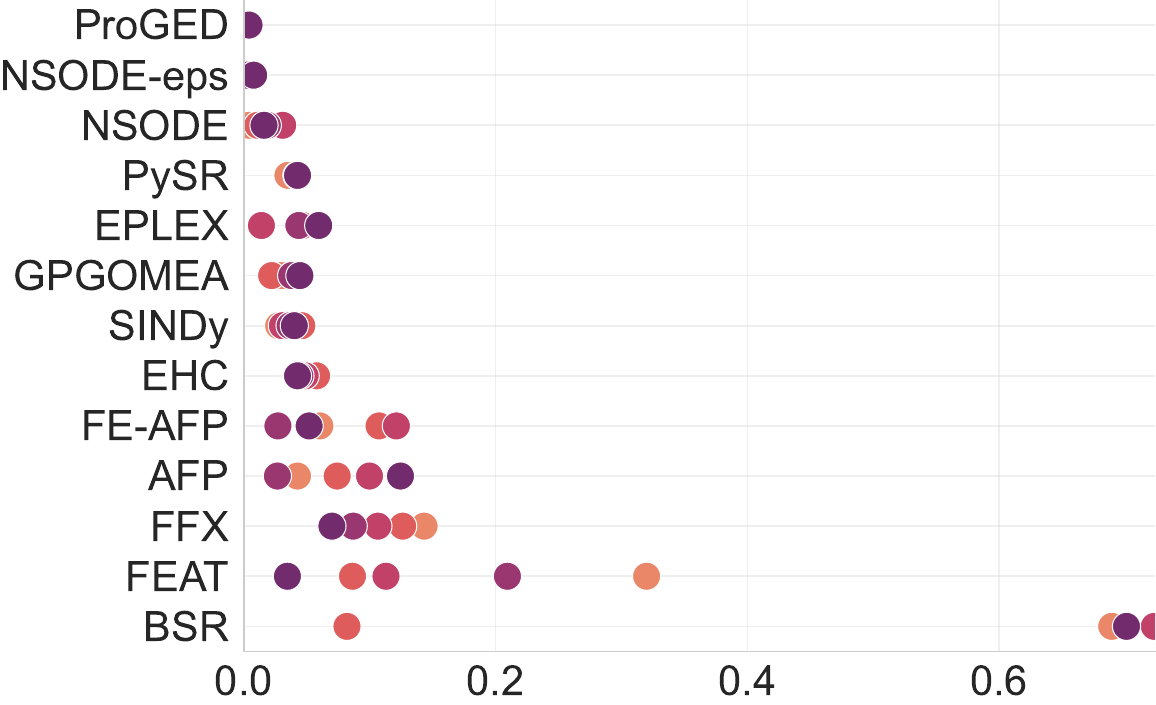}
    \caption{$L_1$, n=192}
\end{subfigure}%
\begin{subfigure}{.3\textwidth}
    \centering
    \includegraphics[width=1\linewidth]{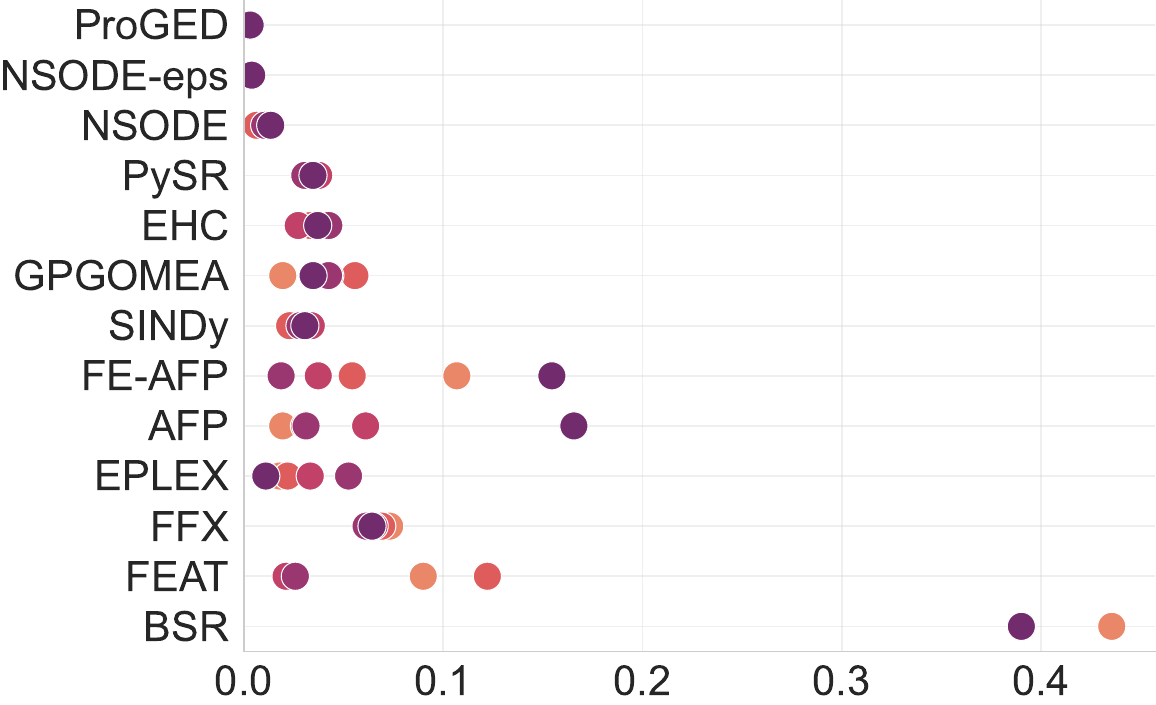}
    \caption{$L_1$, n=256}
\end{subfigure}
\begin{subfigure}{.3\textwidth}
    \centering
    \includegraphics[width=1\linewidth]{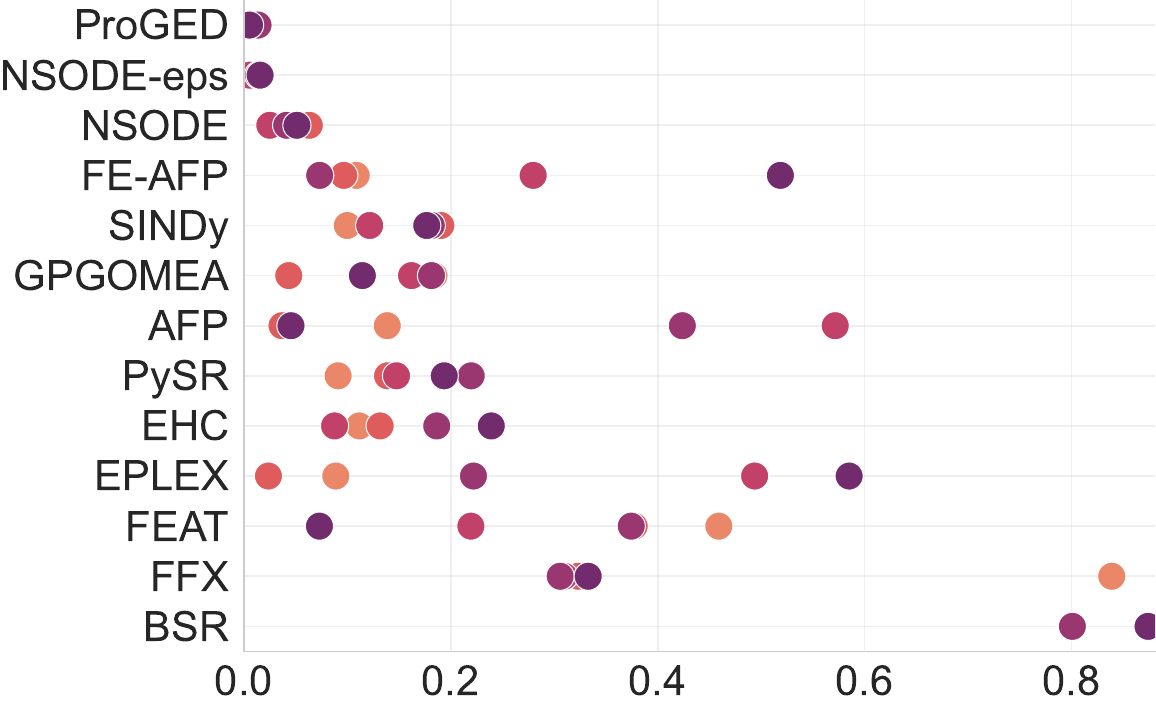}
    \caption{$L_{\infty}$, n=128}
\end{subfigure}%
\begin{subfigure}{.3\textwidth}
    \centering
    \includegraphics[width=1\linewidth]{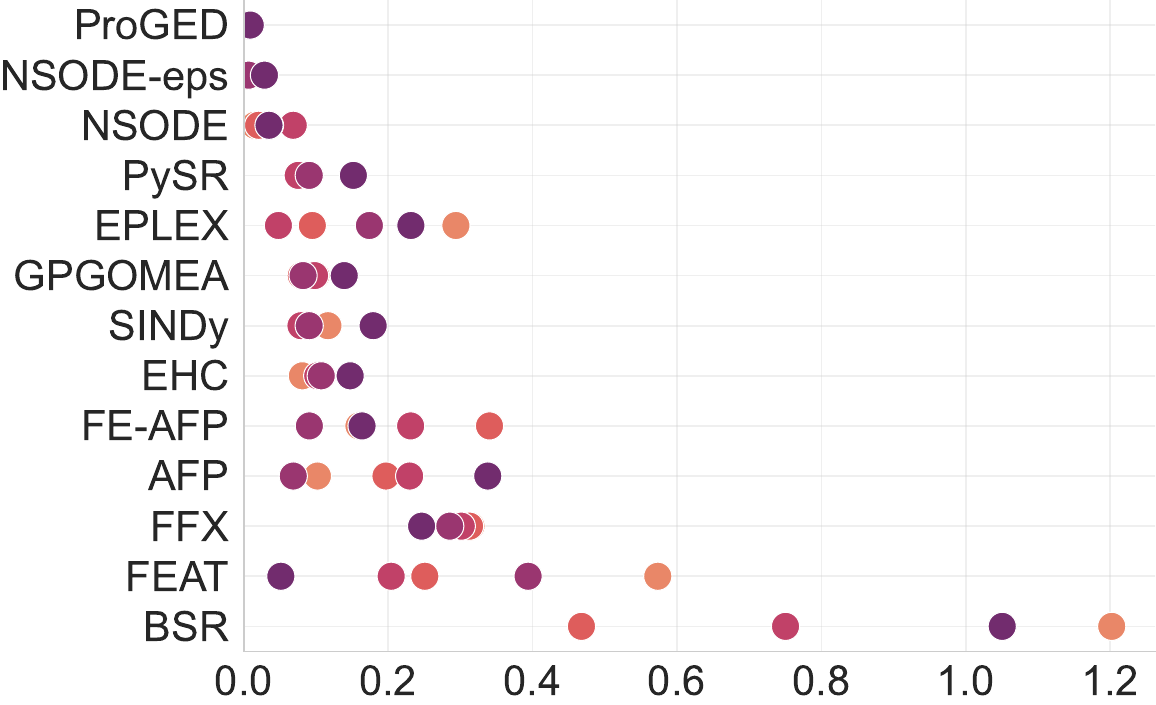}
    \caption{$L_{\infty}$, n=192}
\end{subfigure}%
\begin{subfigure}{.3\textwidth}
    \centering
    \includegraphics[width=1\linewidth]{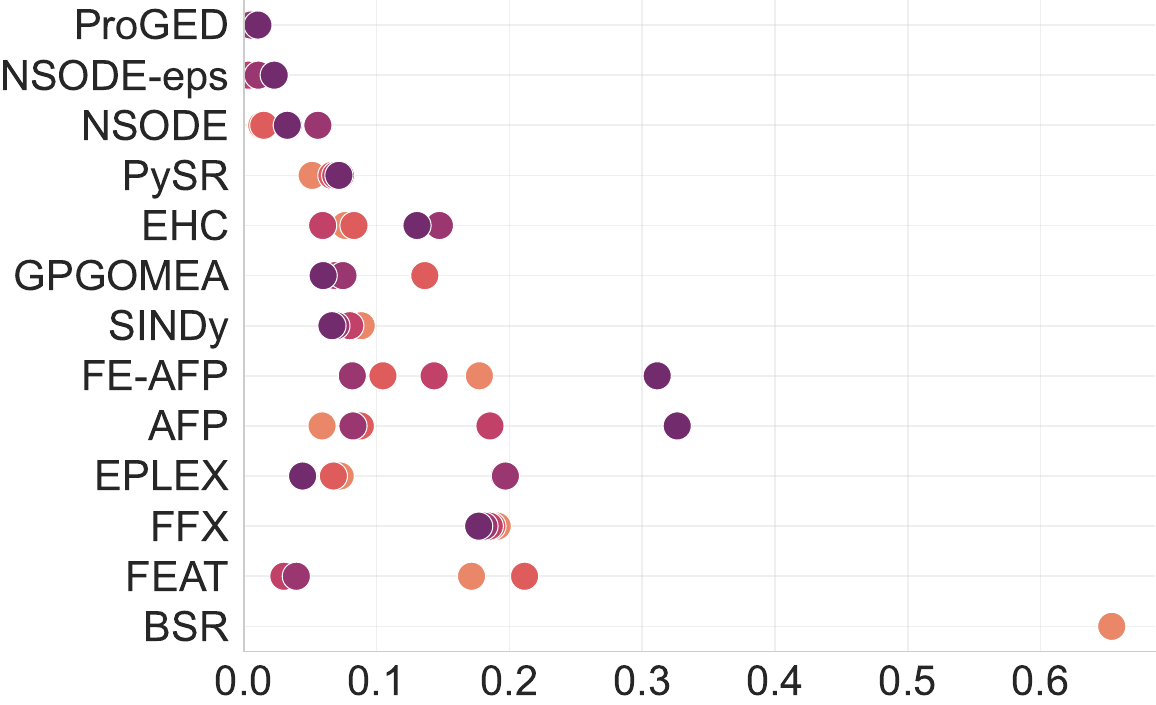}
    \caption{$L_{\infty}$, n=256}
\end{subfigure}
\begin{subfigure}{.3\textwidth}
    \centering
    \includegraphics[width=1\linewidth]{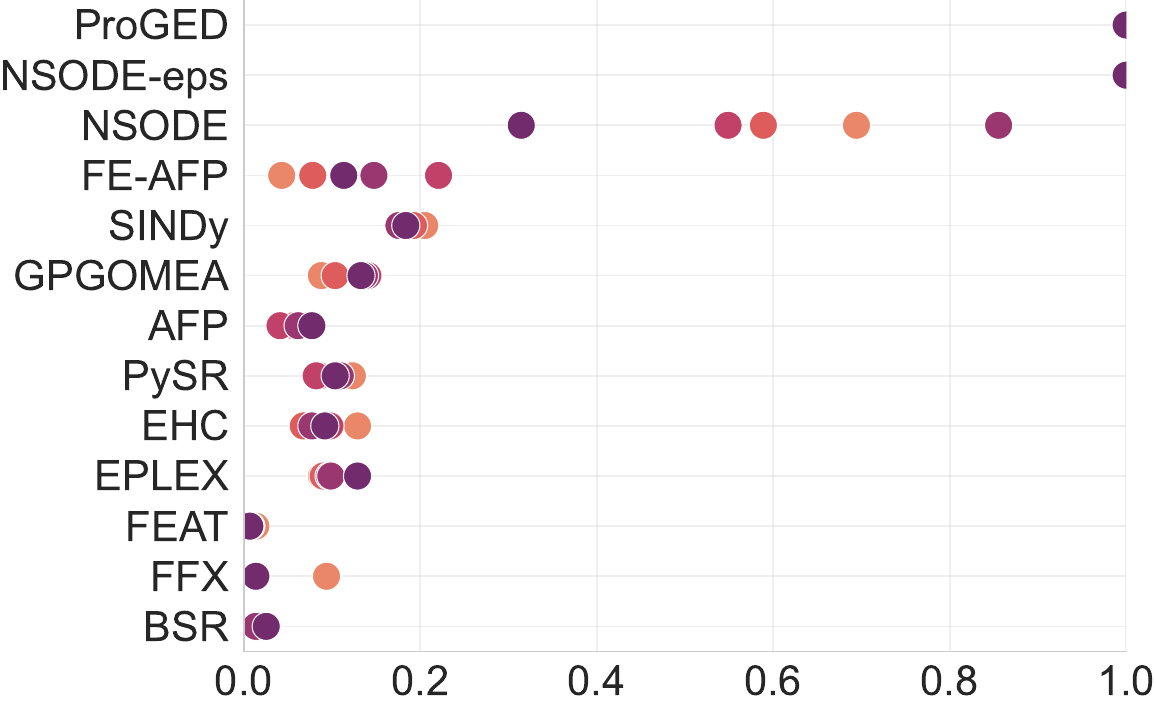}
    \caption{$\nicefrac{\texttt{isclose}}{n}$, n=128}
\end{subfigure}%
\begin{subfigure}{.3\textwidth}
    \centering
    \includegraphics[width=1\linewidth]{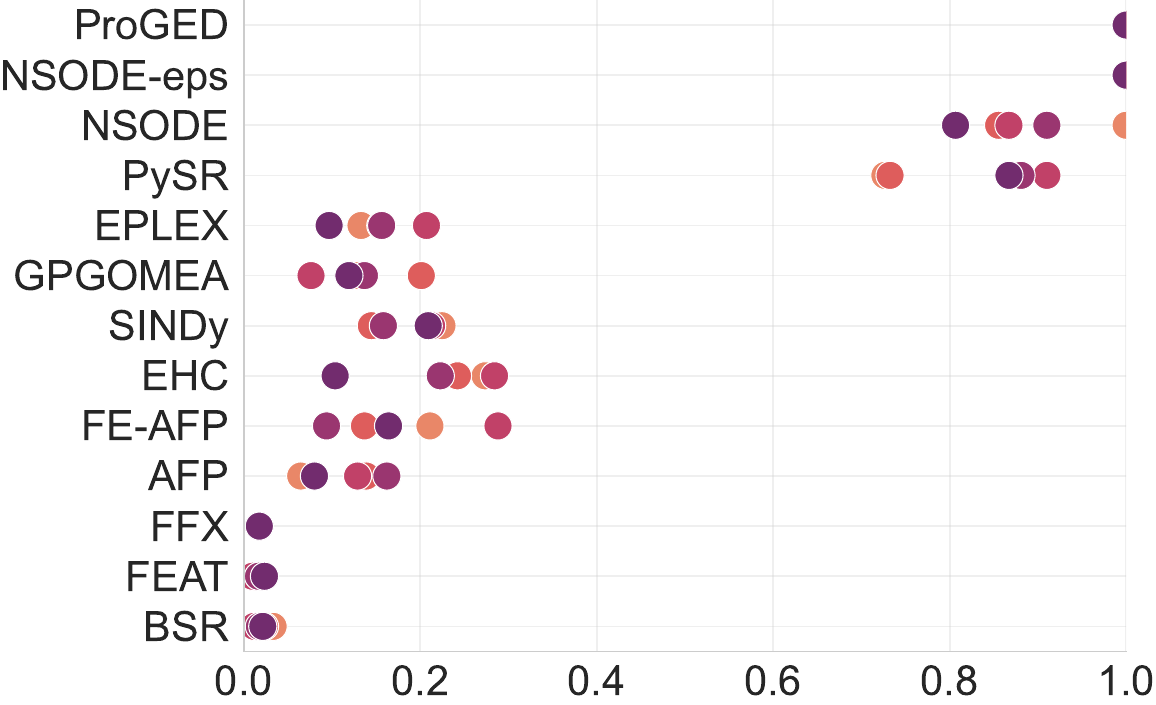}
    \caption{$\nicefrac{\texttt{isclose}}{n}$, n=192}
\end{subfigure}%
\begin{subfigure}{.3\textwidth}
    \centering
    \includegraphics[width=1\linewidth]{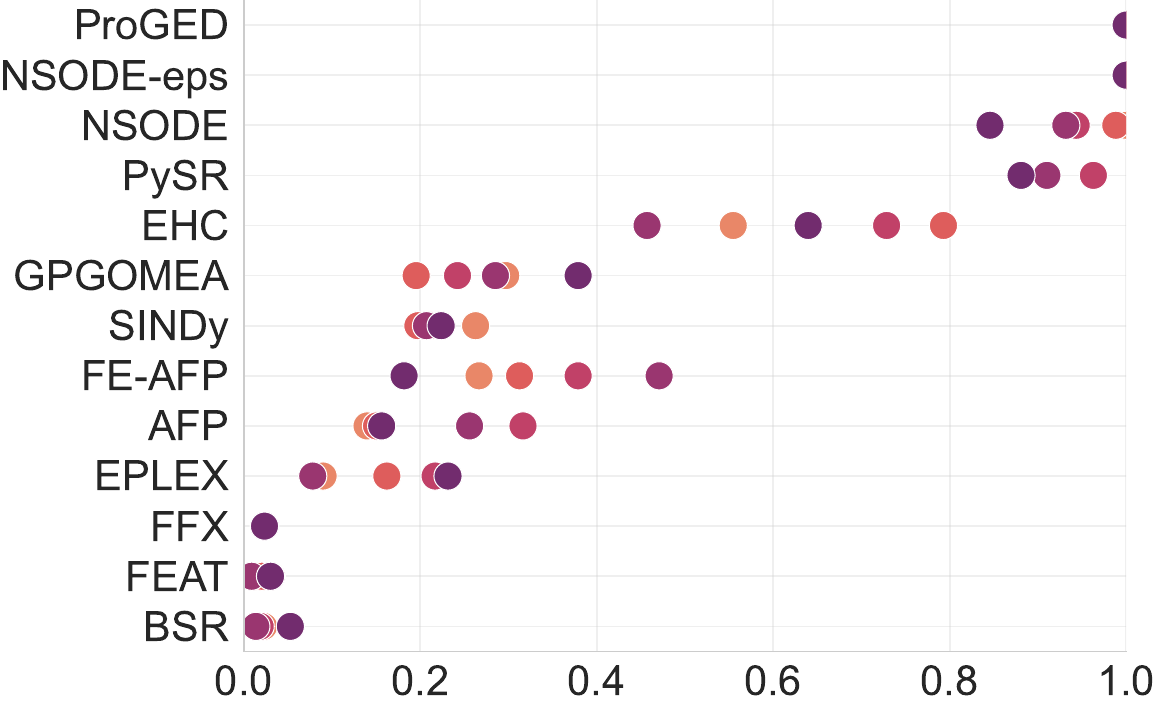}
    \caption{$\nicefrac{\texttt{isclose}}{n}$, n=256}
\end{subfigure}
\begin{subfigure}{.3\textwidth}
    \centering
    \includegraphics[width=1\linewidth]{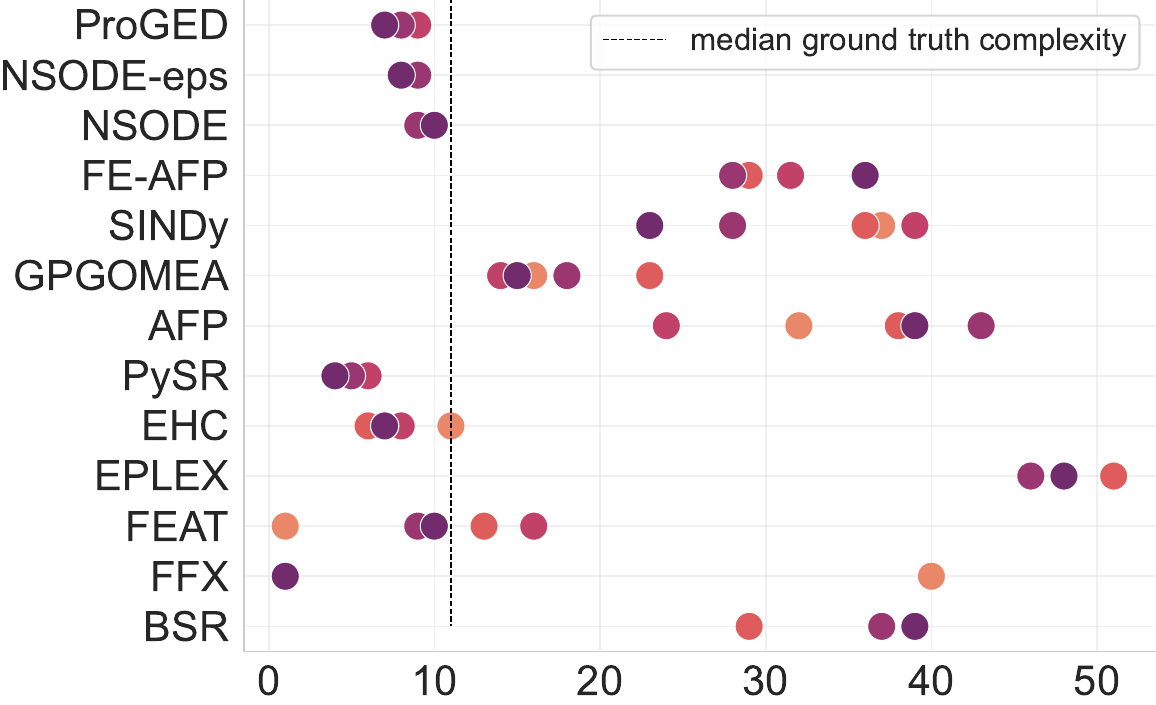}
    \caption{complexity, n=128}
\end{subfigure}%
\begin{subfigure}{.3\textwidth}
    \centering
    \includegraphics[width=1\linewidth]{figs/performance/classic_complexity_192.pdf}
    \caption{complexity, n=192}
\end{subfigure}%
\begin{subfigure}{.3\textwidth}
    \centering
    \includegraphics[width=1\linewidth]{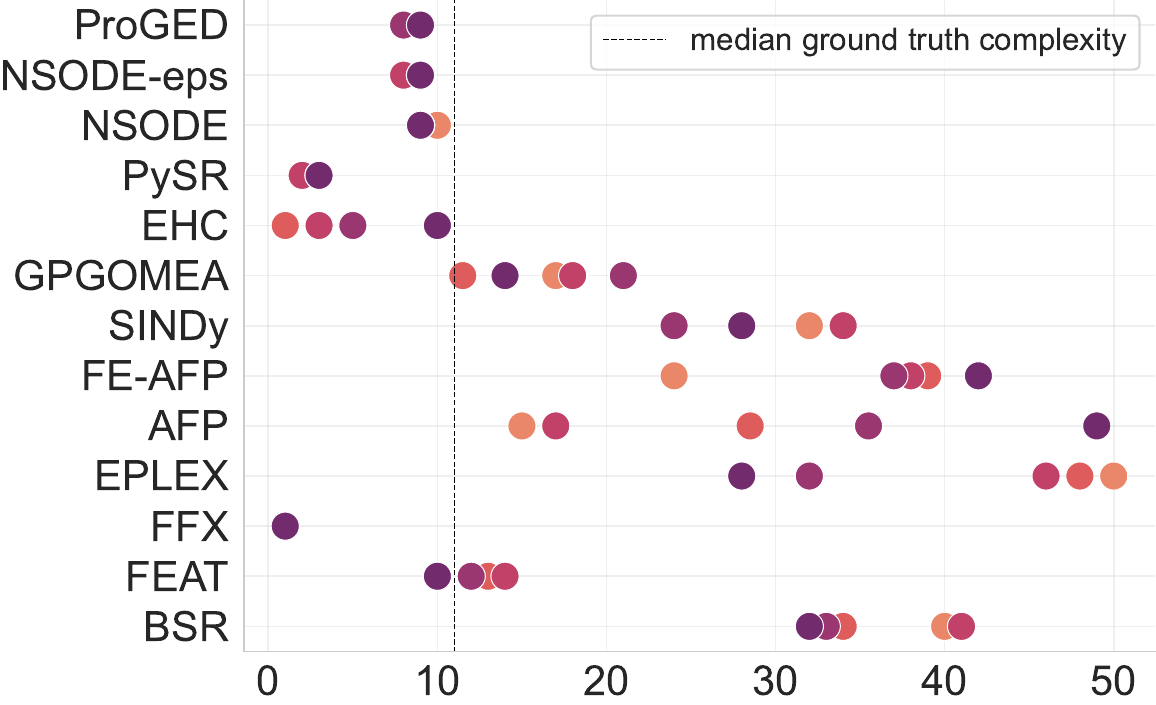}
    \caption{complexity, n=256}
\end{subfigure}
\begin{subfigure}{.3\textwidth}
    \centering
    \includegraphics[width=1\linewidth]{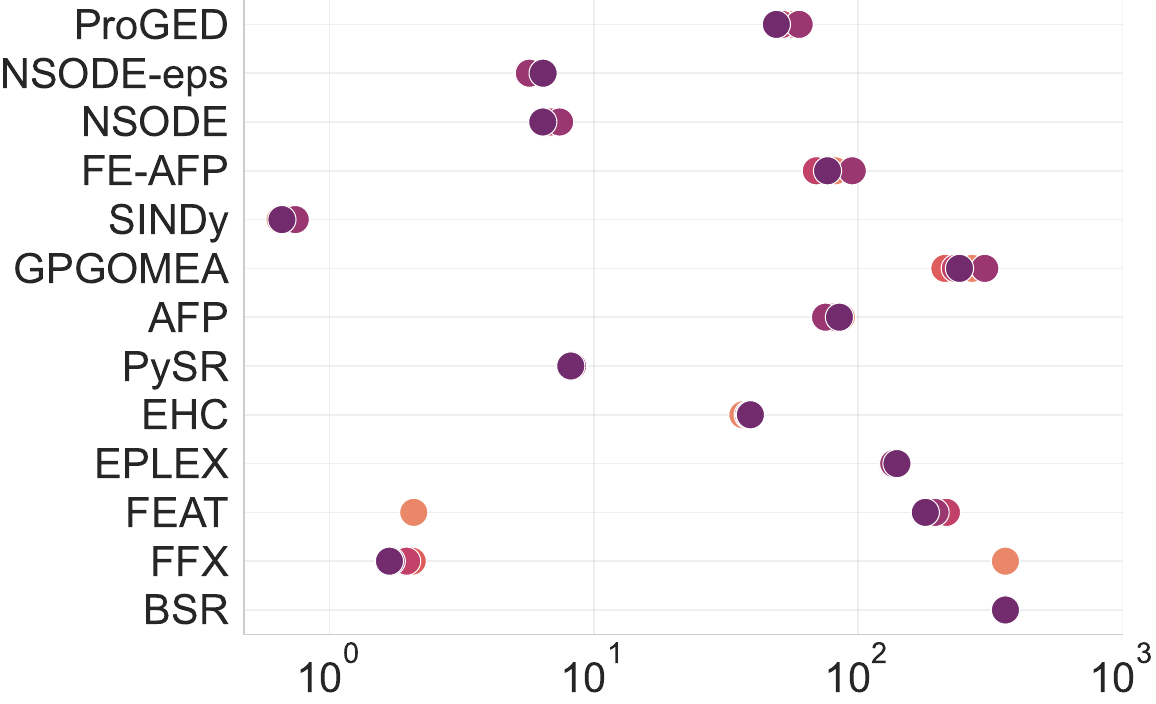}
    \caption{inference time [sec], n=128}
\end{subfigure}%
\begin{subfigure}{.3\textwidth}
    \centering
    \includegraphics[width=1\linewidth]{figs/performance/classic_time_192.pdf}
    \caption{inference time [sec], n=192}
\end{subfigure}%
\begin{subfigure}{.3\textwidth}
    \centering
    \includegraphics[width=1\linewidth]{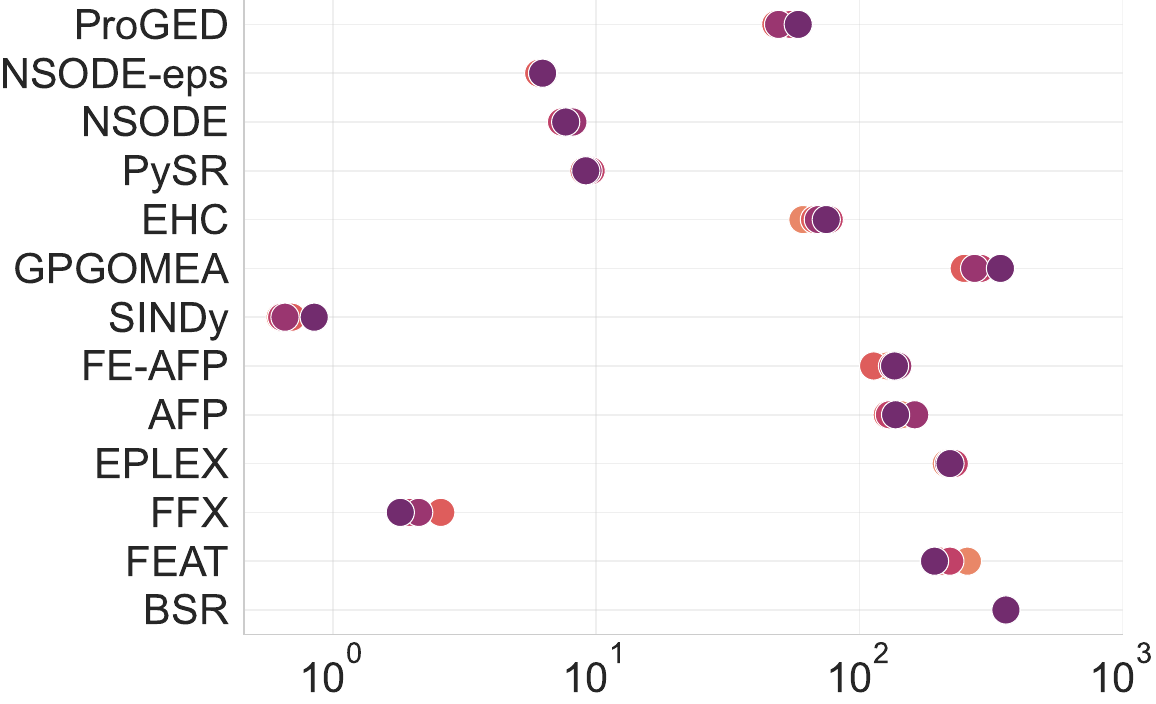}
    \caption{inference time [sec], n=256}
\end{subfigure}%
\caption{Interpolation. Median scores on \textbf{Classic} for $n$ irregularly sampled time points across different noise levels $\sigma$.}
\label{app:results_classic_intra}
\end{figure*}

\begin{figure*}[h!]
\centering
\begin{subfigure}{.3\textwidth}
    \centering
    \includegraphics[width=1\linewidth]{figs/performance/knownODEs_r2_128.pdf}
    \caption{R$^2$, n=128}
\end{subfigure}%
\begin{subfigure}{.3\textwidth}
    \centering
    \includegraphics[width=1\linewidth]{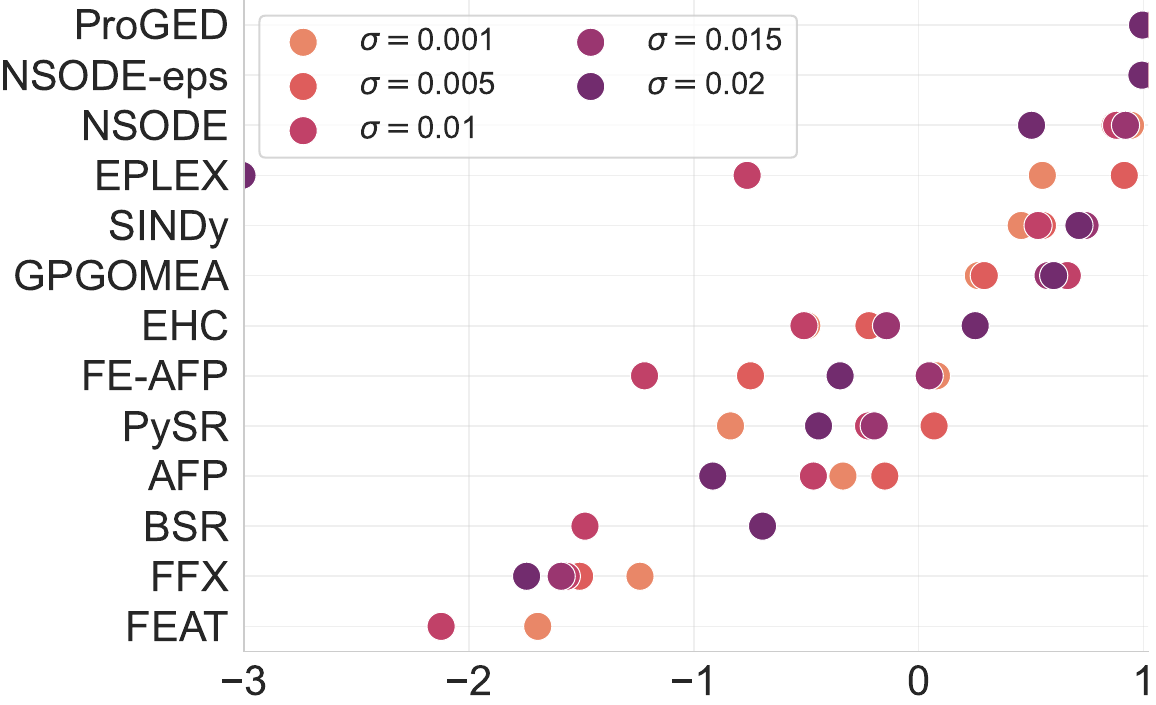}
    \caption{R$^2$, n=192}
\end{subfigure}%
\begin{subfigure}{.3\textwidth}
    \centering
    \includegraphics[width=1\linewidth]{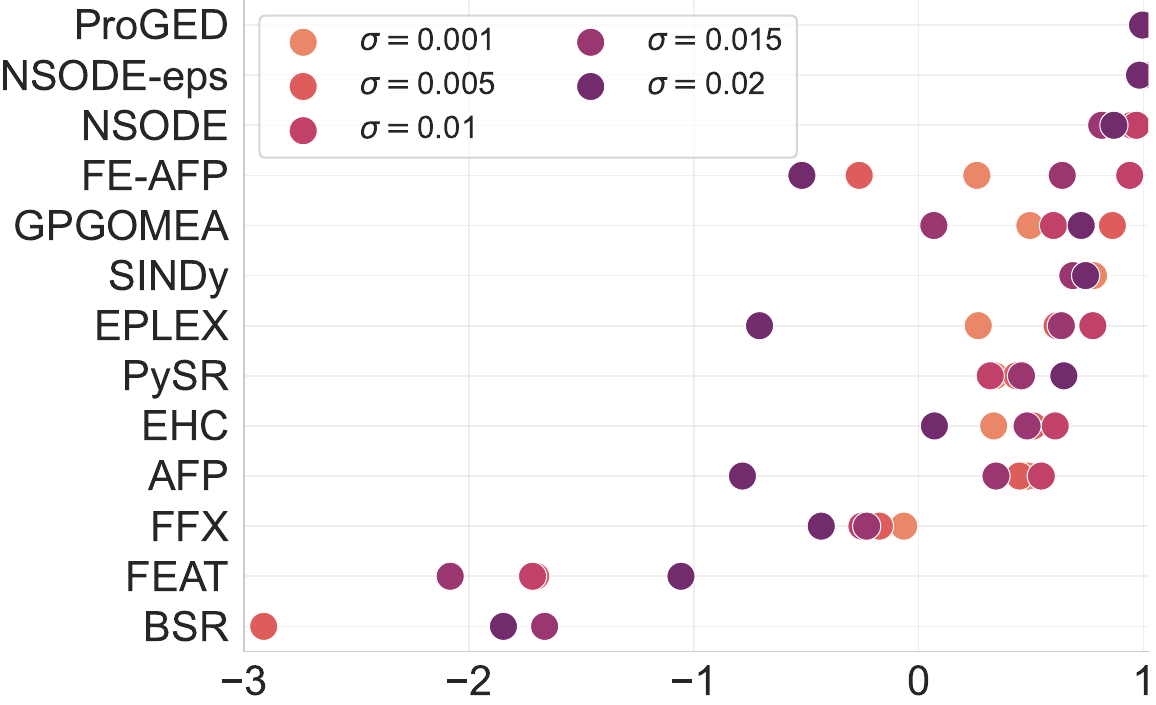}
    \caption{R$^2$, n=256}
\end{subfigure}
\begin{subfigure}{.3\textwidth}
    \centering
    \includegraphics[width=1\linewidth]{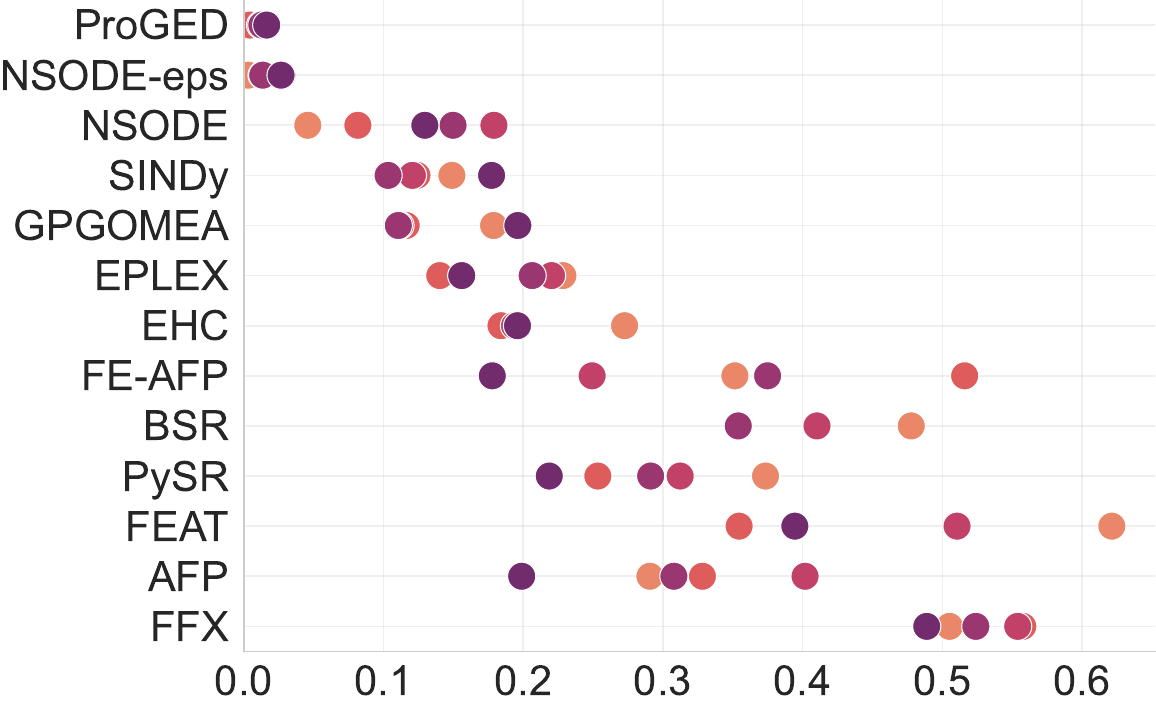}
    \caption{$L_1$, n=128}
\end{subfigure}%
\begin{subfigure}{.3\textwidth}
    \centering
    \includegraphics[width=1\linewidth]{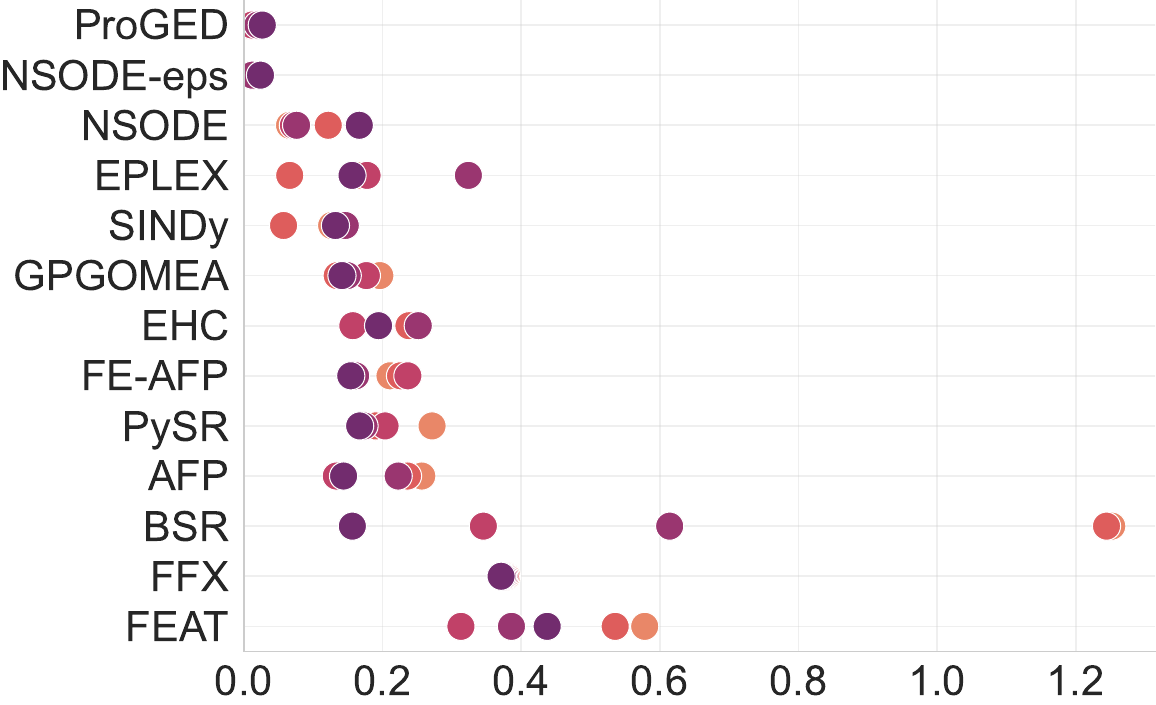}
    \caption{$L_1$, n=192}
\end{subfigure}%
\begin{subfigure}{.3\textwidth}
    \centering
    \includegraphics[width=1\linewidth]{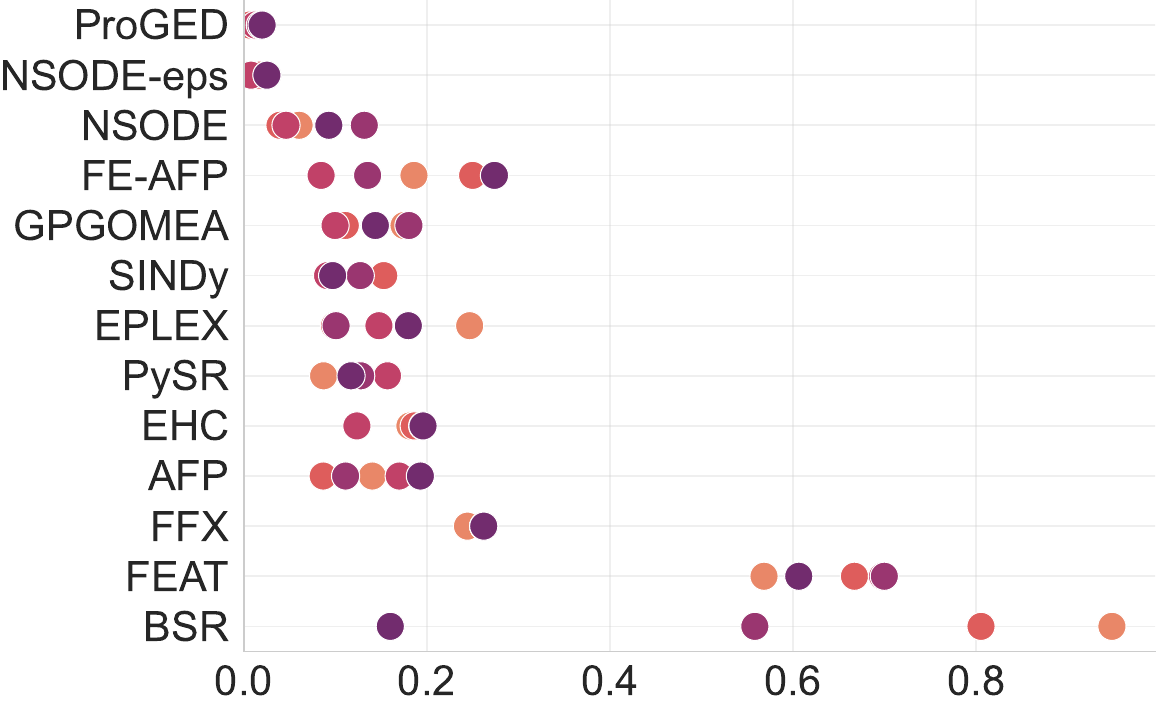}
    \caption{$L_1$, n=256}
\end{subfigure}
\begin{subfigure}{.3\textwidth}
    \centering
    \includegraphics[width=1\linewidth]{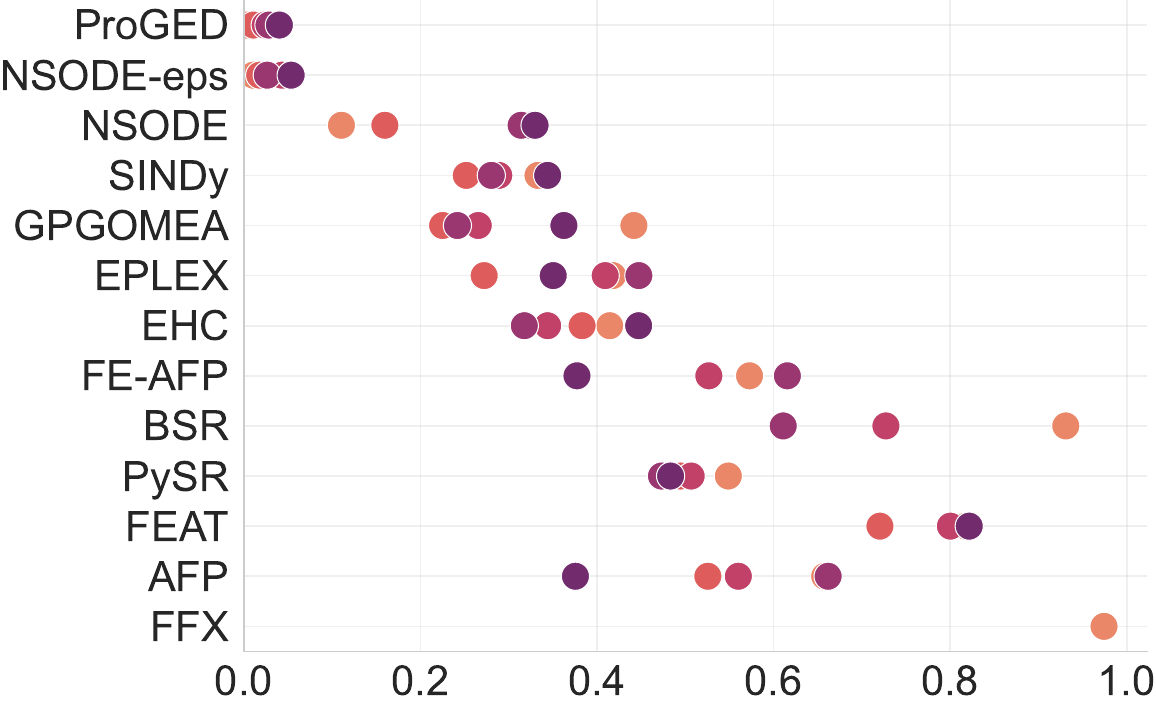}
    \caption{$L_{\infty}$, n=128}
\end{subfigure}%
\begin{subfigure}{.3\textwidth}
    \centering
    \includegraphics[width=1\linewidth]{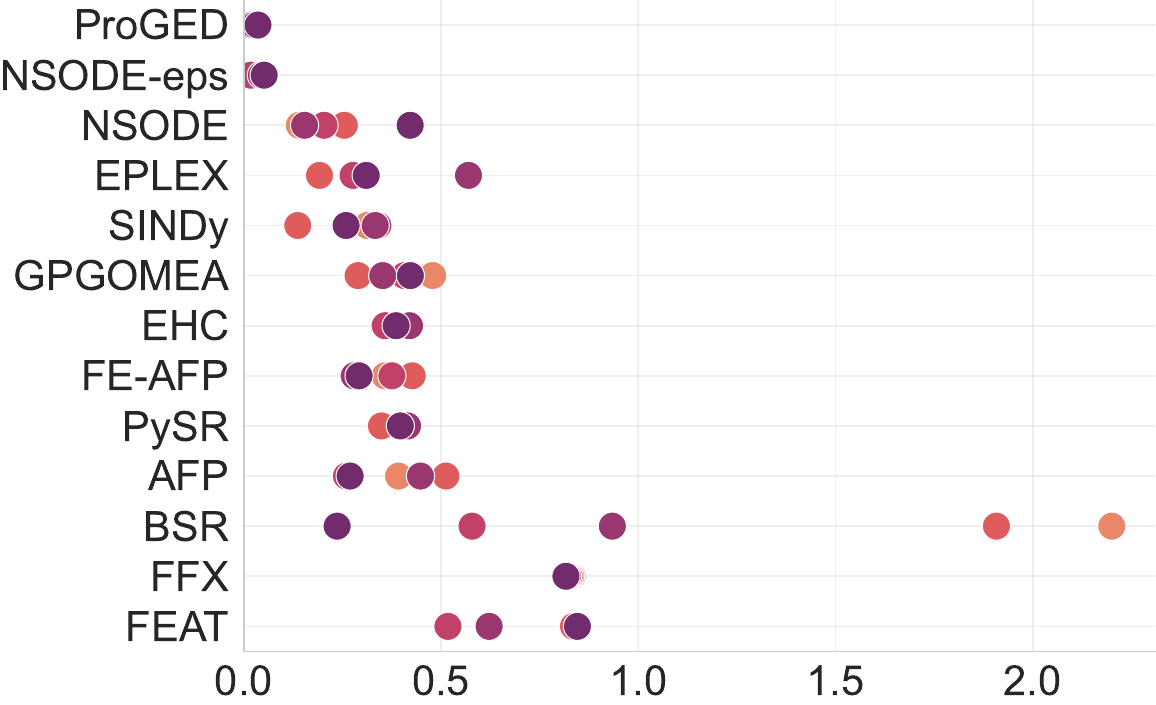}
    \caption{$L_{\infty}$, n=192}
\end{subfigure}%
\begin{subfigure}{.3\textwidth}
    \centering
    \includegraphics[width=1\linewidth]{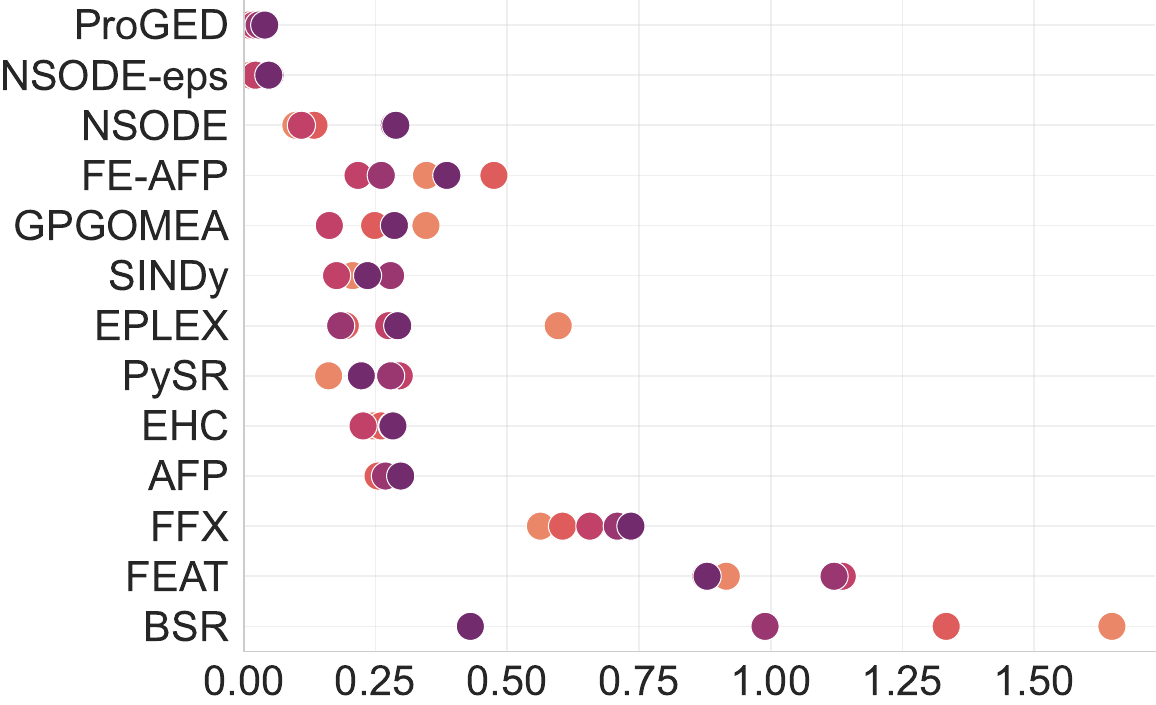}
    \caption{$L_{\infty}$, n=256}
\end{subfigure}
\begin{subfigure}{.3\textwidth}
    \centering
    \includegraphics[width=1\linewidth]{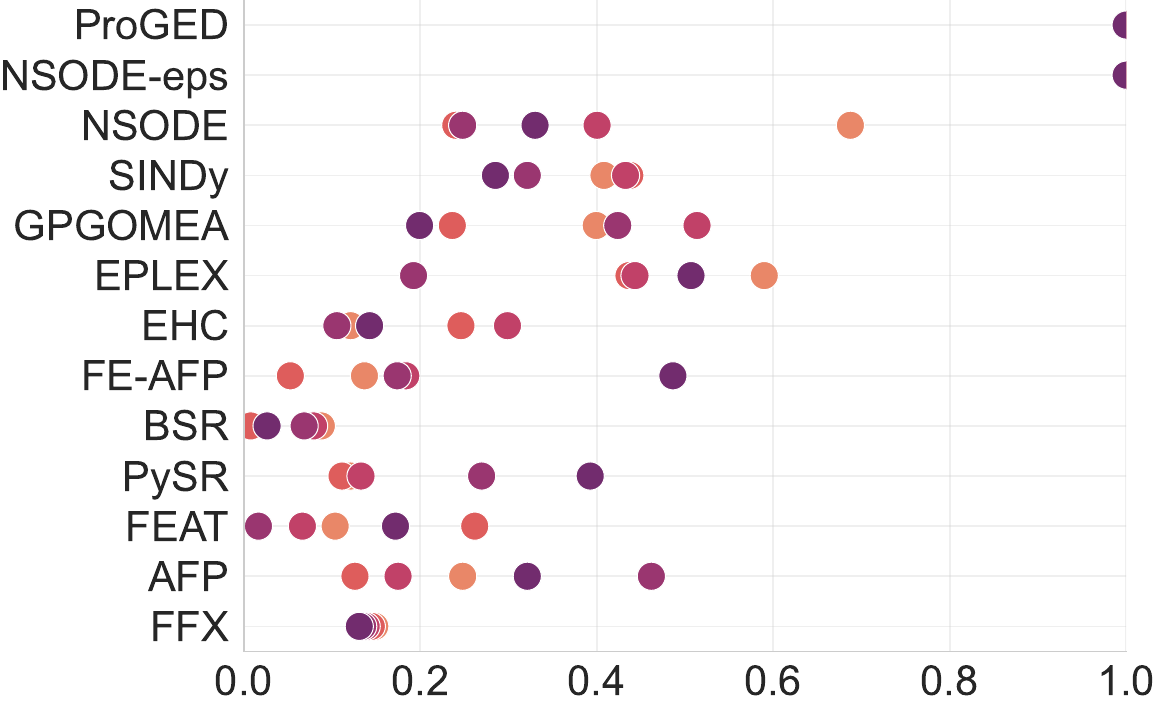}
    \caption{$\nicefrac{\texttt{isclose}}{n}$, n=128}
\end{subfigure}%
\begin{subfigure}{.3\textwidth}
    \centering
    \includegraphics[width=1\linewidth]{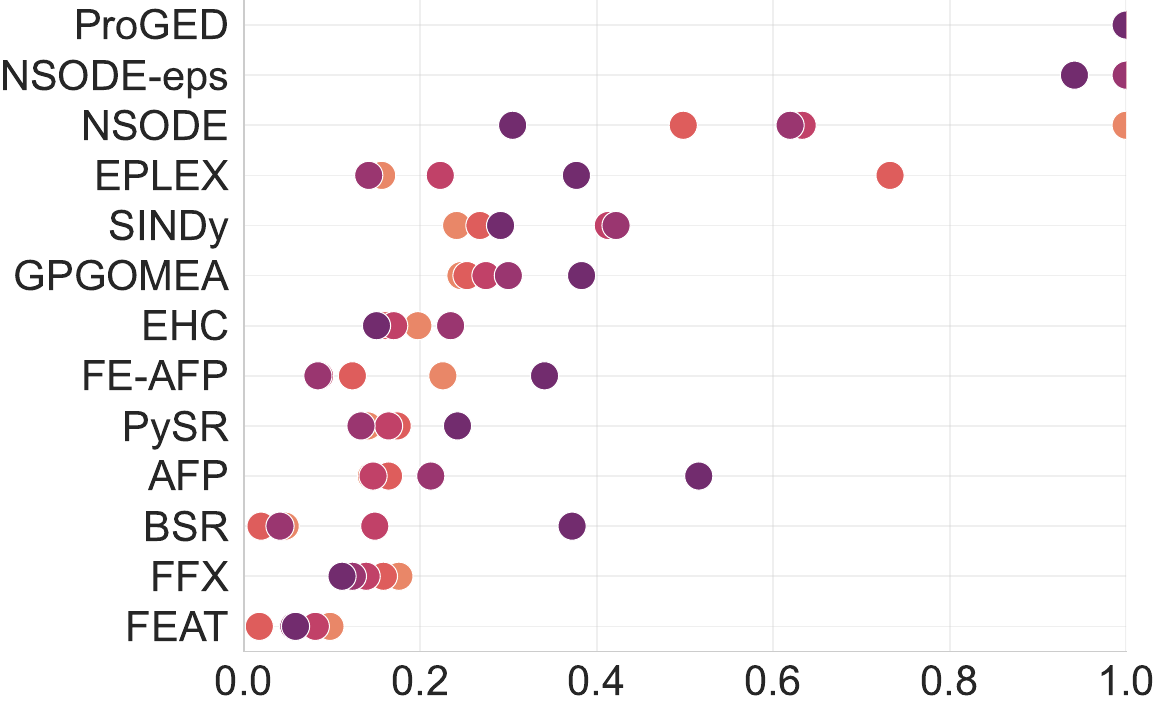}
    \caption{$\nicefrac{\texttt{isclose}}{n}$, n=192}
\end{subfigure}%
\begin{subfigure}{.3\textwidth}
    \centering
    \includegraphics[width=1\linewidth]{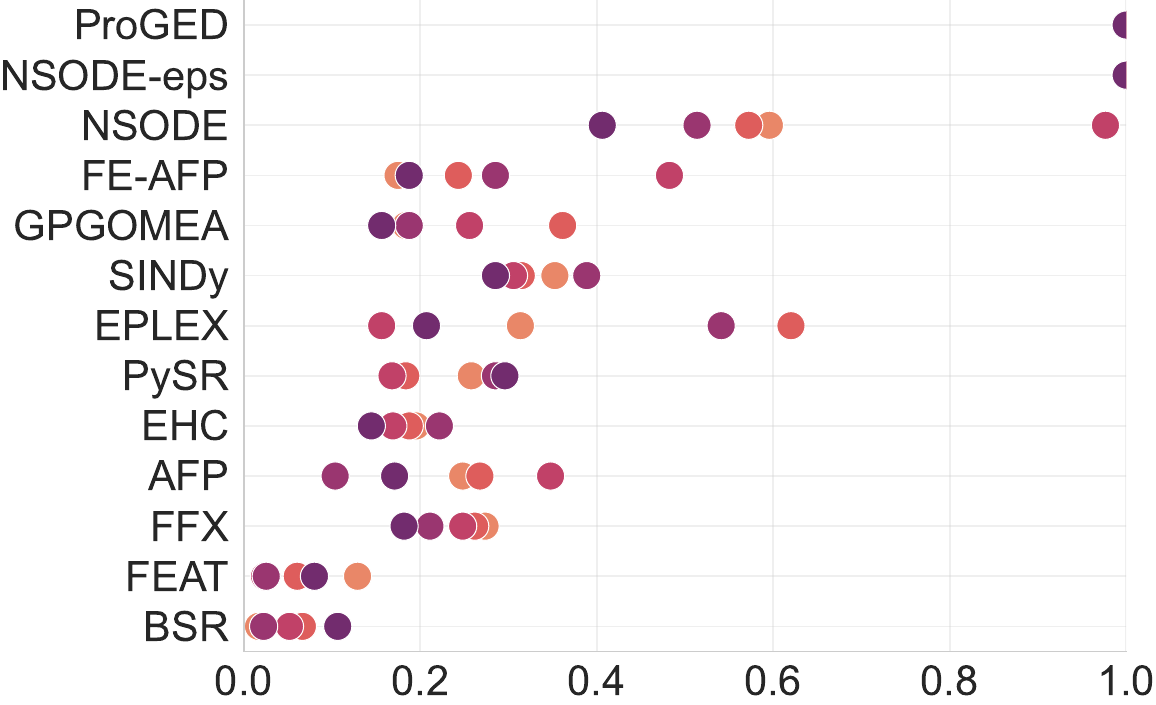}
    \caption{$\nicefrac{\texttt{isclose}}{n}$, n=256}
\end{subfigure}
\begin{subfigure}{.3\textwidth}
    \centering
    \includegraphics[width=1\linewidth]{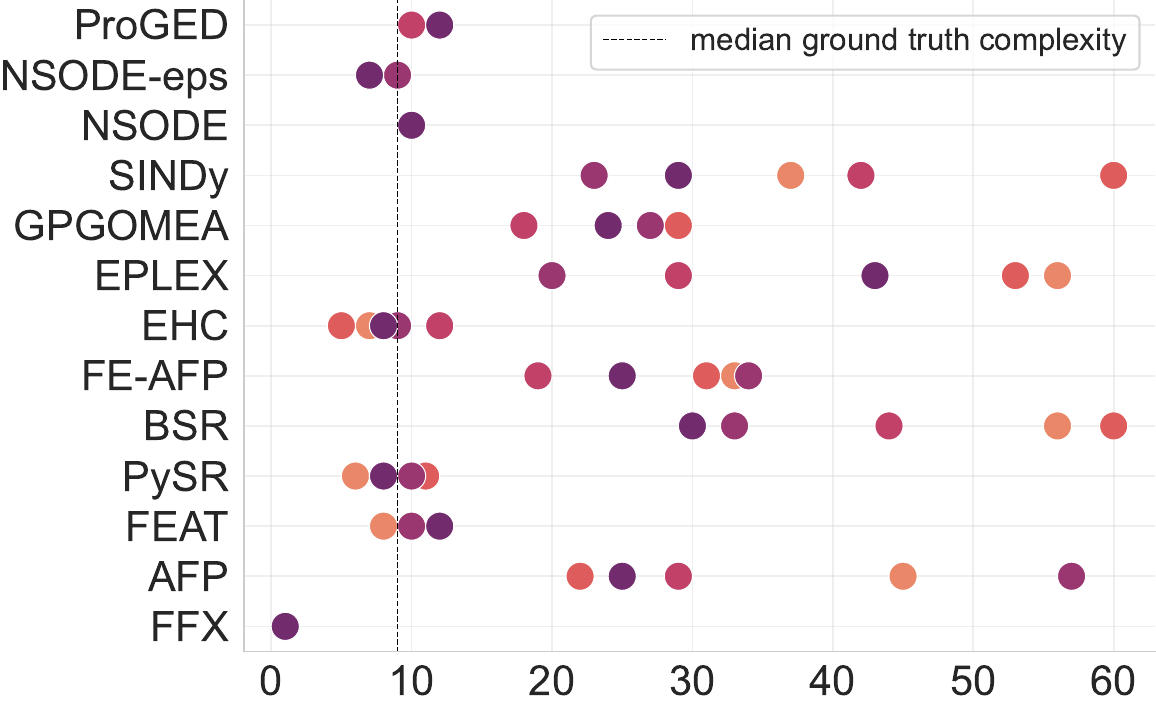}
    \caption{complexity, n=128}
\end{subfigure}%
\begin{subfigure}{.3\textwidth}
    \centering
    \includegraphics[width=1\linewidth]{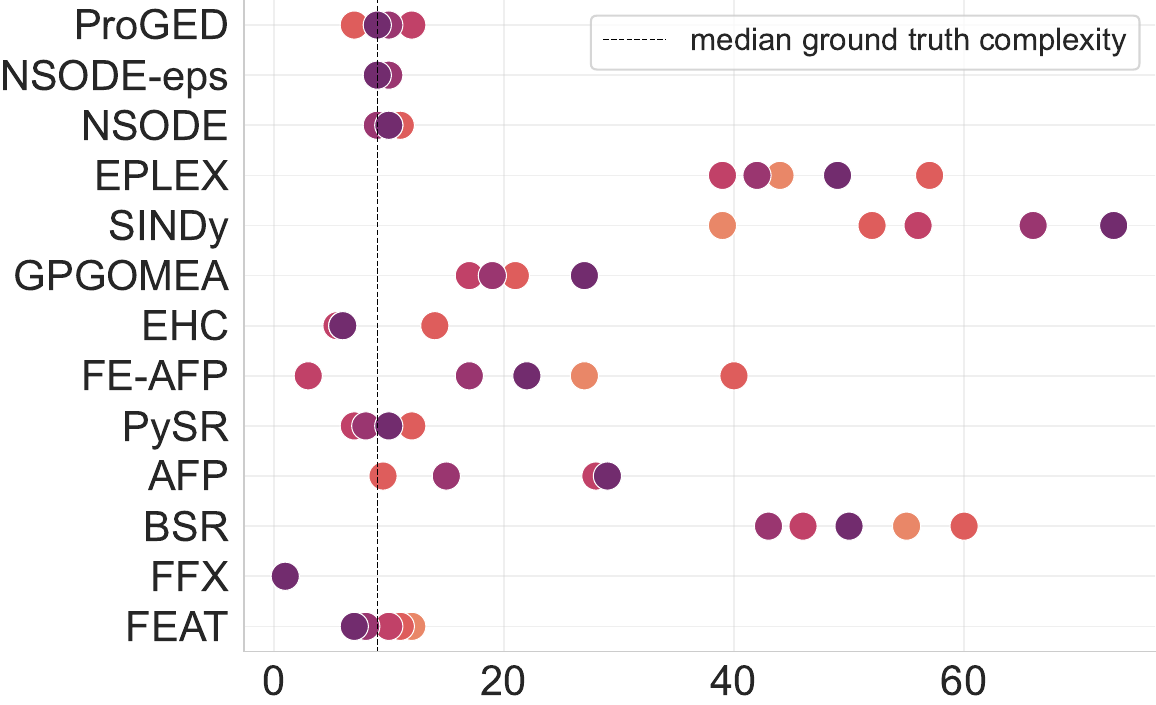}
    \caption{complexity, n=192}
\end{subfigure}%
\begin{subfigure}{.3\textwidth}
    \centering
    \includegraphics[width=1\linewidth]{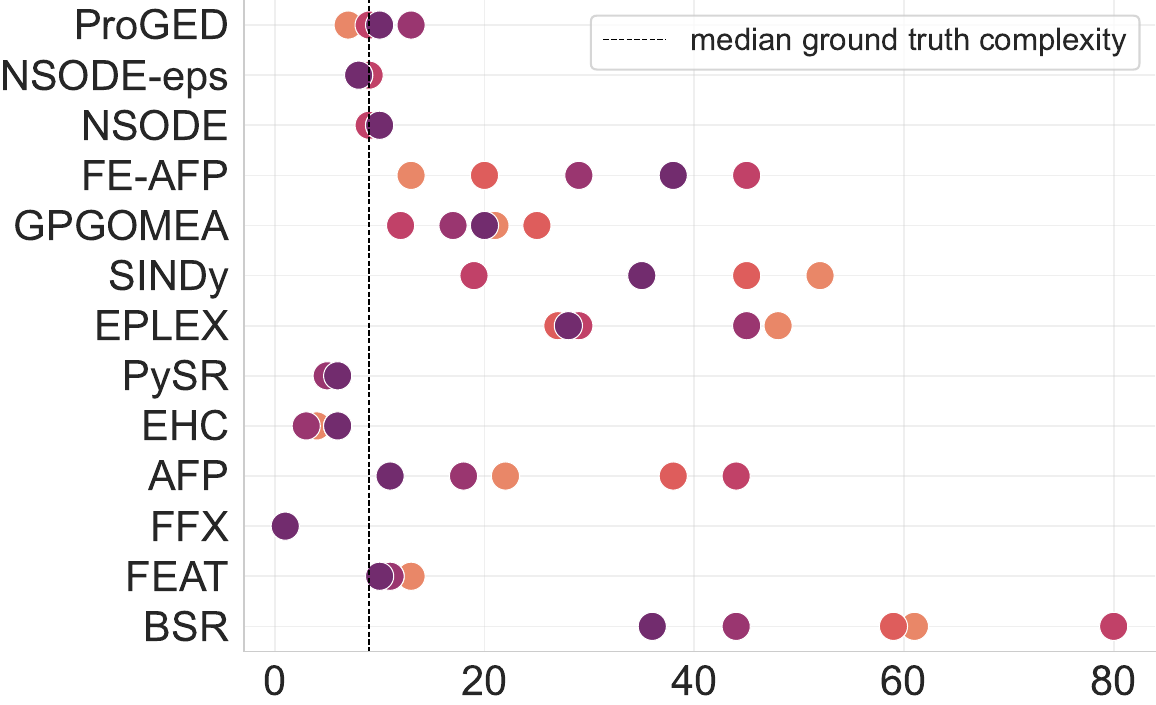}
    \caption{complexity, n=256}
\end{subfigure}
\begin{subfigure}{.3\textwidth}
    \centering
    \includegraphics[width=1\linewidth]{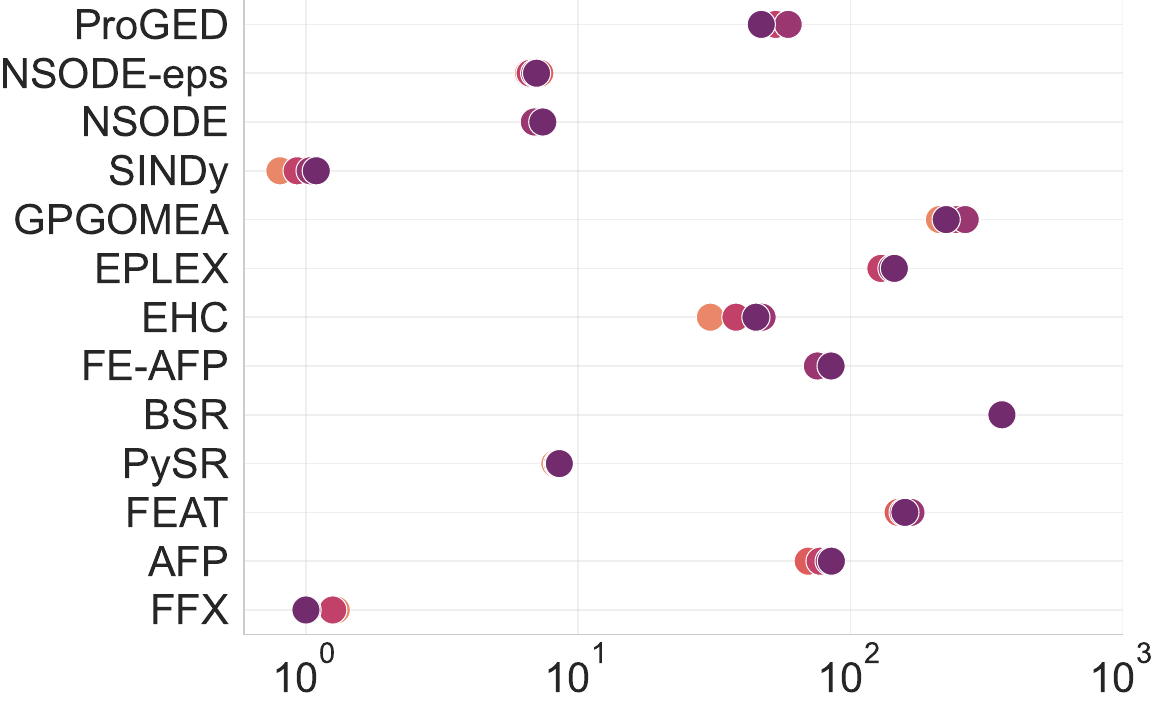}
    \caption{inference time [sec], n=128}
\end{subfigure}%
\begin{subfigure}{.3\textwidth}
    \centering
    \includegraphics[width=1\linewidth]{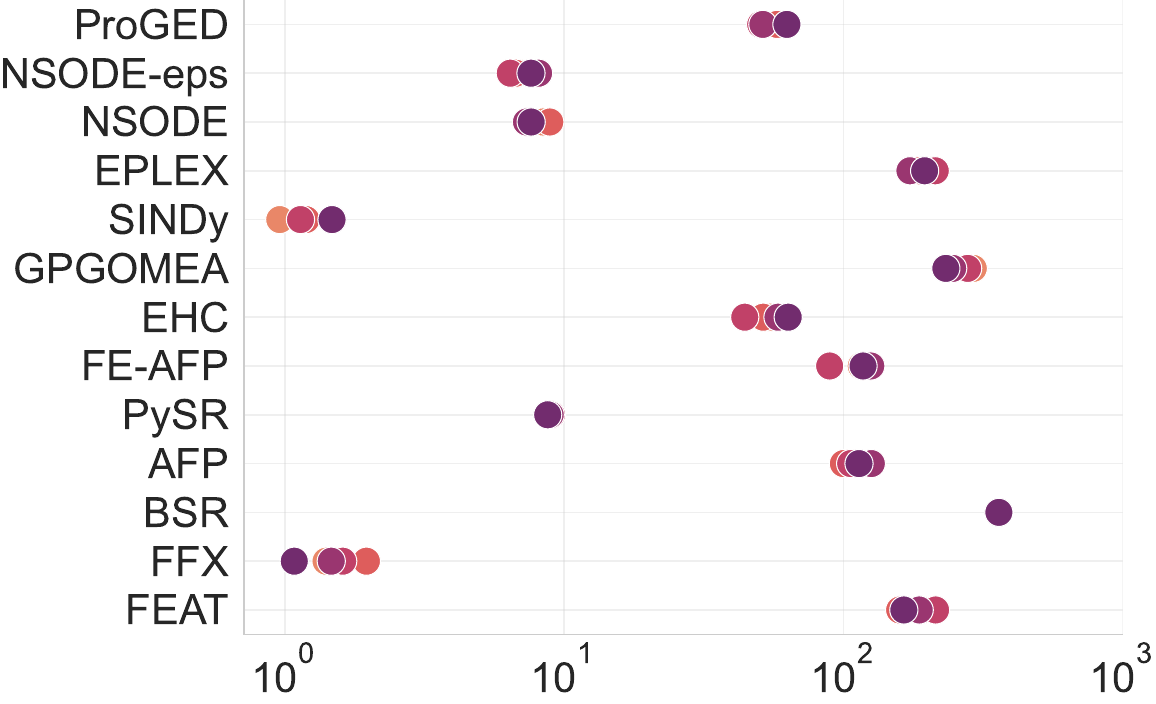}
    \caption{inference time [sec], n=192}
\end{subfigure}%
\begin{subfigure}{.3\textwidth}
    \centering
    \includegraphics[width=1\linewidth]{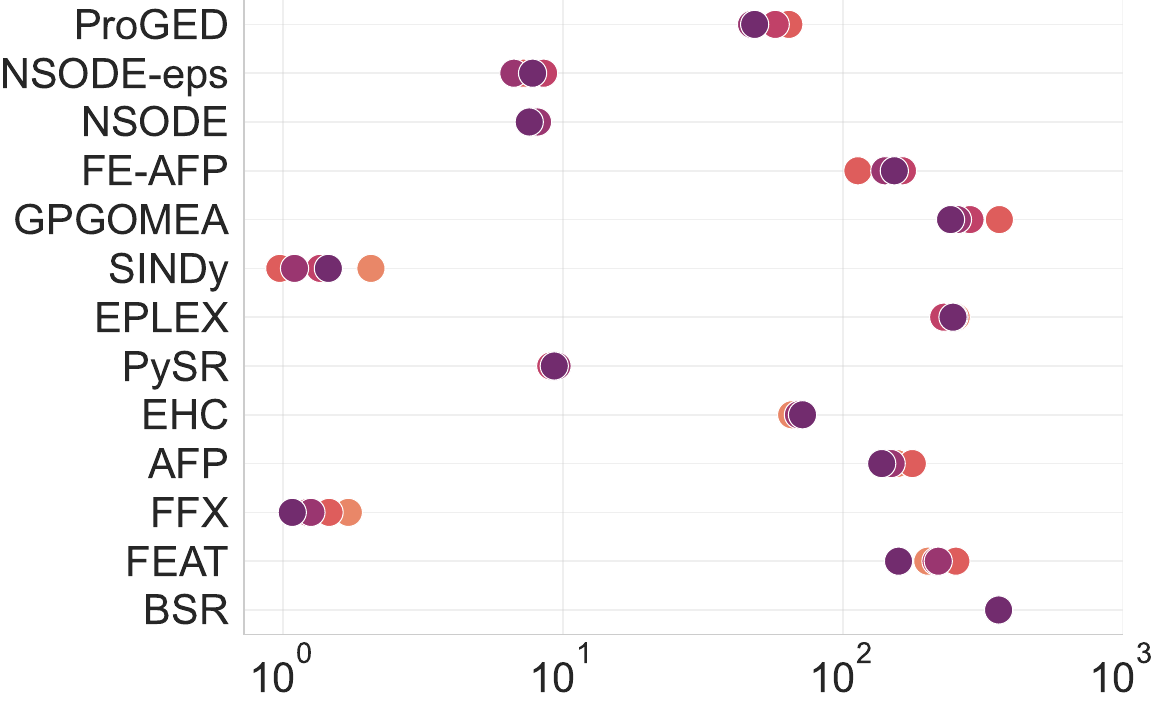}
    \caption{inference time [sec], n=256}
\end{subfigure}%
\caption{Interpolation. Median scores on \textbf{Textbook} for $n$ irregularly sampled time points across different noise levels $\sigma$.}
\label{app:results_textbook_intra}
\end{figure*}

\begin{figure*}[h!]
\centering
\begin{subfigure}{.3\textwidth}
    \centering
    \includegraphics[width=1\linewidth]{figs/performance/163_r2_128.pdf}
    \caption{R$^2$, n=128}
\end{subfigure}%
\begin{subfigure}{.3\textwidth}
    \centering
    \includegraphics[width=1\linewidth]{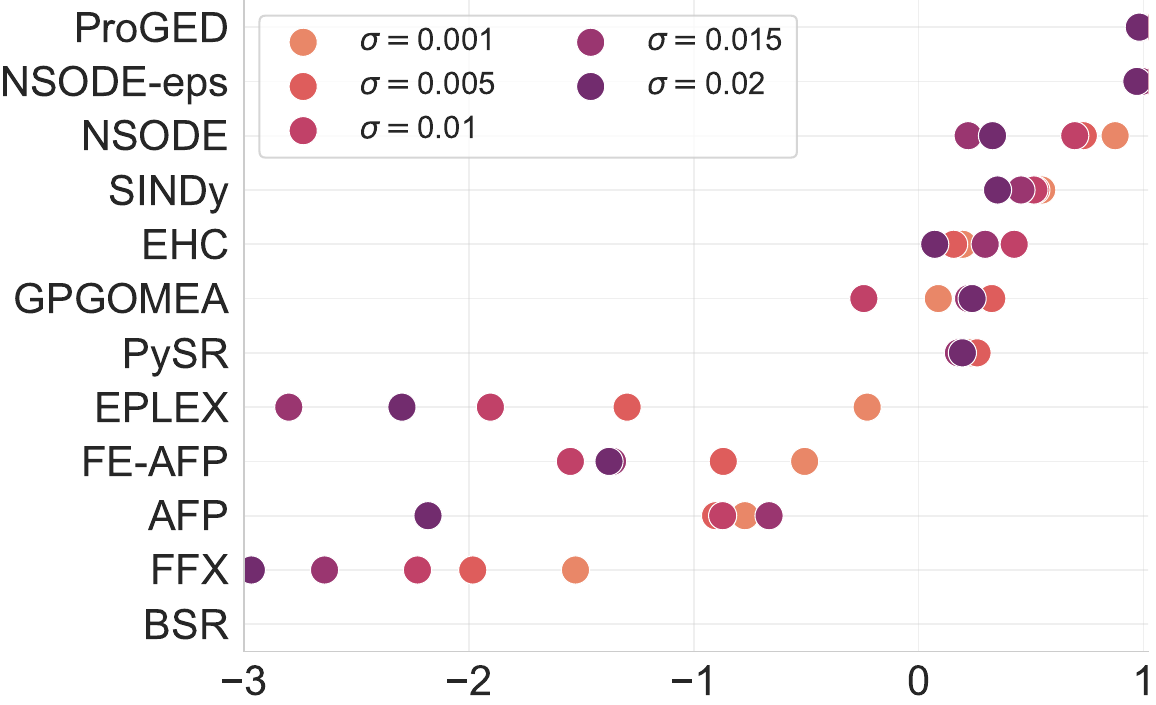}
    \caption{R$^2$, n=192}
\end{subfigure}%
\begin{subfigure}{.3\textwidth}
    \centering
    \includegraphics[width=1\linewidth]{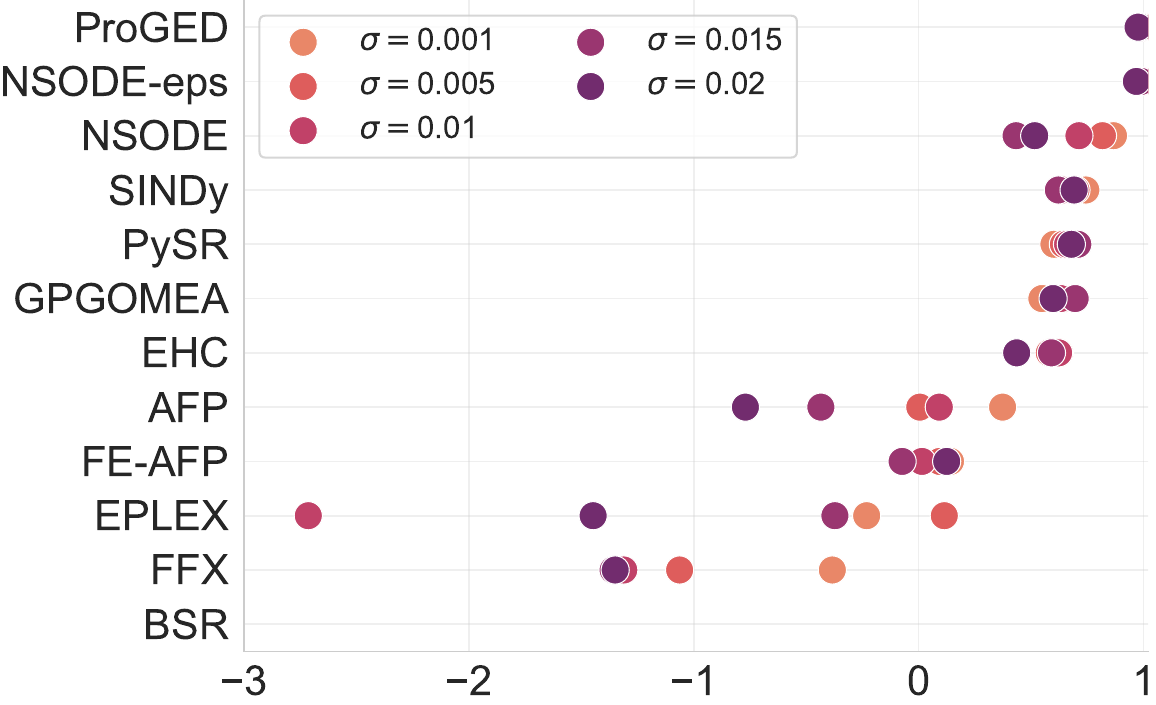}
    \caption{R$^2$, n=256}
\end{subfigure}
\begin{subfigure}{.3\textwidth}
    \centering
    \includegraphics[width=1\linewidth]{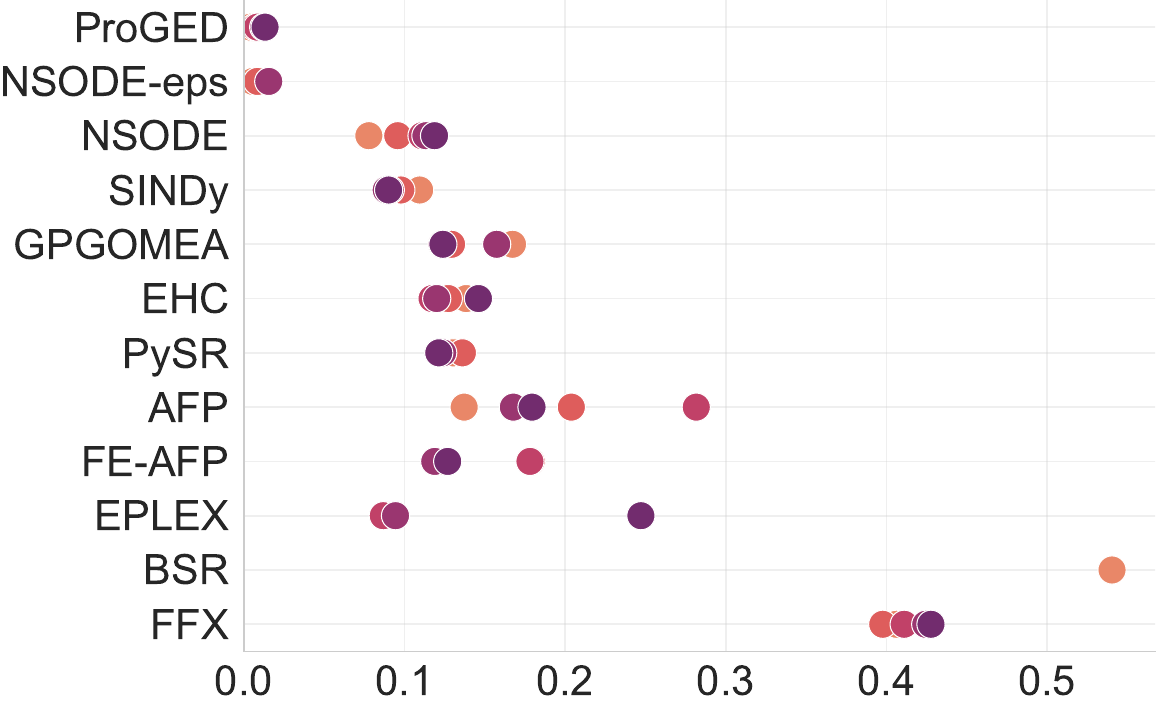}
    \caption{$L_1$, n=128}
\end{subfigure}%
\begin{subfigure}{.3\textwidth}
    \centering
    \includegraphics[width=1\linewidth]{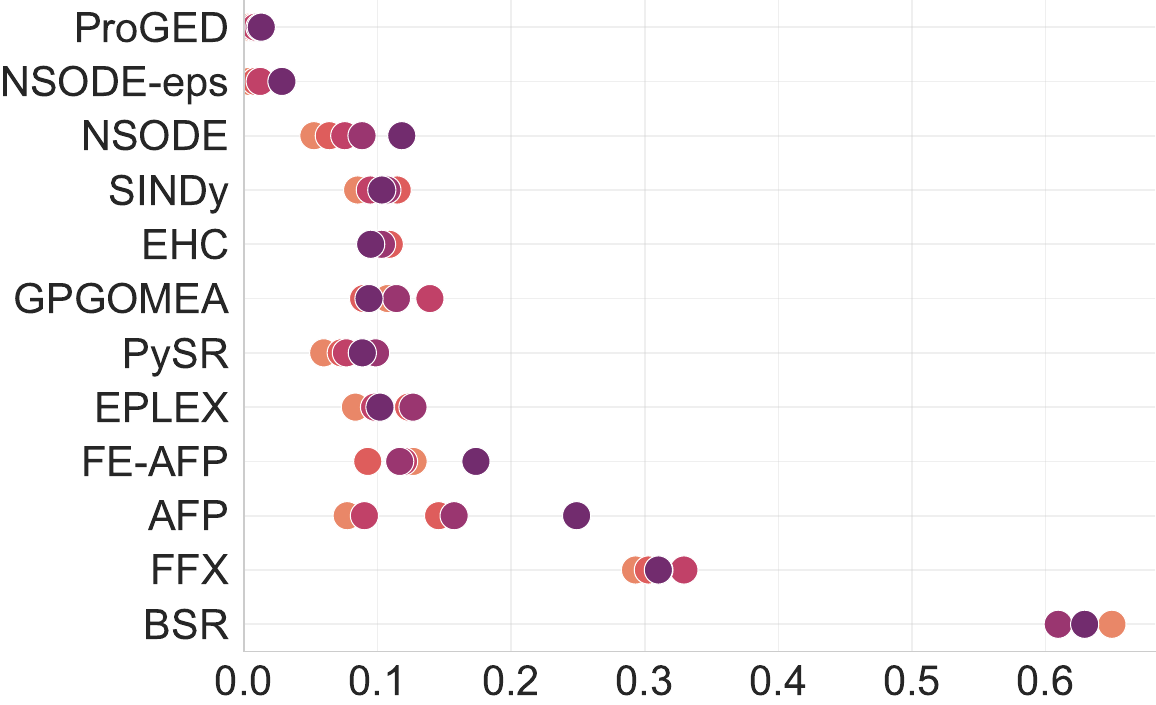}
    \caption{$L_1$, n=192}
\end{subfigure}%
\begin{subfigure}{.3\textwidth}
    \centering
    \includegraphics[width=1\linewidth]{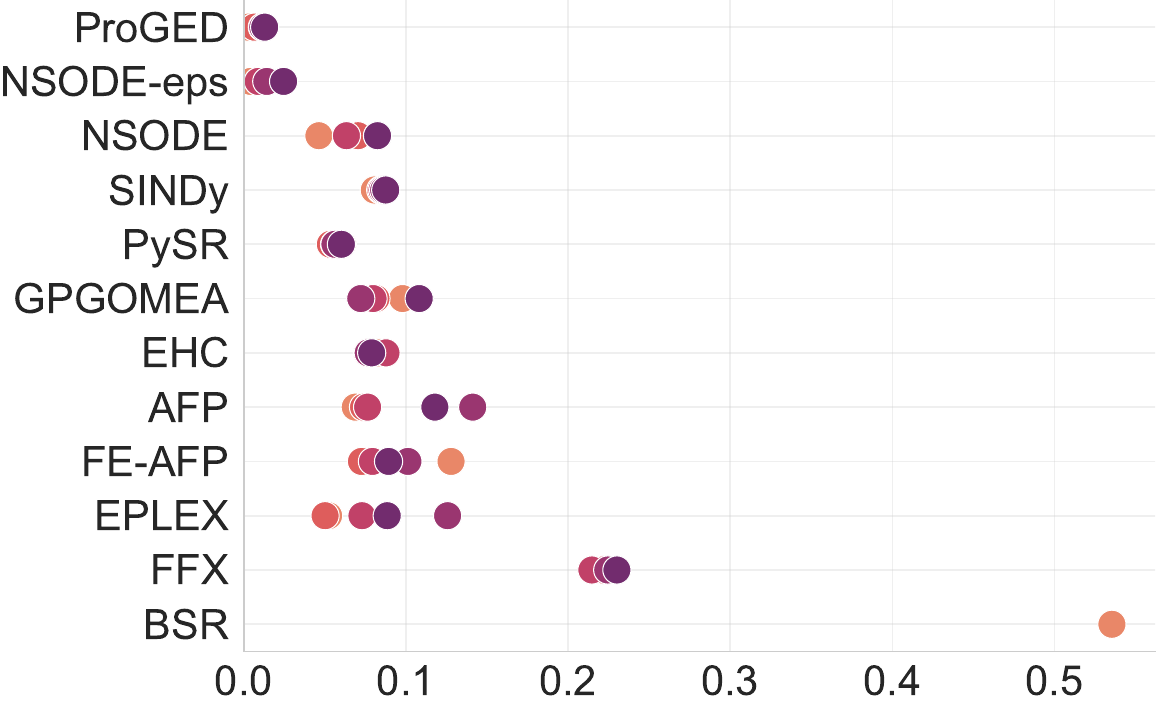}
    \caption{$L_1$, n=256}
\end{subfigure}
\begin{subfigure}{.3\textwidth}
    \centering
    \includegraphics[width=1\linewidth]{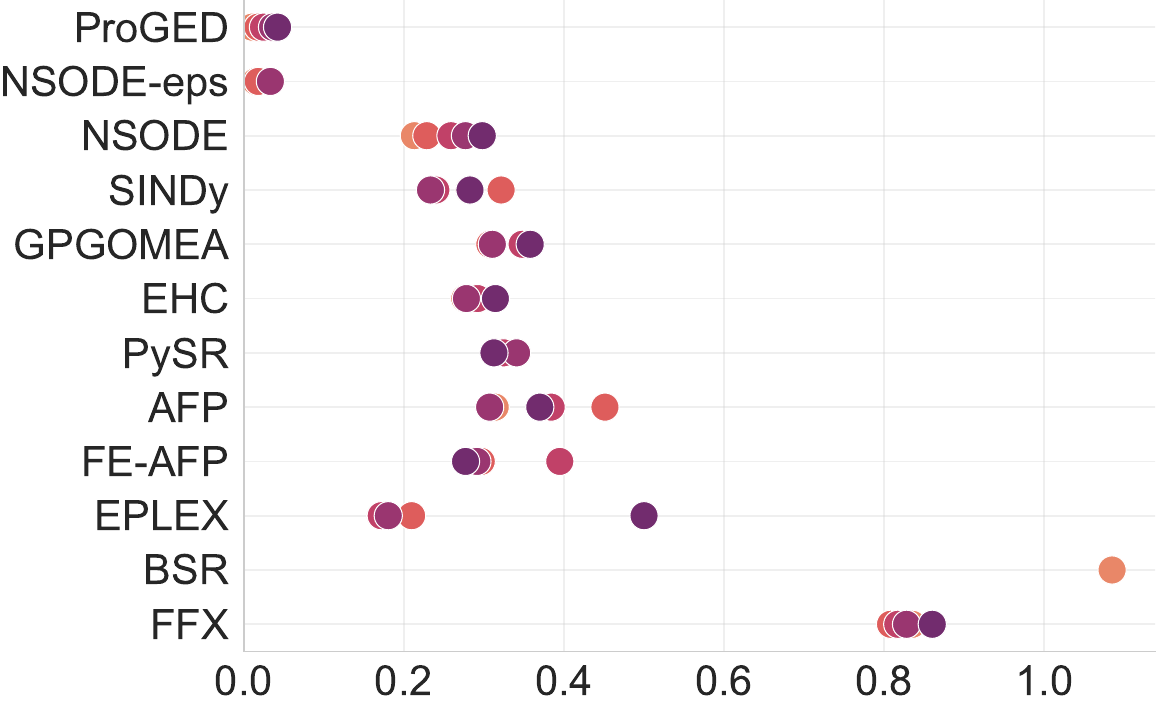}
    \caption{$L_{\infty}$, n=128}
\end{subfigure}%
\begin{subfigure}{.3\textwidth}
    \centering
    \includegraphics[width=1\linewidth]{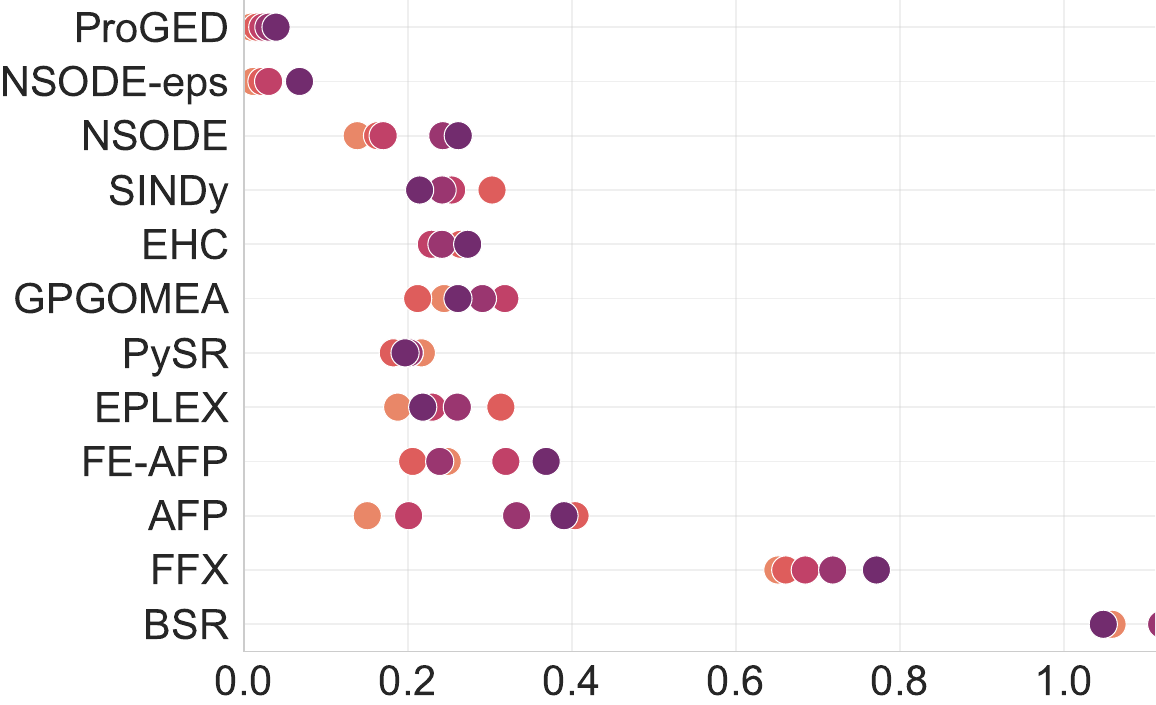}
    \caption{$L_{\infty}$, n=192}
\end{subfigure}%
\begin{subfigure}{.3\textwidth}
    \centering
    \includegraphics[width=1\linewidth]{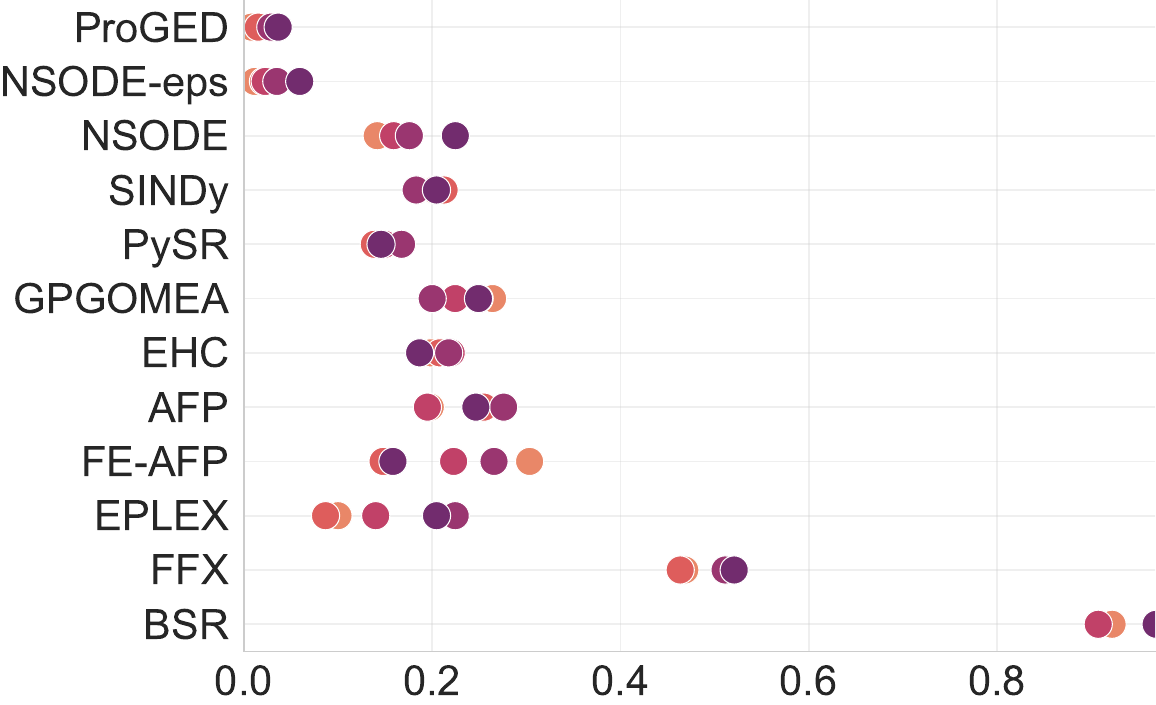}
    \caption{$L_{\infty}$, n=256}
\end{subfigure}
\begin{subfigure}{.3\textwidth}
    \centering
    \includegraphics[width=1\linewidth]{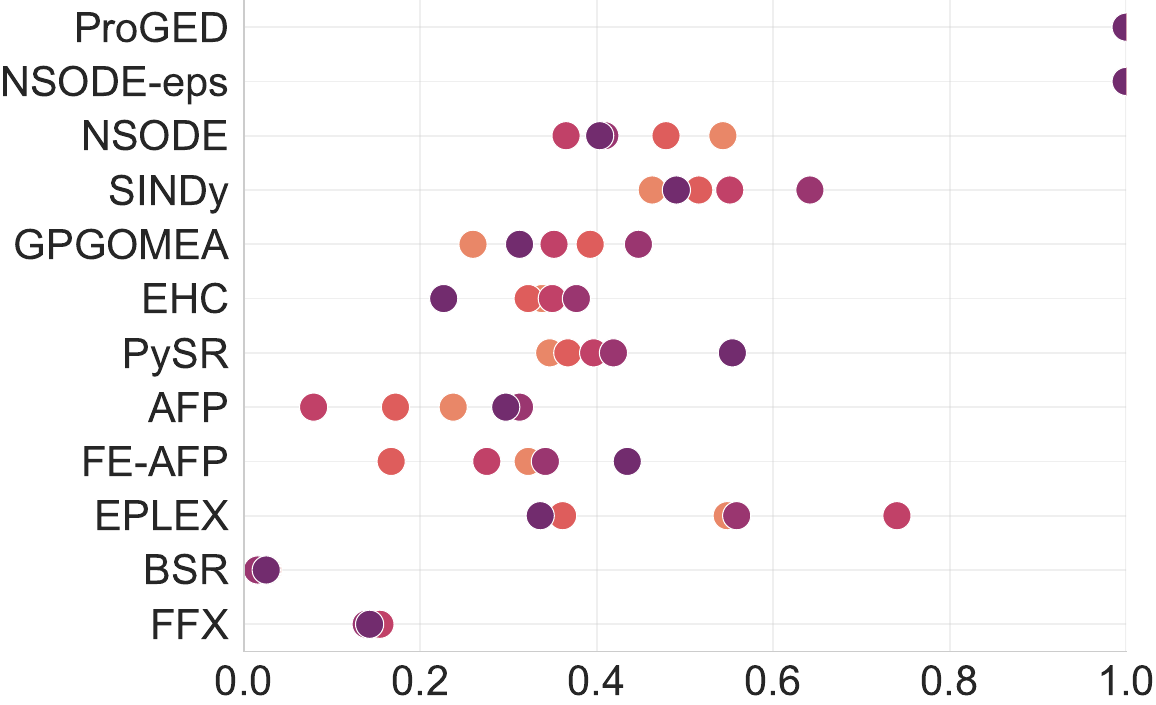}
    \caption{$\nicefrac{\texttt{isclose}}{n}$, n=128}
\end{subfigure}%
\begin{subfigure}{.3\textwidth}
    \centering
    \includegraphics[width=1\linewidth]{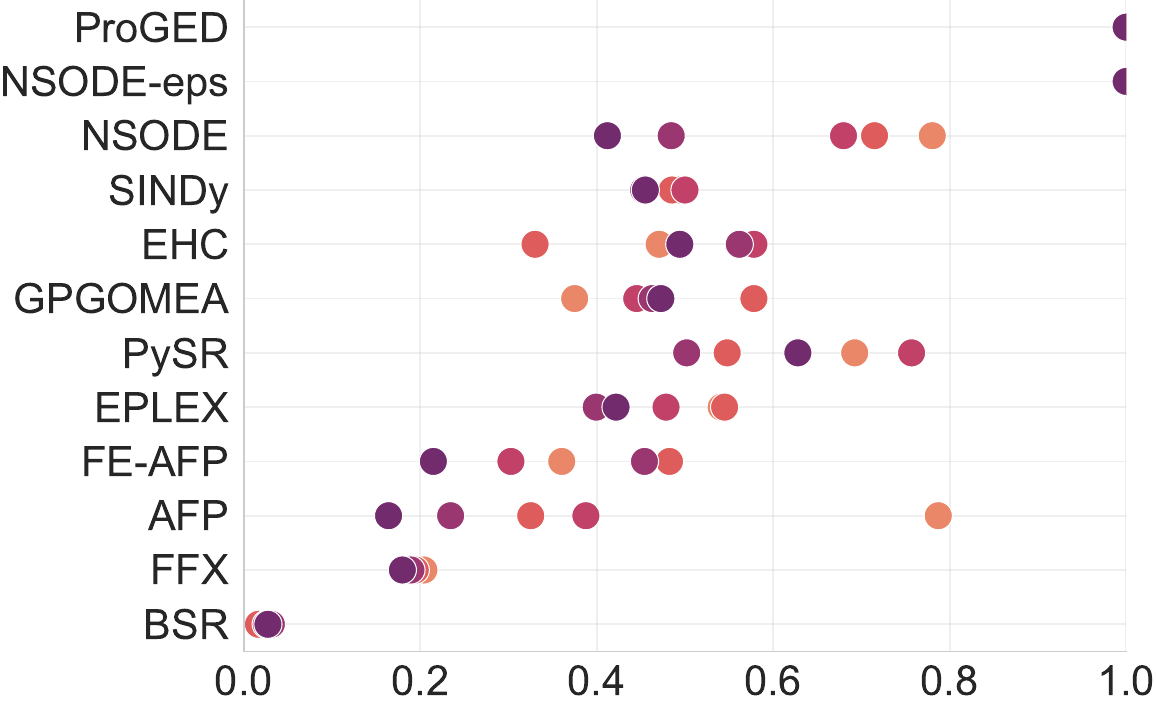}
    \caption{$\nicefrac{\texttt{isclose}}{n}$, n=192}
\end{subfigure}%
\begin{subfigure}{.3\textwidth}
    \centering
    \includegraphics[width=1\linewidth]{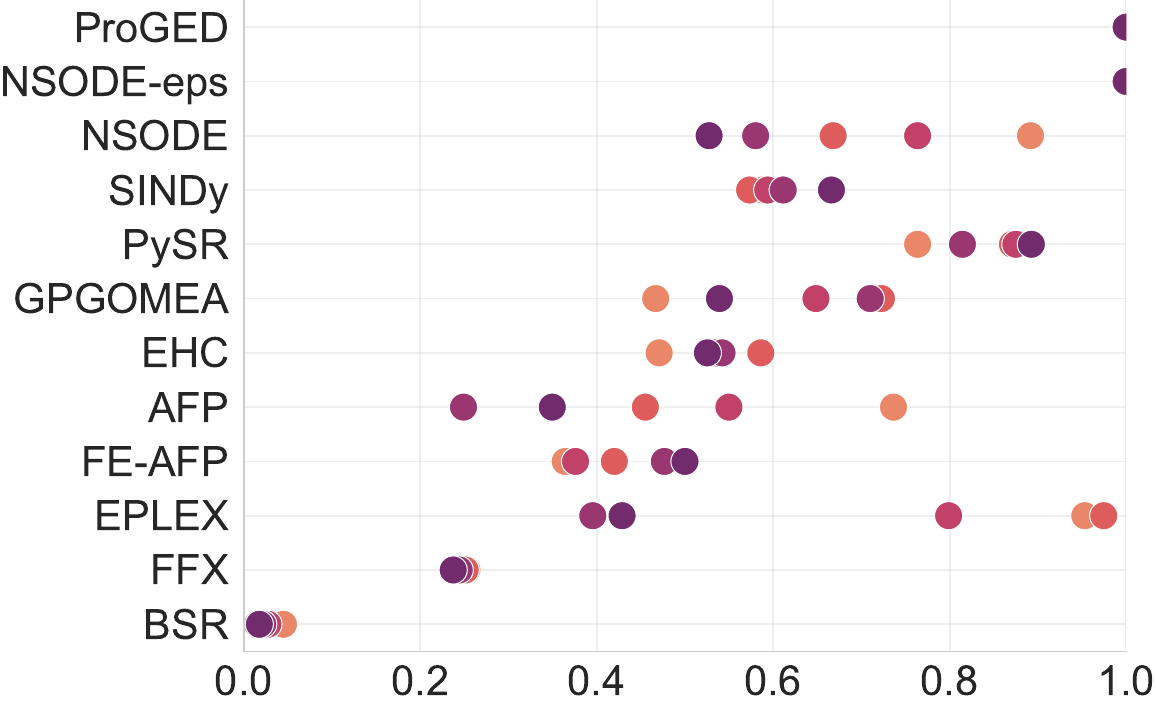}
    \caption{$\nicefrac{\texttt{isclose}}{n}$, n=256}
\end{subfigure}
\begin{subfigure}{.3\textwidth}
    \centering
    \includegraphics[width=1\linewidth]{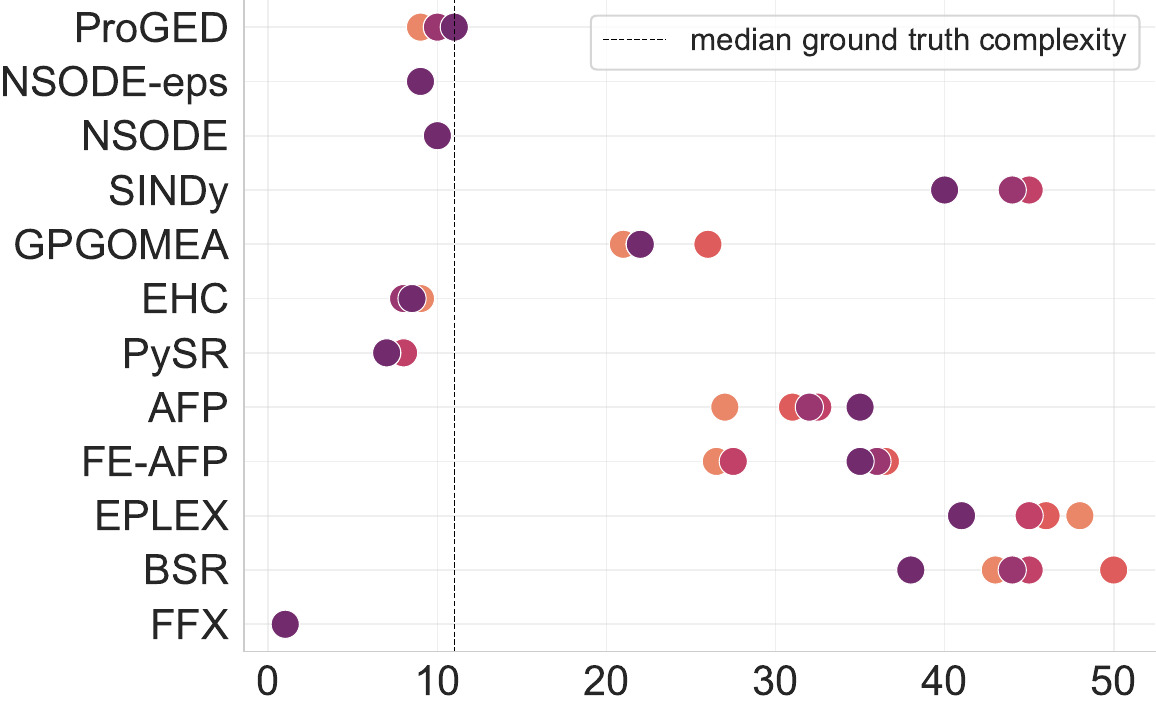}
    \caption{complexity, n=128}
\end{subfigure}%
\begin{subfigure}{.3\textwidth}
    \centering
    \includegraphics[width=1\linewidth]{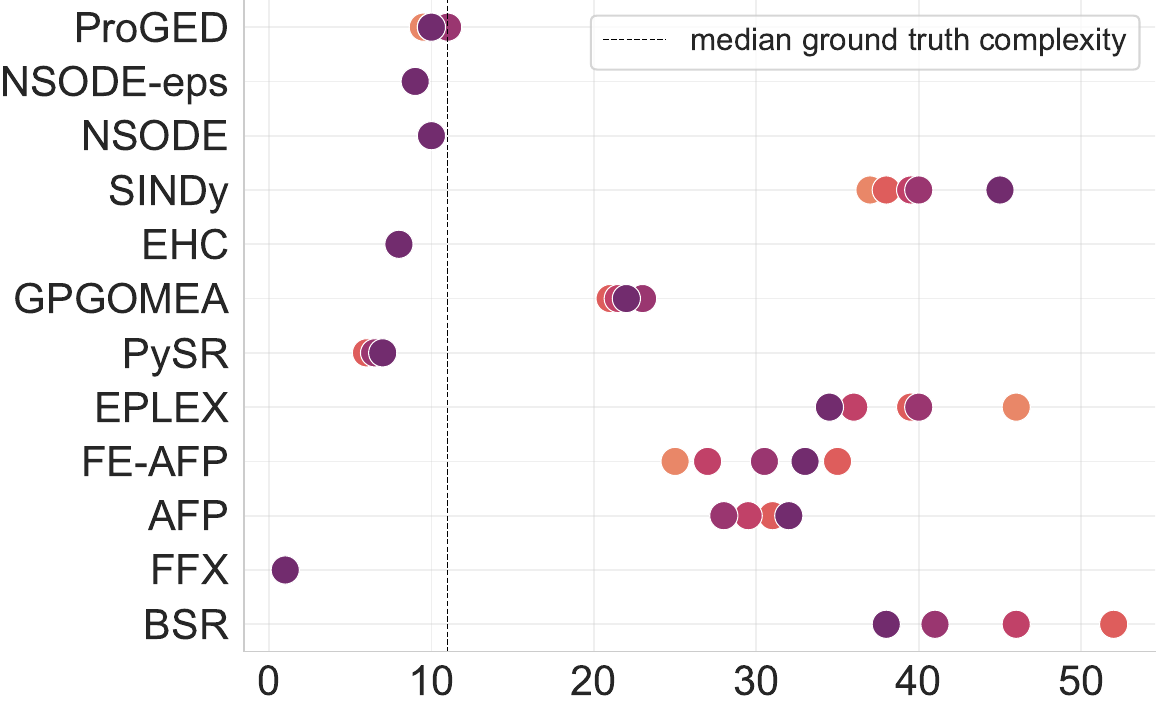}
    \caption{complexity, n=192}
\end{subfigure}%
\begin{subfigure}{.3\textwidth}
    \centering
    \includegraphics[width=1\linewidth]{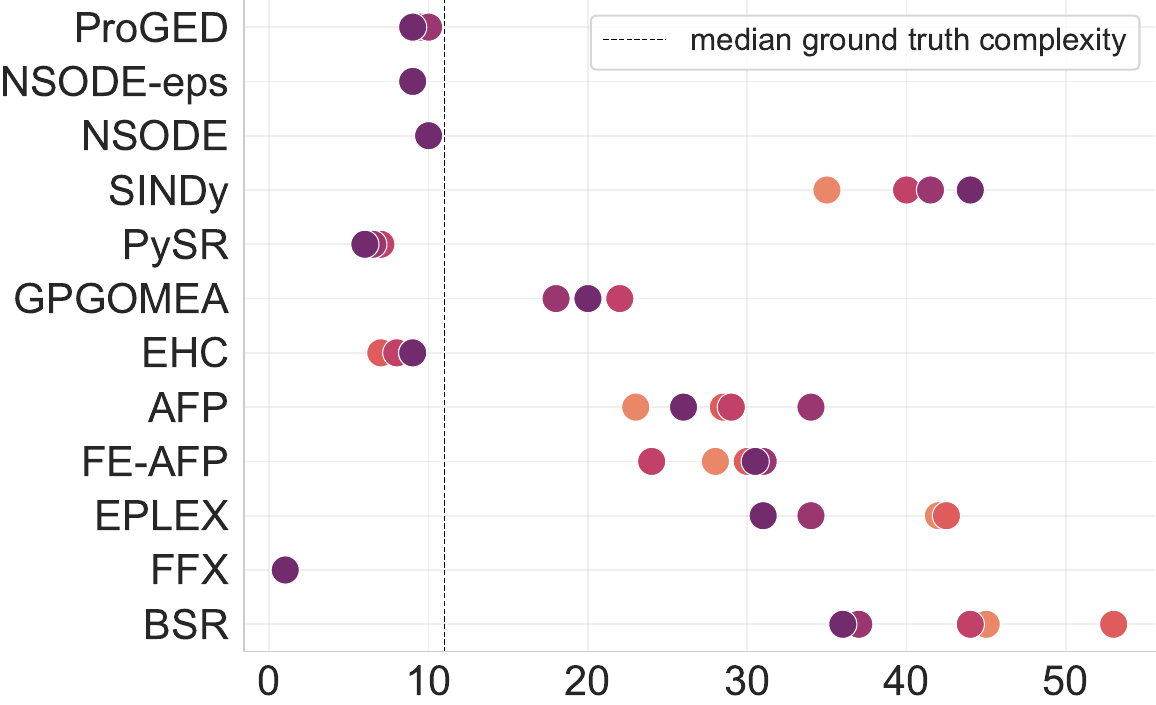}
    \caption{complexity, n=256}
\end{subfigure}
\begin{subfigure}{.3\textwidth}
    \centering
    \includegraphics[width=1\linewidth]{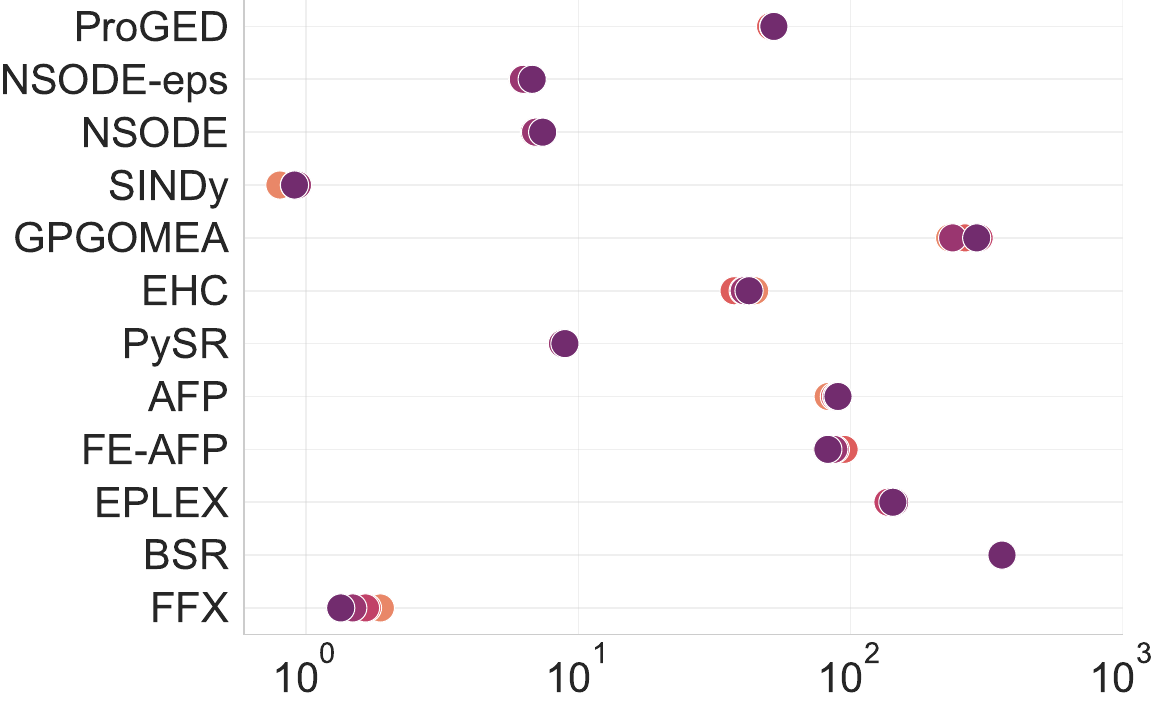}
    \caption{inference time [sec], n=128}
\end{subfigure}%
\begin{subfigure}{.3\textwidth}
    \centering
    \includegraphics[width=1\linewidth]{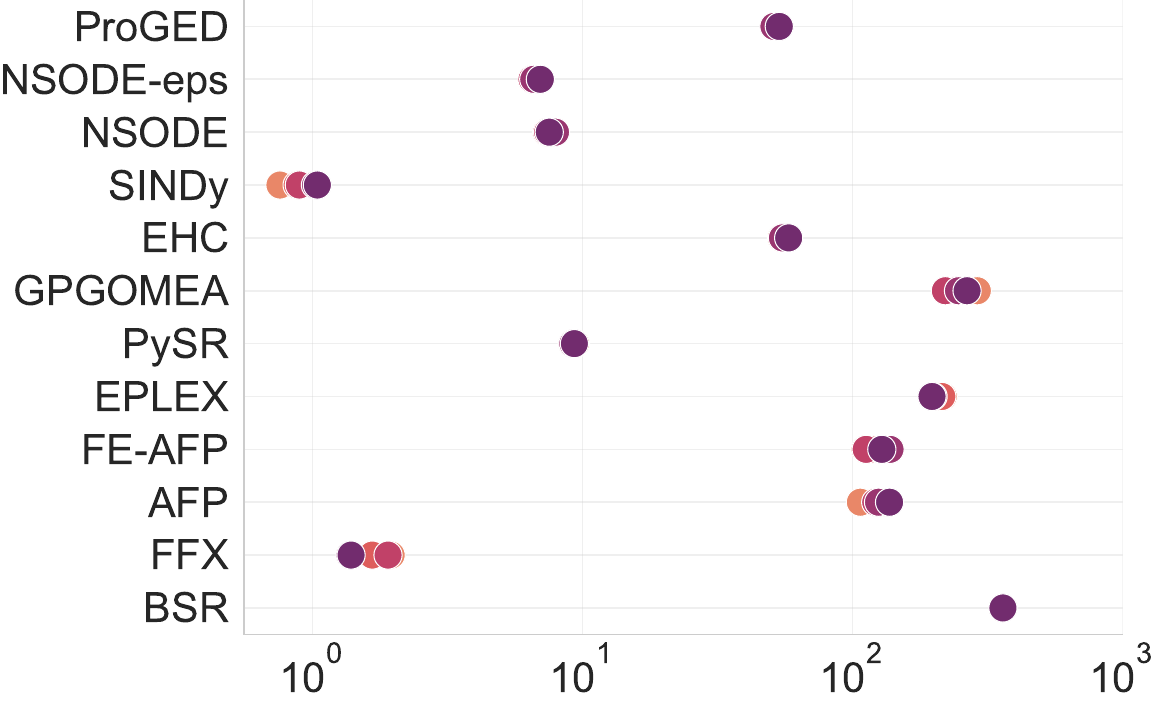}
    \caption{inference time [sec], n=192}
\end{subfigure}%
\begin{subfigure}{.3\textwidth}
    \centering
    \includegraphics[width=1\linewidth]{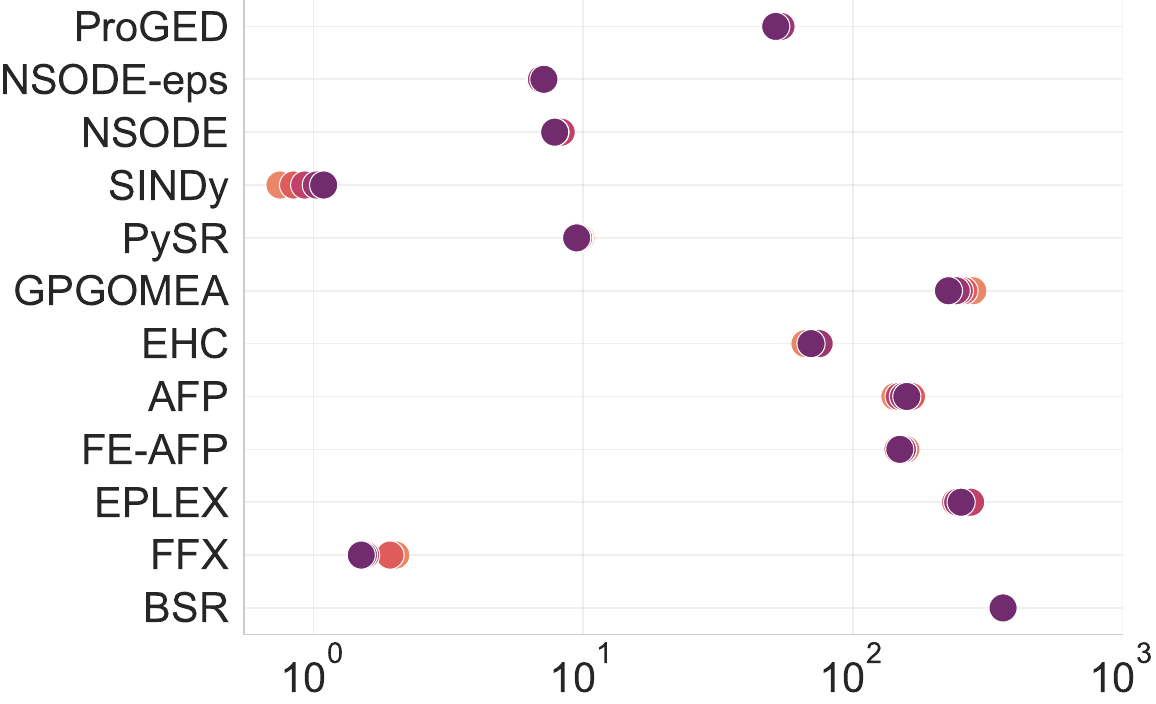}
    \caption{inference time [sec], n=256}
\end{subfigure}%
\caption{Interpolation. Median scores on \textbf{Large} for $n$ irregularly sampled time points across different noise levels $\sigma$.}
\label{app:results_large_intra}
\end{figure*}

Across all datasets, the direct comparison of results with $n=128$ time-points vs $n=192$ time-points vs $n=256$ time-points reveals a gradual performance degradation of models relying on finite difference approximation with increasing sparsity of the observed data both in terms of R$^2$ (top rows in \cref{app:results_classic_intra,app:results_textbook_intra,app:results_large_intra}). NSODE-eps and ProGED on the other hand perform consistently well across these settings. 


\clearpage
\subsection{Extrapolation results}
Performance evaluation on the extrapolation interval $[\maxtime, T_{\mathrm{extra}}]$.

\begin{figure*}[b!]
\centering
\begin{subfigure}{.33\textwidth}
    \centering
    \includegraphics[width=1\linewidth]{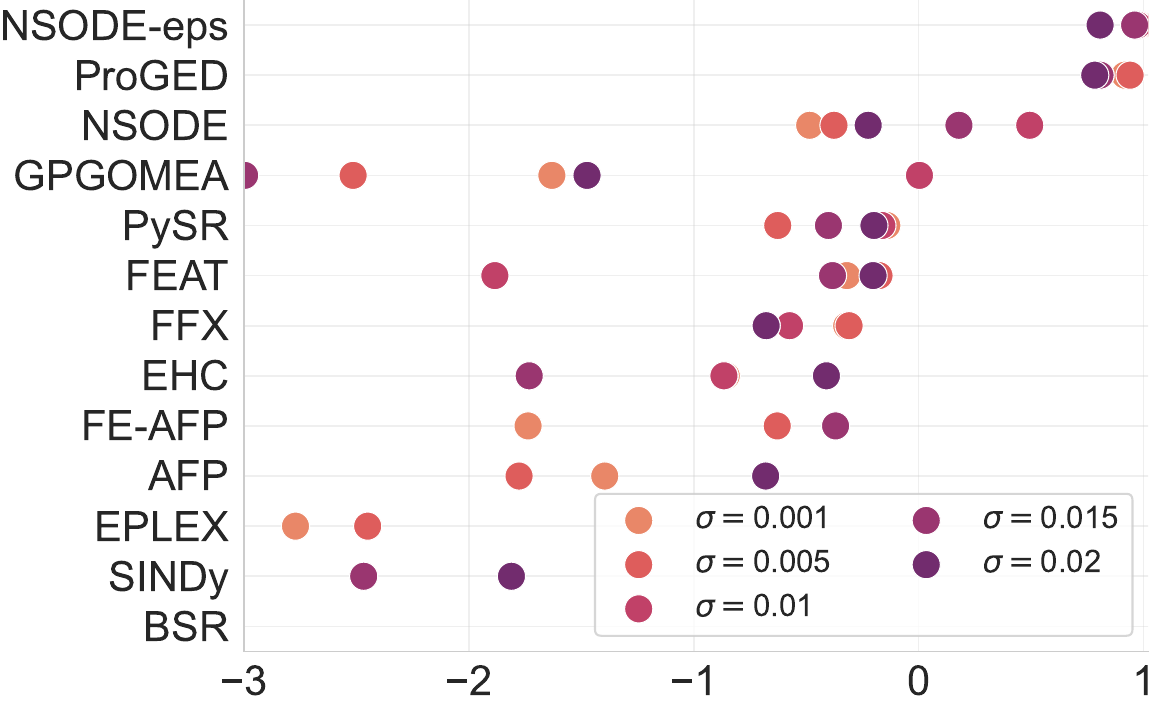}
    \caption{R$^2$, n=128}
\end{subfigure}%
\begin{subfigure}{.33\textwidth}
    \centering
    \includegraphics[width=1\linewidth]{figs/performance/classic_r2_192_extra.pdf}
    \caption{R$^2$, n=192}
\end{subfigure}%
\begin{subfigure}{.33\textwidth}
    \centering
    \includegraphics[width=1\linewidth]{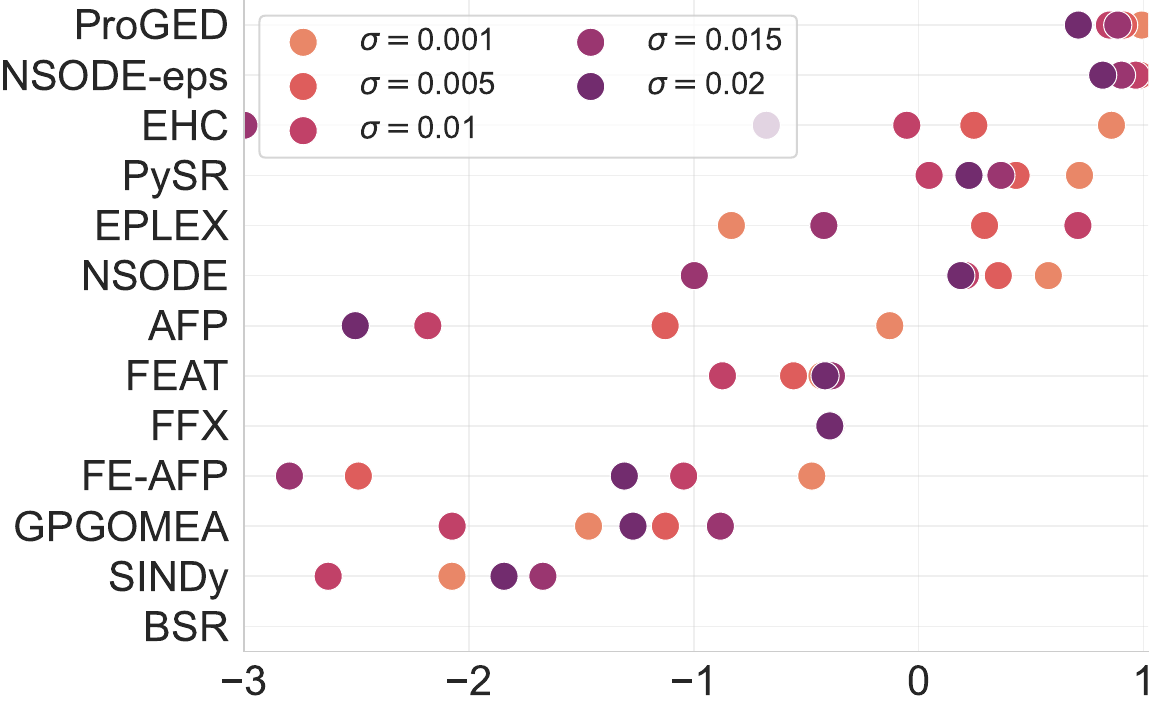}
    \caption{R$^2$, n=256}
\end{subfigure}%
\vspace{0.4cm}
\begin{subfigure}{.33\textwidth}
    \centering
    \includegraphics[width=1\linewidth]{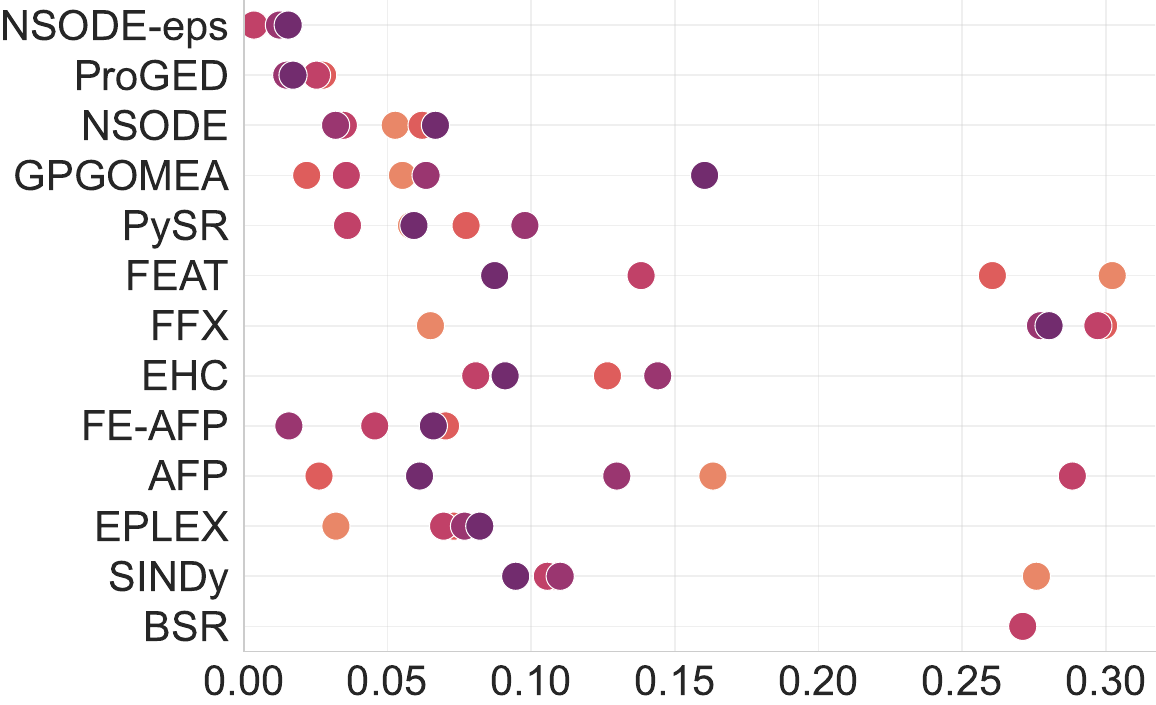}
    \caption{$L_1$, n=128}
\end{subfigure}%
\begin{subfigure}{.33\textwidth}
    \centering
    \includegraphics[width=1\linewidth]{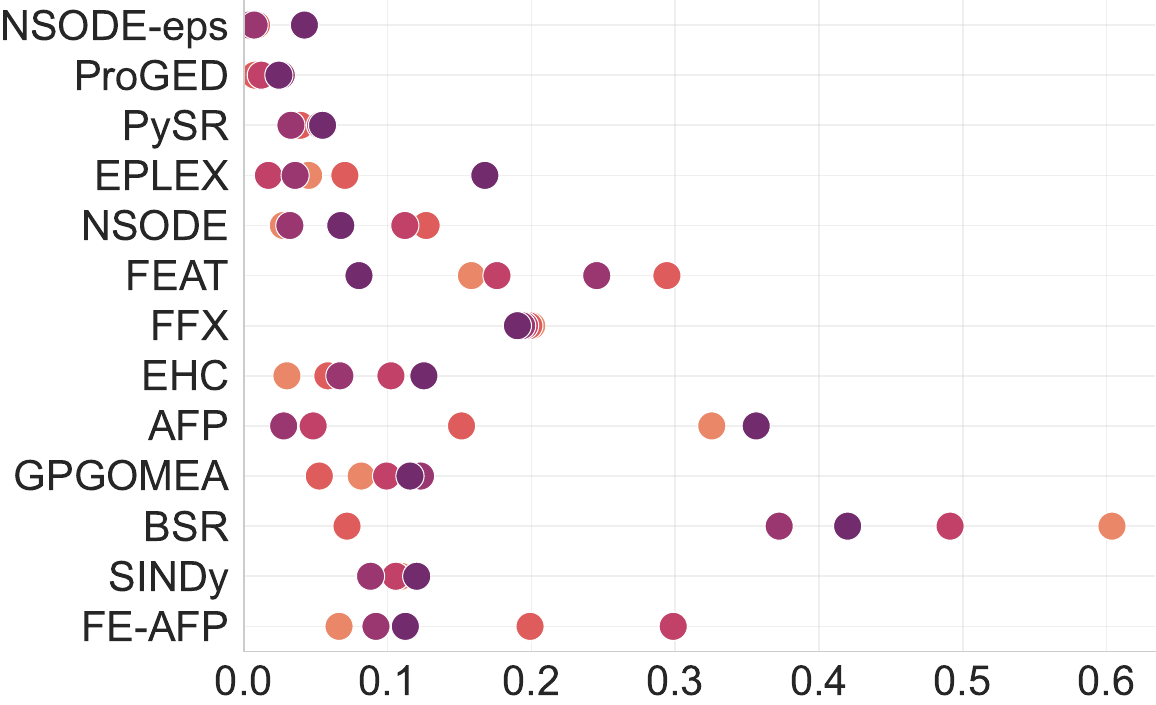}
    \caption{$L_1$, n=192}
\end{subfigure}%
\begin{subfigure}{.33\textwidth}
    \centering
    \includegraphics[width=1\linewidth]{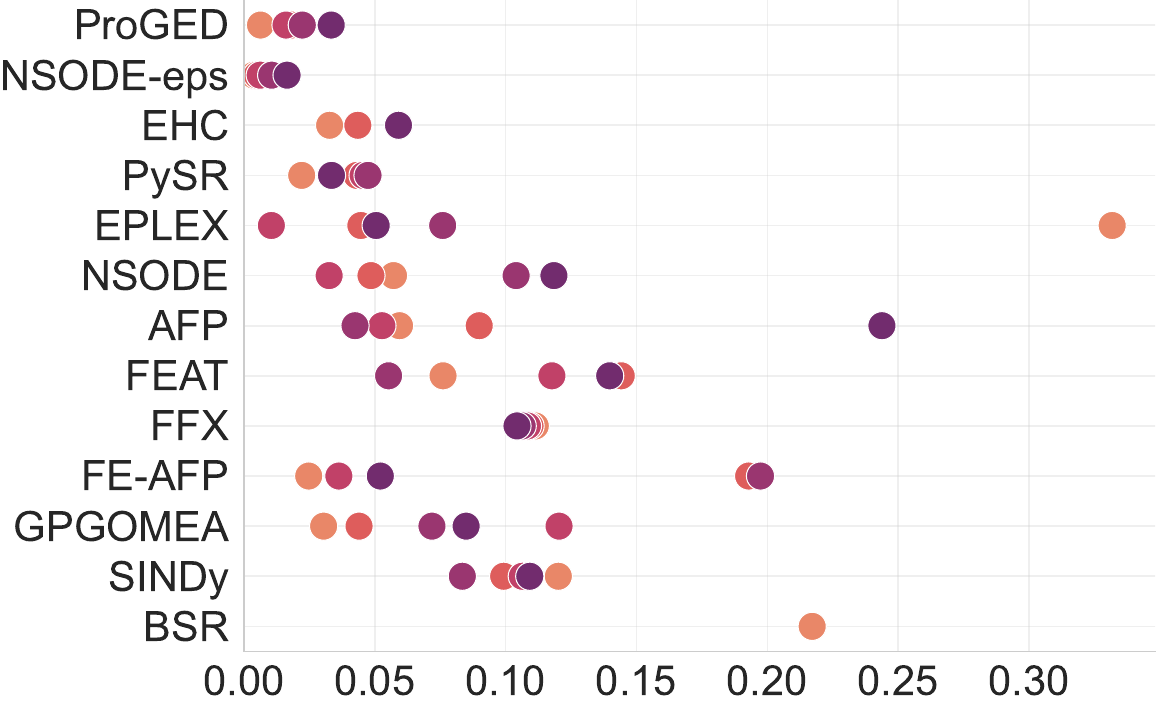}
    \caption{$L_1$, n=256}
\end{subfigure}%
\vspace{0.4cm}
\begin{subfigure}{.33\textwidth}
    \centering
    \includegraphics[width=1\linewidth]{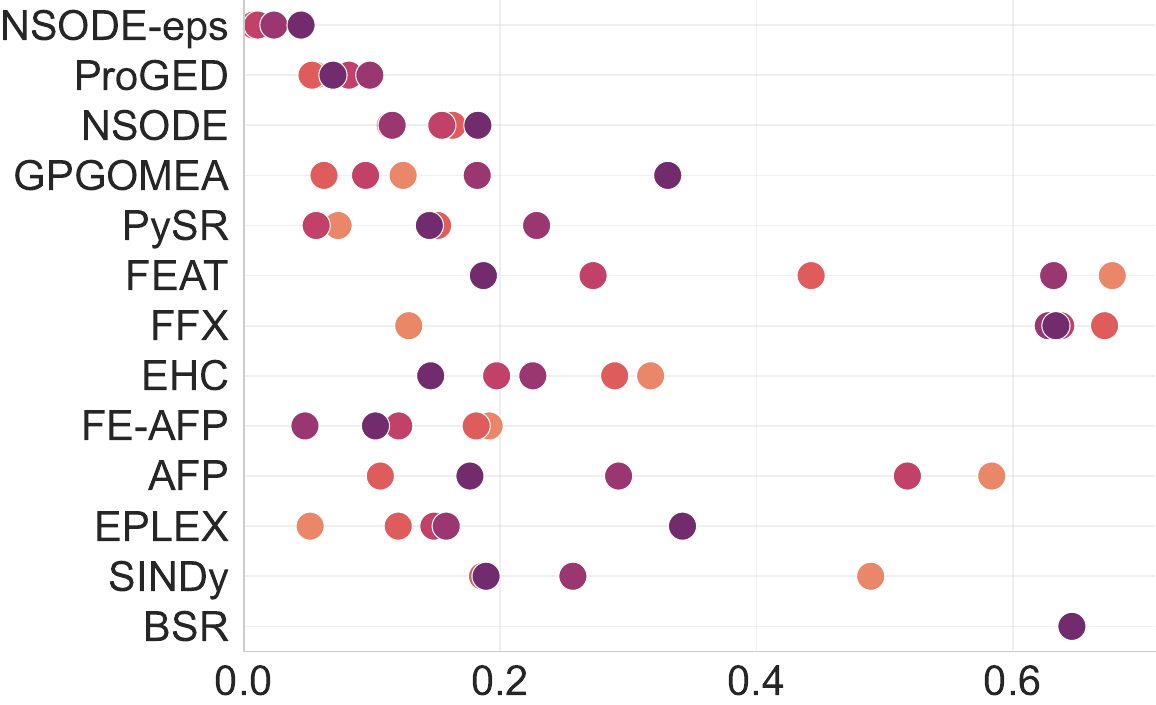}
    \caption{$L_{\infty}$, n=128}
\end{subfigure}%
\begin{subfigure}{.33\textwidth}
    \centering
    \includegraphics[width=1\linewidth]{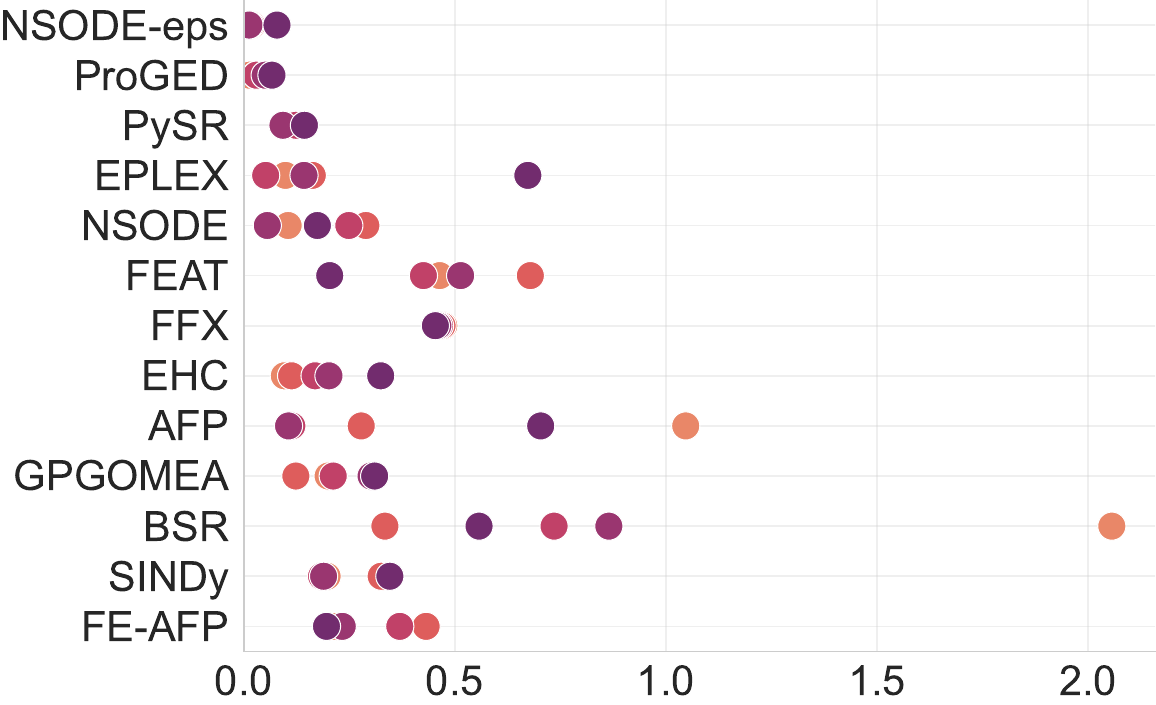}
    \caption{$L_{\infty}$, n=192}
\end{subfigure}%
\begin{subfigure}{.33\textwidth}
    \centering
    \includegraphics[width=1\linewidth]{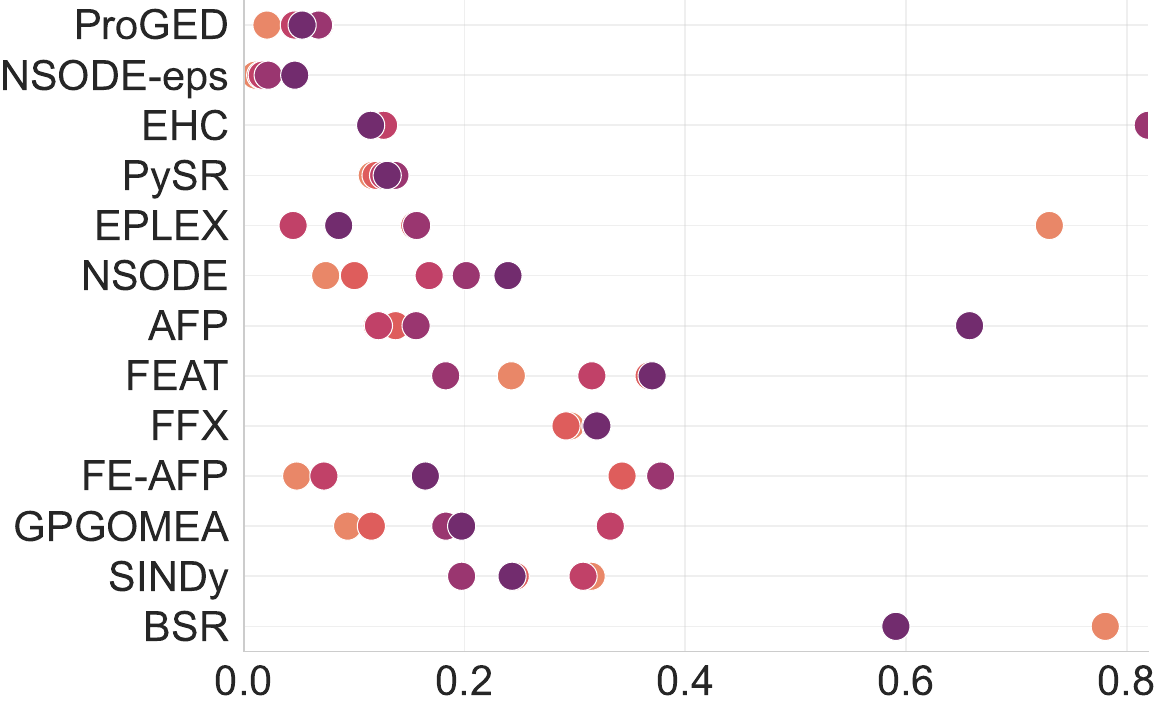}
    \caption{$L_{\infty}$, n=256}
\end{subfigure}%
\vspace{0.4cm}
\begin{subfigure}{.33\textwidth}
    \centering
    \includegraphics[width=1\linewidth]{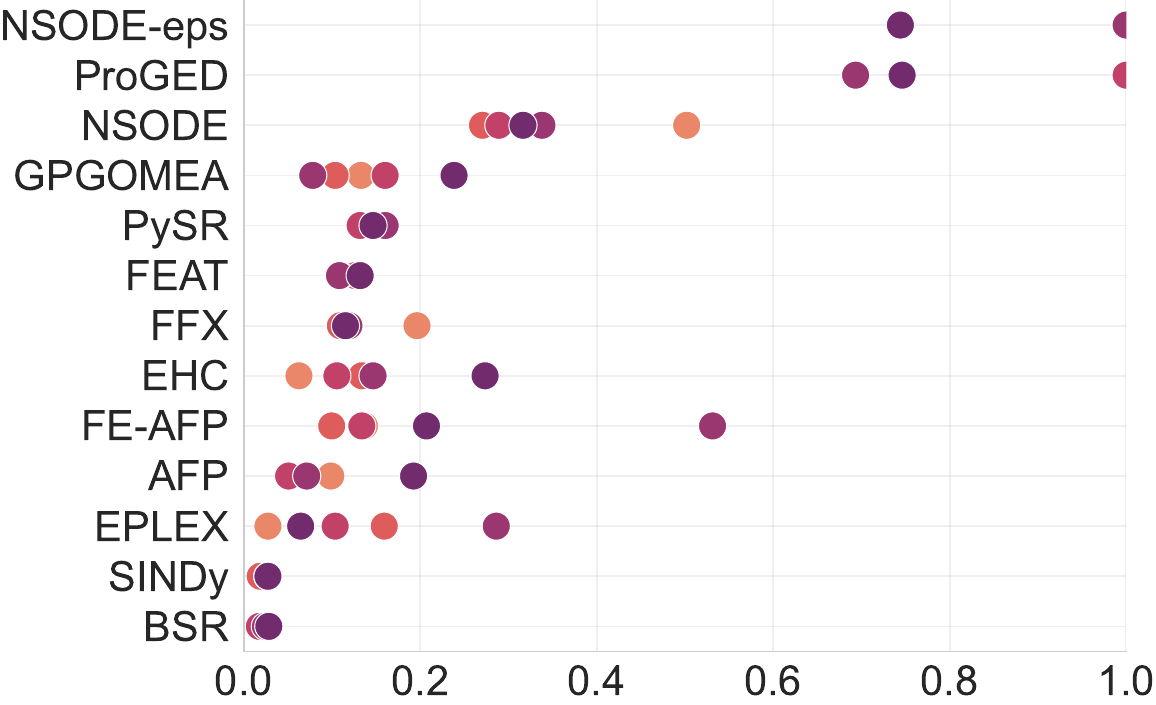}
    \caption{$\nicefrac{\texttt{isclose}}{n}$, n=128}
\end{subfigure}%
\begin{subfigure}{.33\textwidth}
    \centering
    \includegraphics[width=1\linewidth]{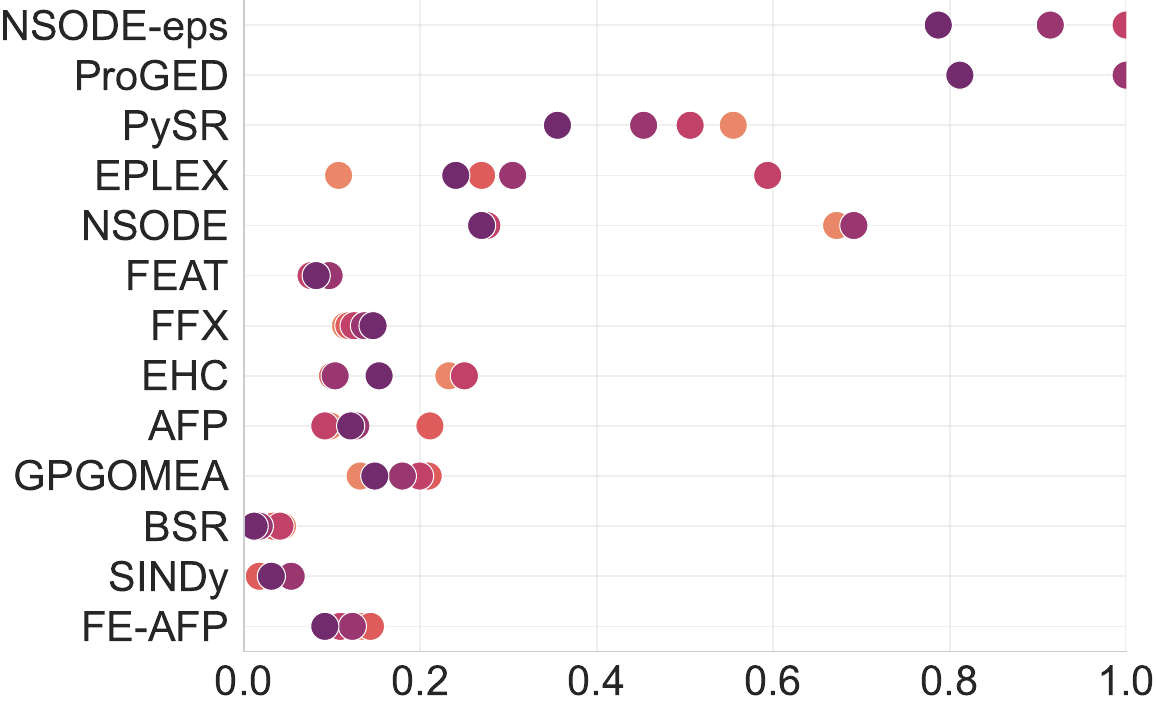}
    \caption{$\nicefrac{\texttt{isclose}}{n}$, n=192}
\end{subfigure}%
\begin{subfigure}{.33\textwidth}
    \centering
    \includegraphics[width=1\linewidth]{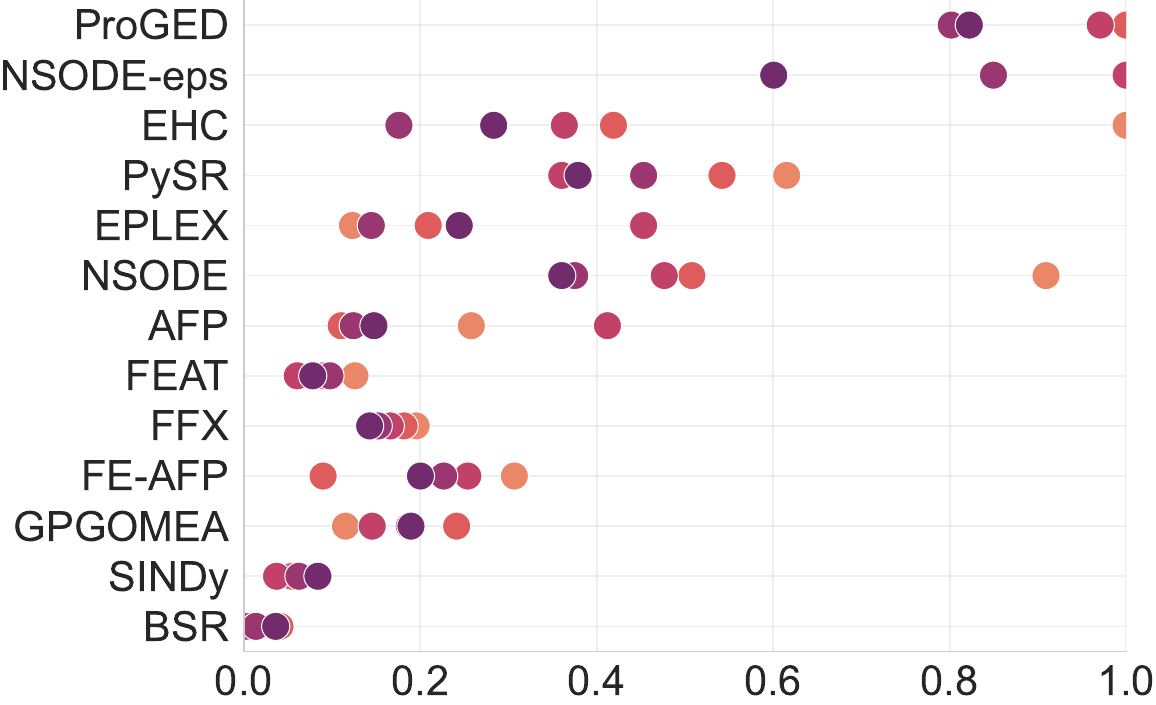}
    \caption{$\nicefrac{\texttt{isclose}}{n}$, n=256}
\end{subfigure}%
\caption{Extrapolation. Median scores on \textbf{Classic} for $n$ irregularly sampled time points across different noise levels $\sigma$.}
\label{app:results_classic_extra}
\end{figure*}

\begin{figure*}[h!]
\centering
\begin{subfigure}{.33\textwidth}
    \centering
    \includegraphics[width=1\linewidth]{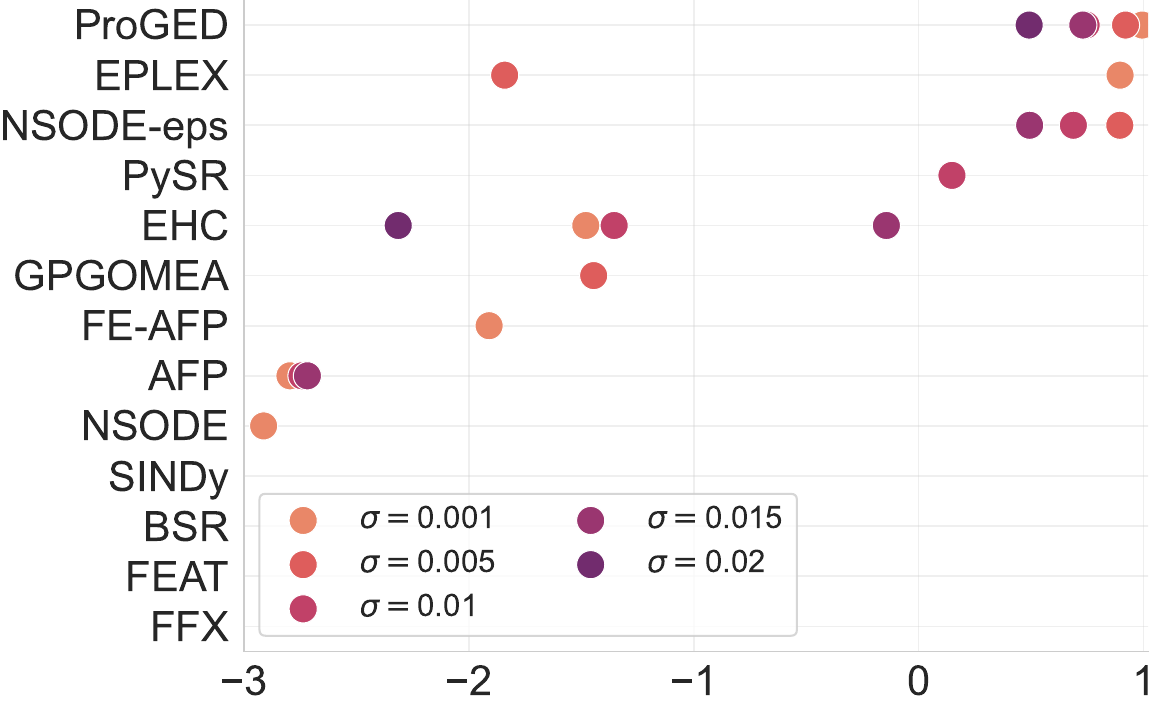}
    \caption{R$^2$, n=128}
\end{subfigure}%
\begin{subfigure}{.33\textwidth}
    \centering
    \includegraphics[width=1\linewidth]{figs/performance/knownODEs_r2_192_extra.pdf}
    \caption{R$^2$, n=192}
\end{subfigure}%
\begin{subfigure}{.33\textwidth}
    \centering
    \includegraphics[width=1\linewidth]{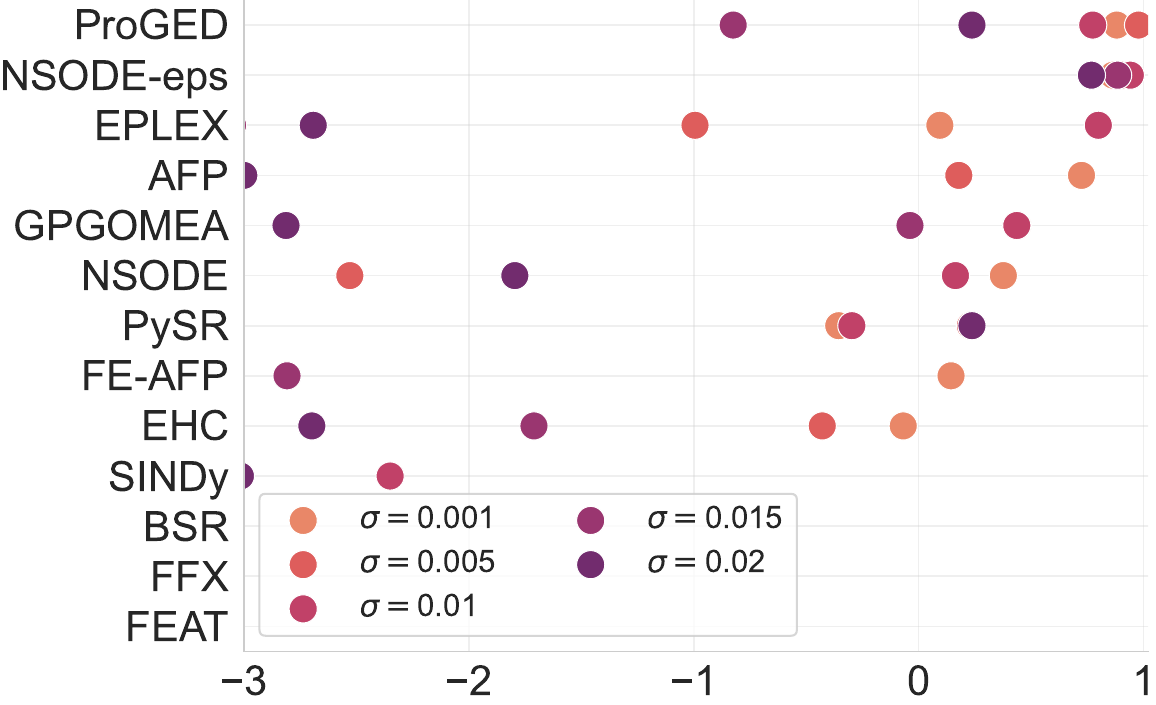}
    \caption{R$^2$, n=256}
\end{subfigure}
\vspace{0.4cm}
\begin{subfigure}{.33\textwidth}
    \centering
    \includegraphics[width=1\linewidth]{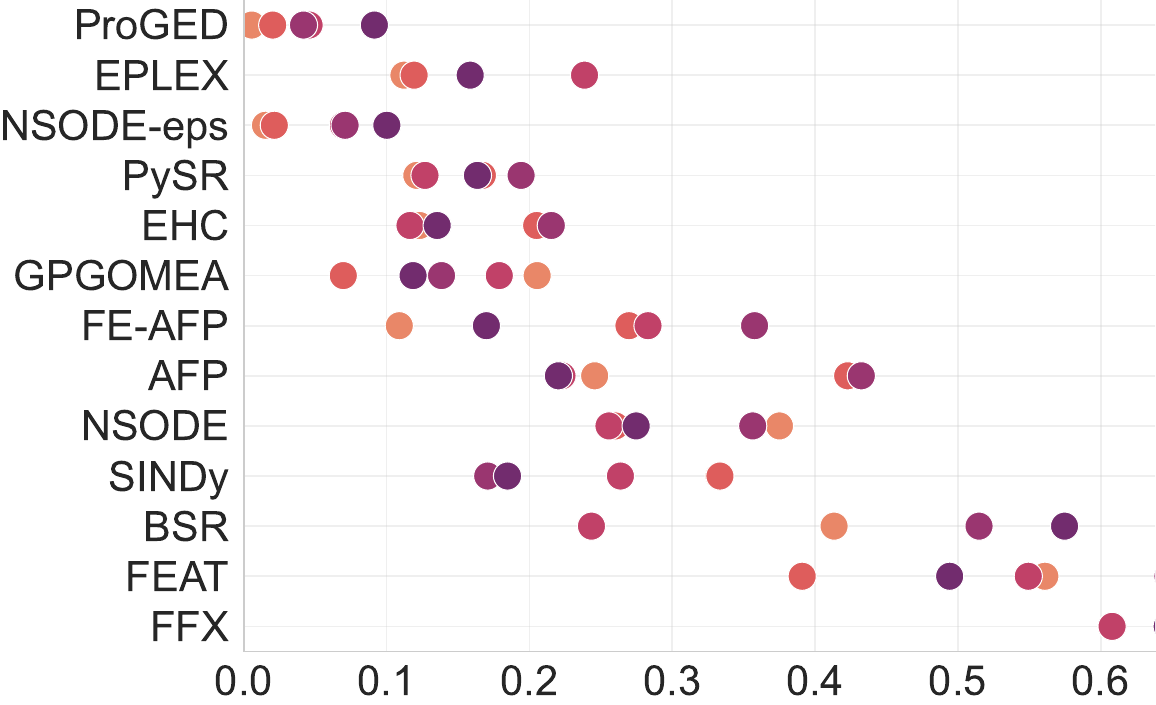}
    \caption{$L_1$, n=128}
\end{subfigure}%
\begin{subfigure}{.33\textwidth}
    \centering
    \includegraphics[width=1\linewidth]{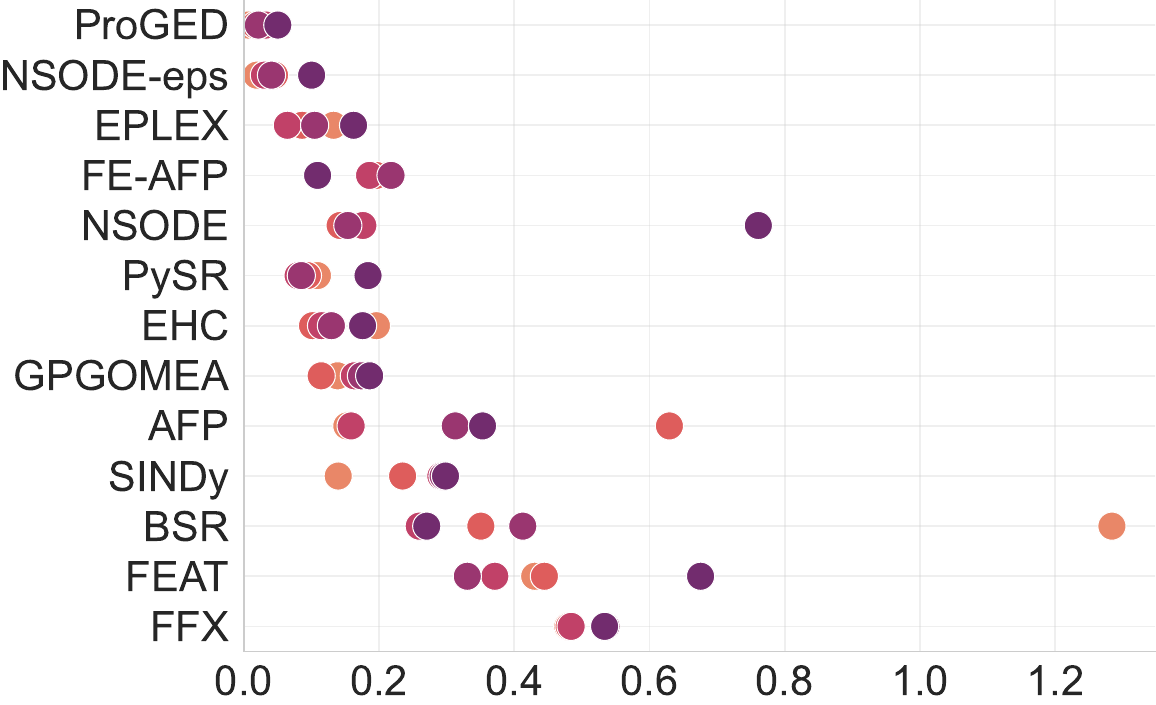}
    \caption{$L_1$, n=192}
\end{subfigure}%
\begin{subfigure}{.33\textwidth}
    \centering
    \includegraphics[width=1\linewidth]{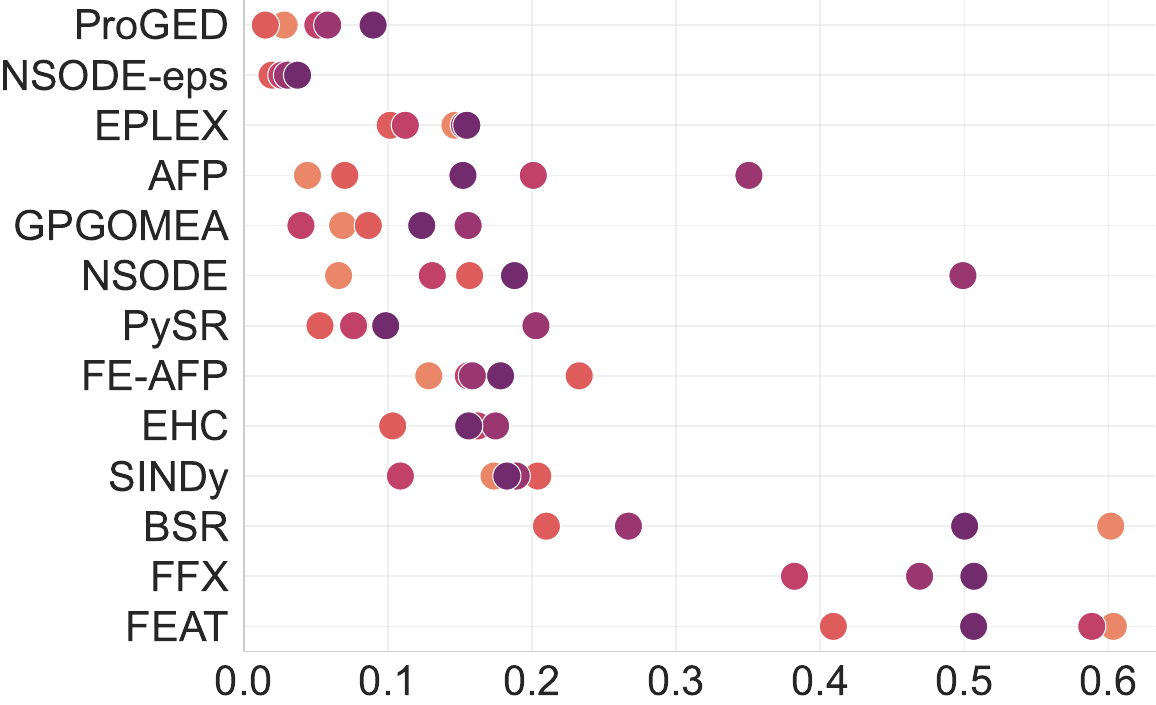}
    \caption{$L_1$, n=256}
\end{subfigure}
\vspace{0.4cm}
\begin{subfigure}{.33\textwidth}
    \centering
    \includegraphics[width=1\linewidth]{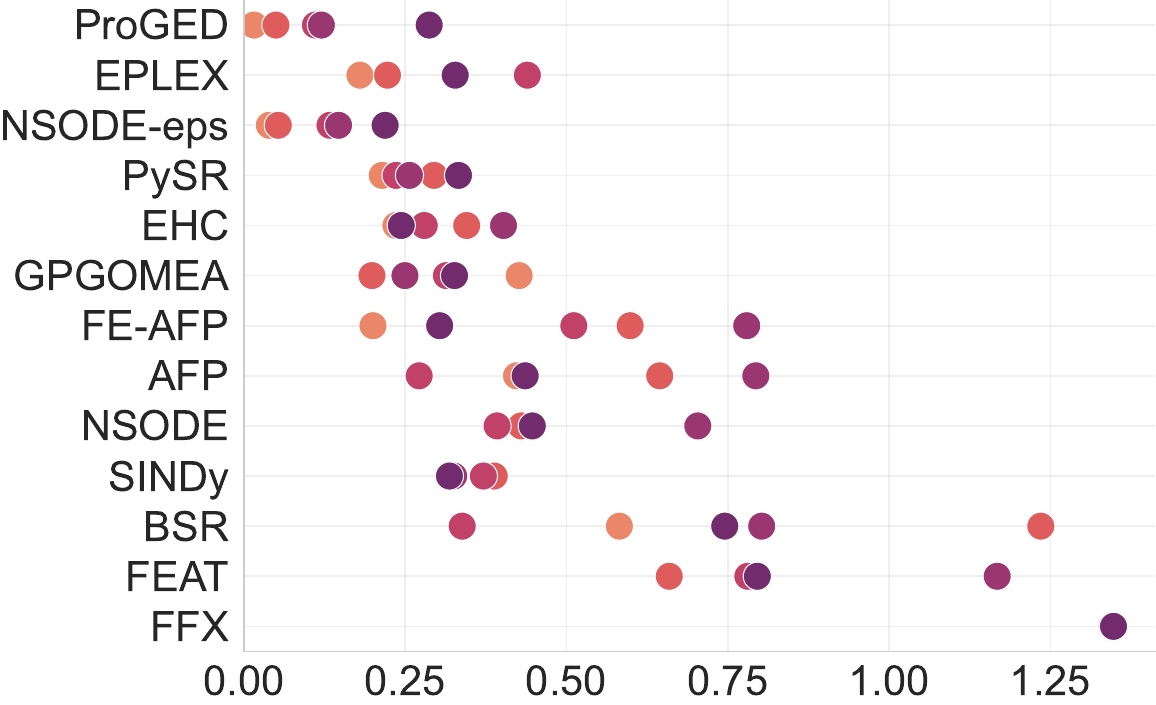}
    \caption{$L_{\infty}$, n=128}
\end{subfigure}%
\begin{subfigure}{.33\textwidth}
    \centering
    \includegraphics[width=1\linewidth]{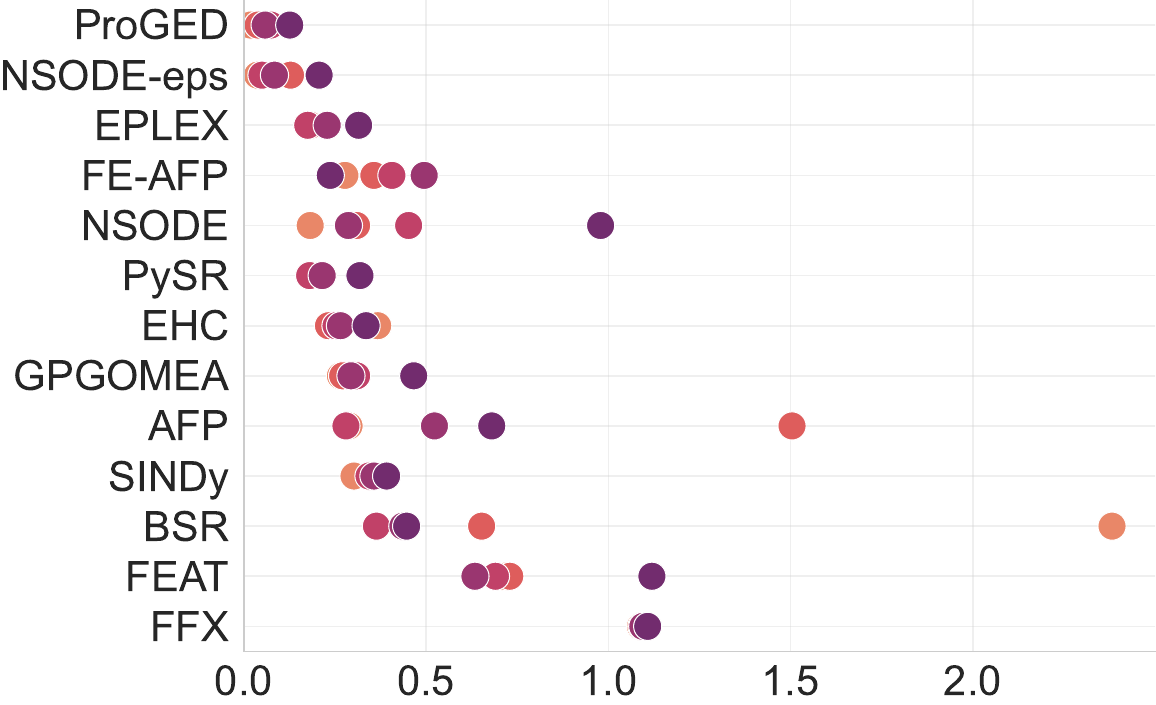}
    \caption{$L_{\infty}$, n=192}
\end{subfigure}%
\begin{subfigure}{.33\textwidth}
    \centering
    \includegraphics[width=1\linewidth]{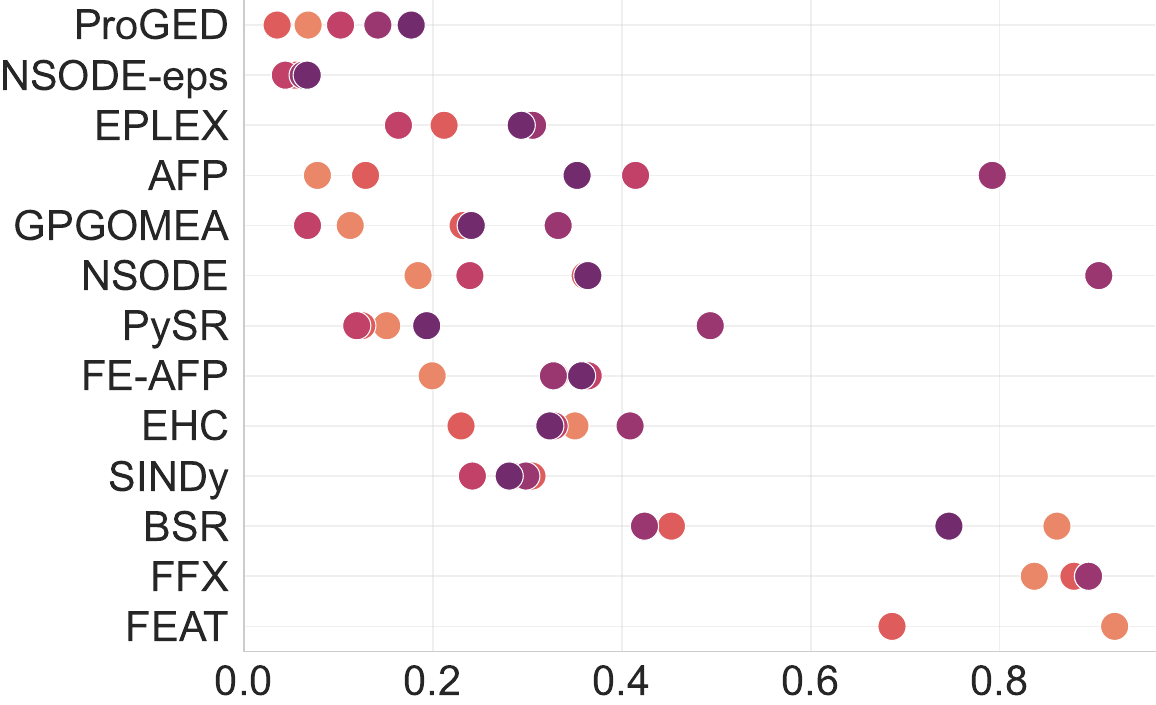}
    \caption{$L_{\infty}$, n=256}
\end{subfigure}
\vspace{0.4cm}
\begin{subfigure}{.33\textwidth}
    \centering
    \includegraphics[width=1\linewidth]{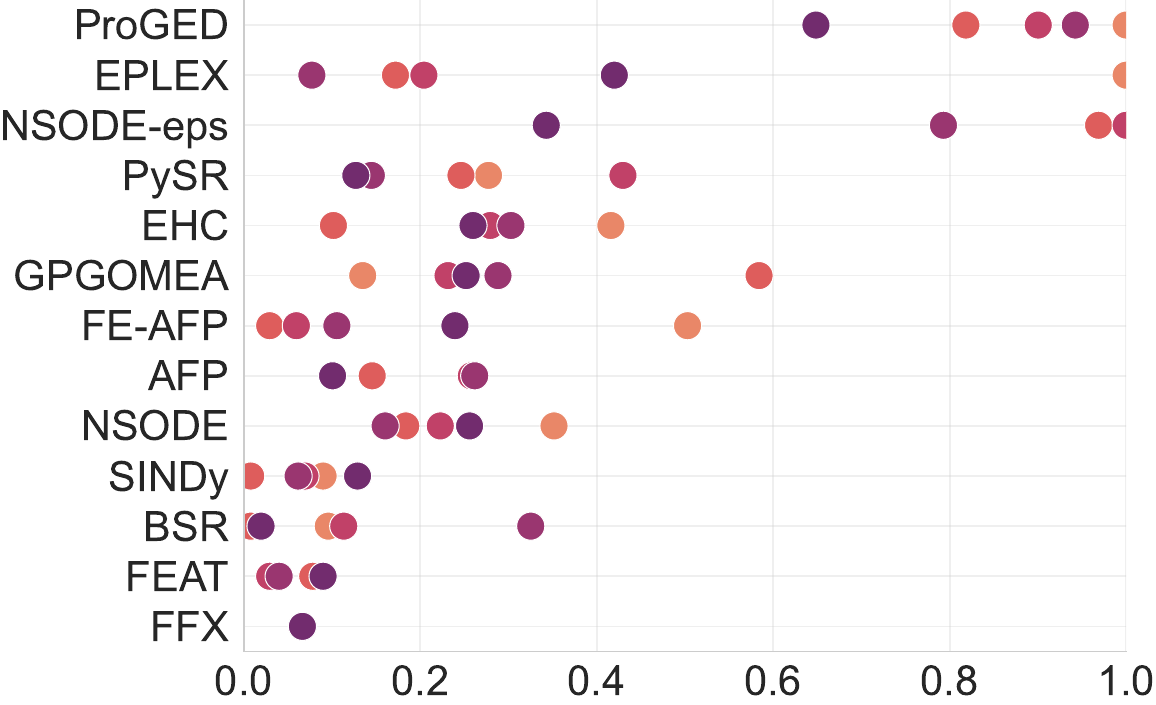}
    \caption{$\nicefrac{\texttt{isclose}}{n}$, n=128}
\end{subfigure}%
\begin{subfigure}{.33\textwidth}
    \centering
    \includegraphics[width=1\linewidth]{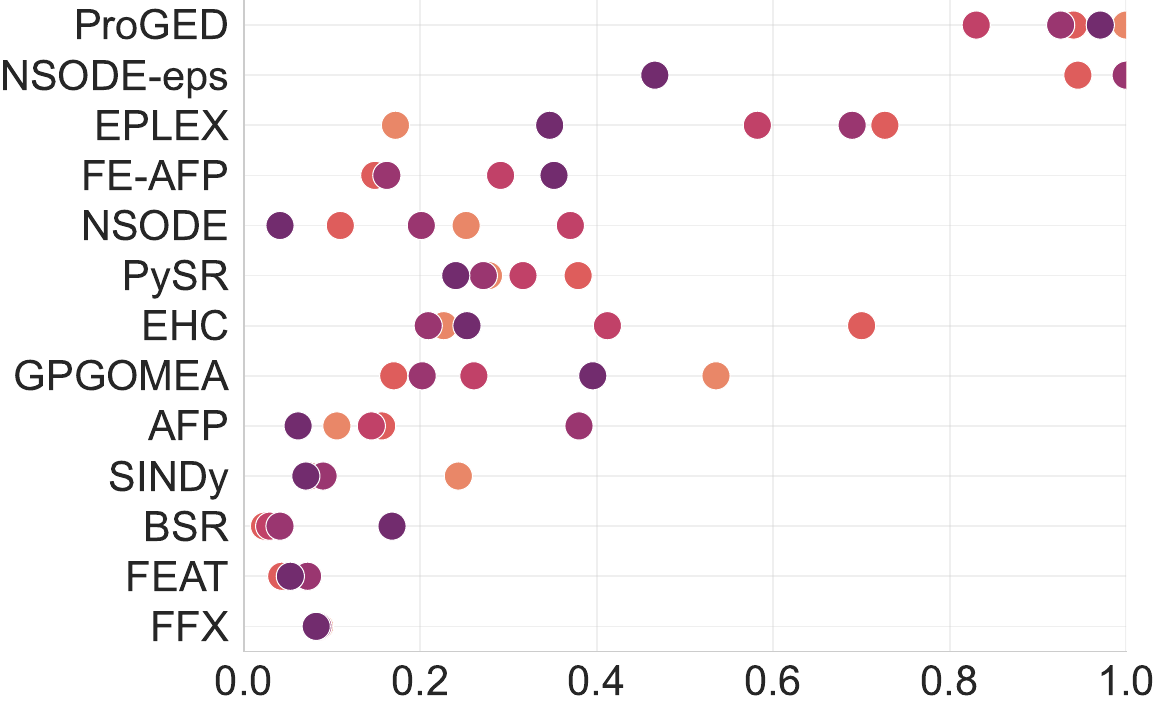}
    \caption{$\nicefrac{\texttt{isclose}}{n}$, n=192}
\end{subfigure}%
\begin{subfigure}{.33\textwidth}
    \centering
    \includegraphics[width=1\linewidth]{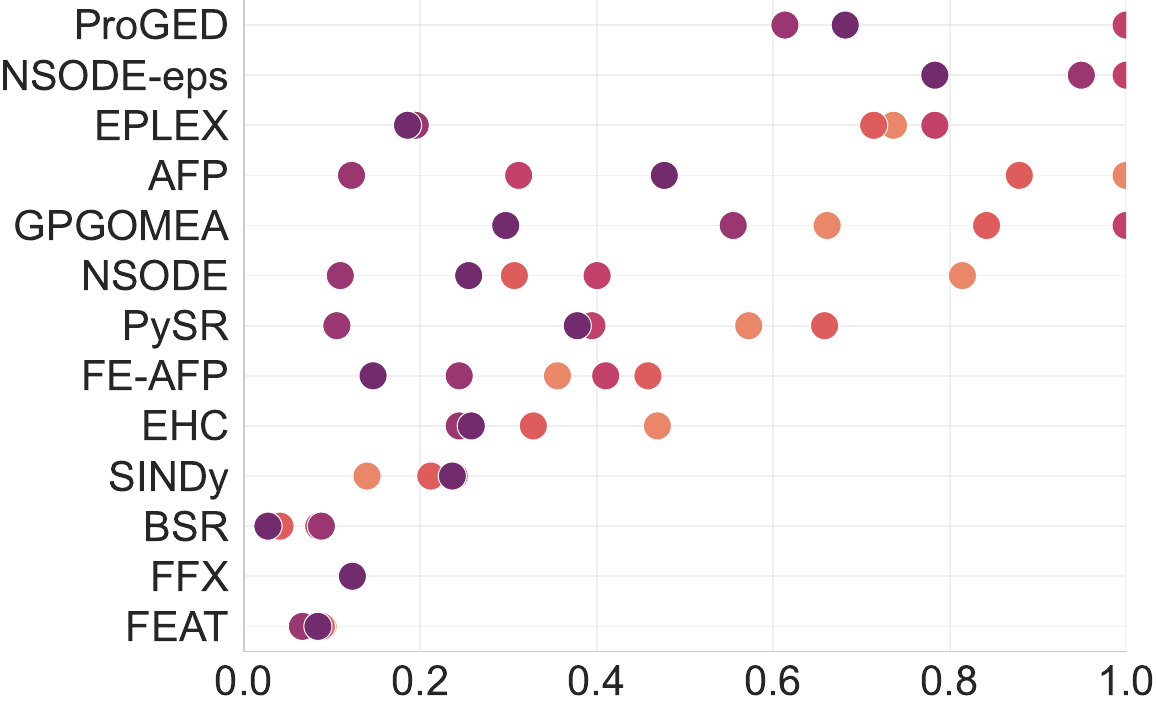}
    \caption{$\nicefrac{\texttt{isclose}}{n}$, n=256}
\end{subfigure}%
\caption{Extrapolation. Median scores on \textbf{Textbook} for $n$ irregularly sampled time points across different noise levels $\sigma$.}
\label{app:results_textbook_extra}
\end{figure*}

\begin{figure*}[h!]
\centering
\begin{subfigure}{.33\textwidth}
    \centering
    \includegraphics[width=1\linewidth]{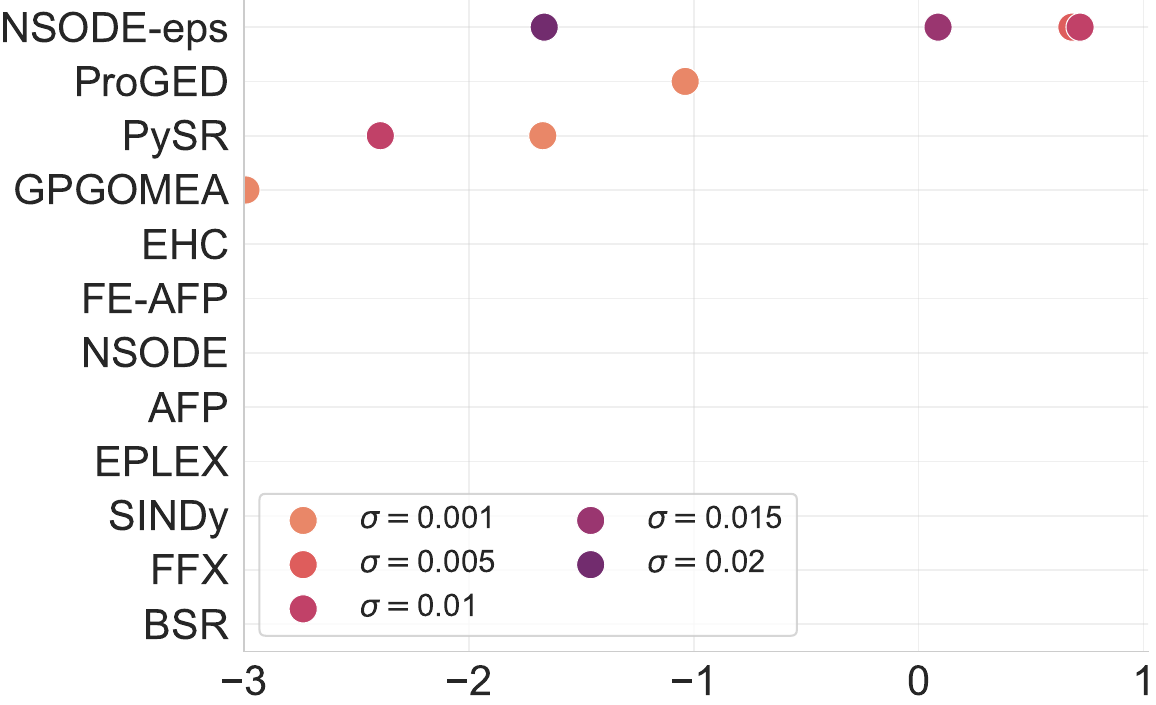}
    \caption{R$^2$, n=128}
\end{subfigure}%
\begin{subfigure}{.33\textwidth}
    \centering
    \includegraphics[width=1\linewidth]{figs/performance/163_r2_192_extra.pdf}
    \caption{R$^2$, n=192}
\end{subfigure}%
\begin{subfigure}{.33\textwidth}
    \centering
    \includegraphics[width=1\linewidth]{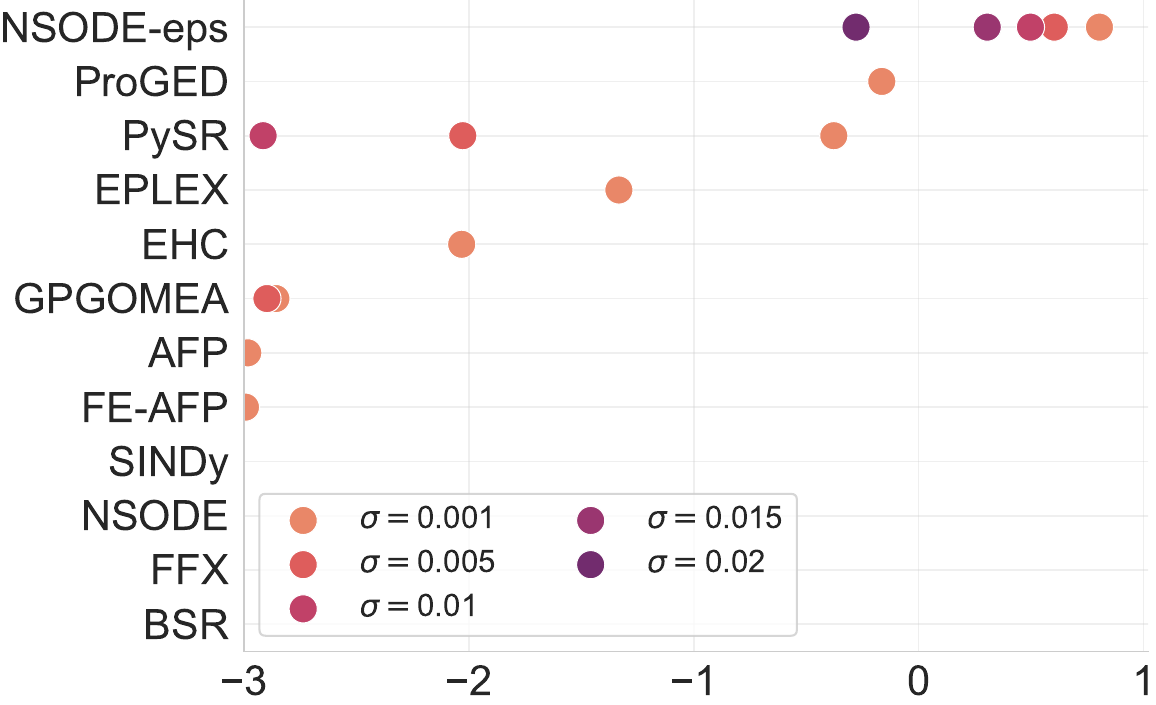}
    \caption{R$^2$, n=256}
\end{subfigure}
\vspace{0.4cm}
\begin{subfigure}{.33\textwidth}
    \centering
    \includegraphics[width=1\linewidth]{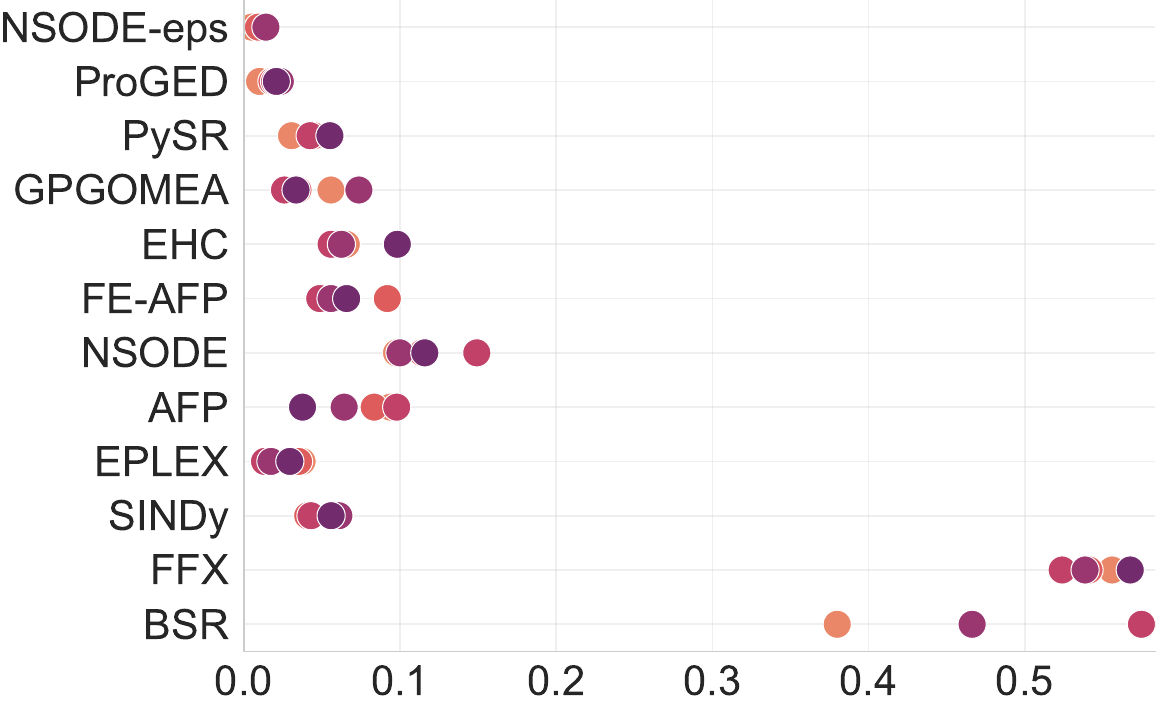}
    \caption{$L_1$, n=128}
\end{subfigure}%
\begin{subfigure}{.33\textwidth}
    \centering
    \includegraphics[width=1\linewidth]{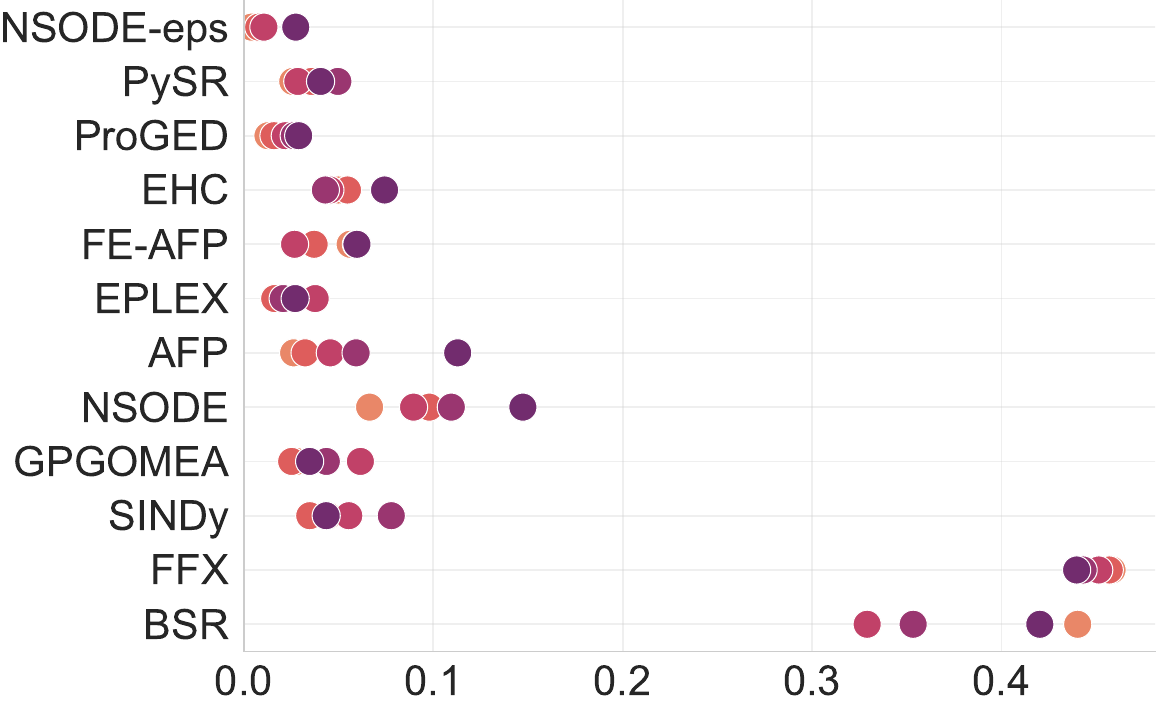}
    \caption{$L_1$, n=192}
\end{subfigure}%
\begin{subfigure}{.33\textwidth}
    \centering
    \includegraphics[width=1\linewidth]{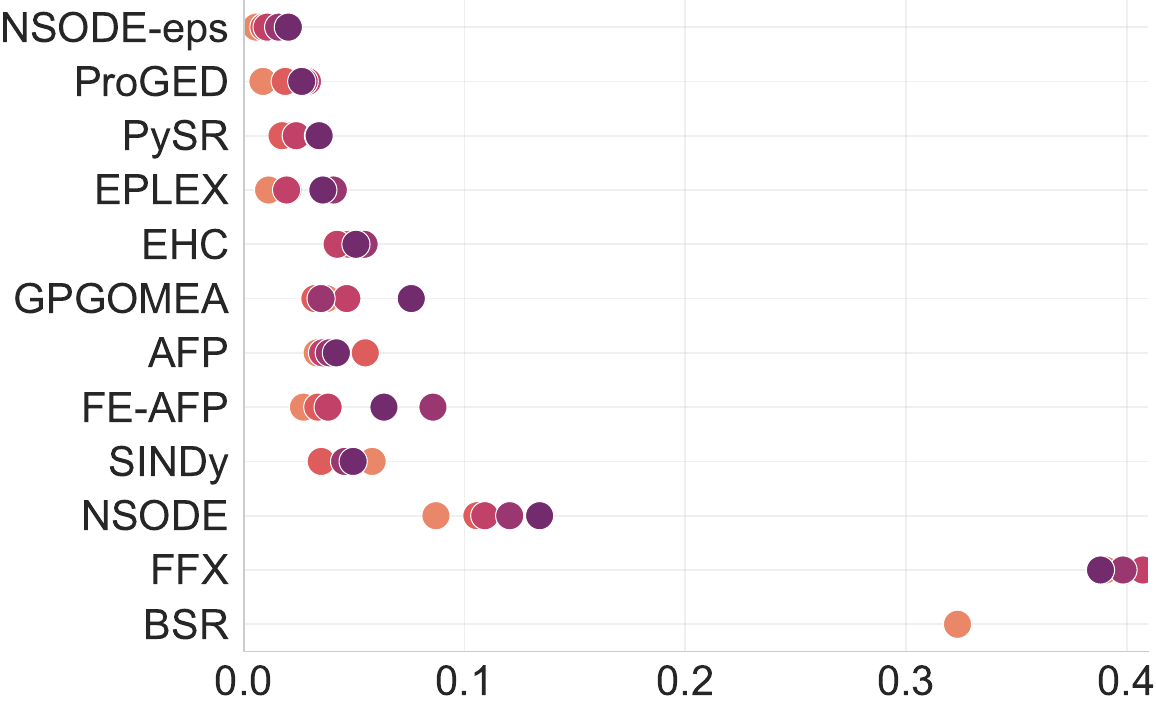}
    \caption{$L_1$, n=256}
\end{subfigure}
\vspace{0.4cm}
\begin{subfigure}{.33\textwidth}
    \centering
    \includegraphics[width=1\linewidth]{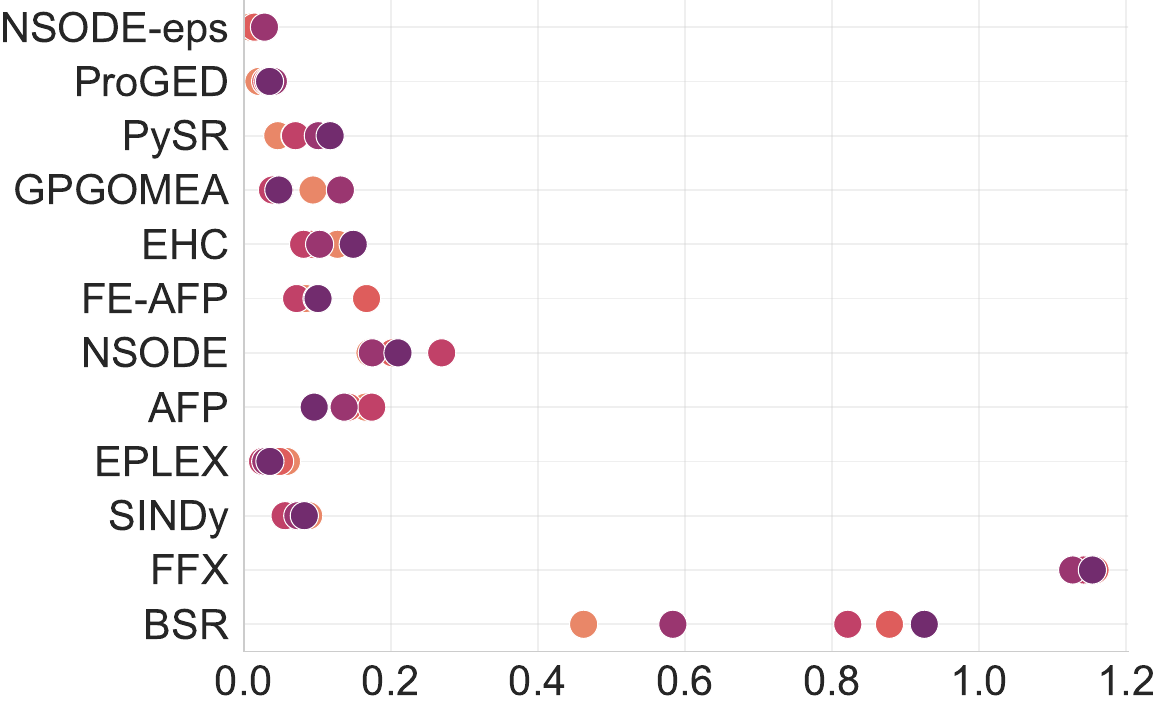}
    \caption{$L_{\infty}$, n=128}
\end{subfigure}%
\begin{subfigure}{.33\textwidth}
    \centering
    \includegraphics[width=1\linewidth]{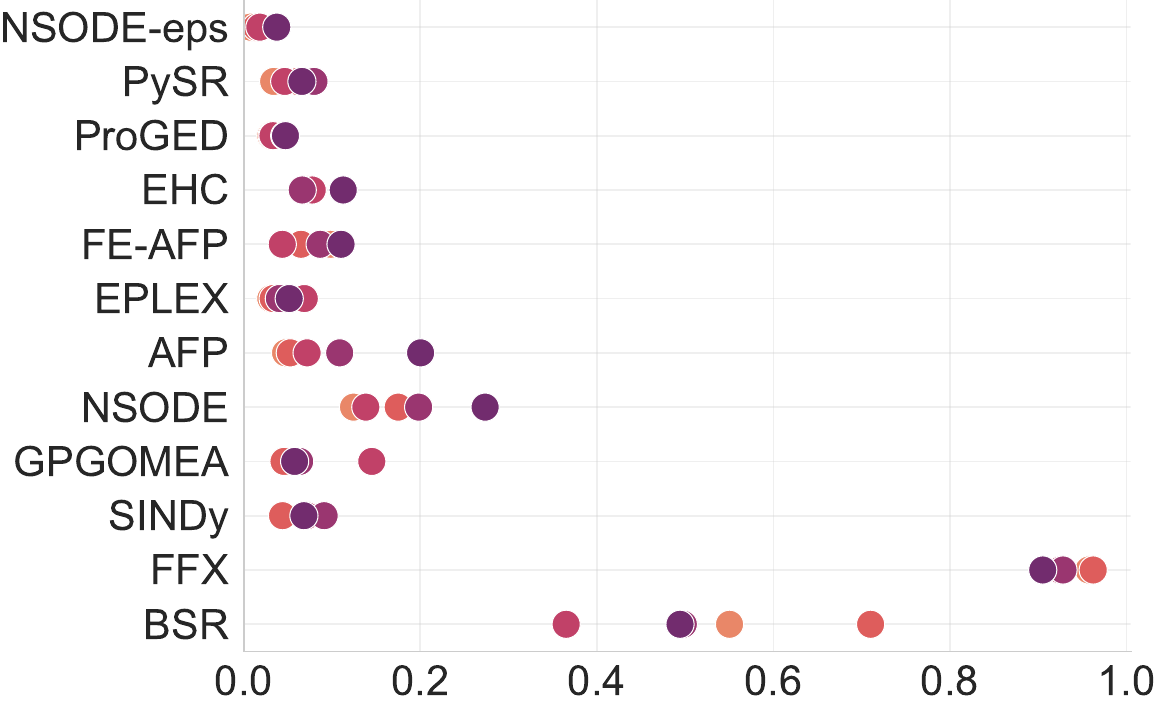}
    \caption{$L_{\infty}$, n=192}
\end{subfigure}%
\begin{subfigure}{.33\textwidth}
    \centering
    \includegraphics[width=1\linewidth]{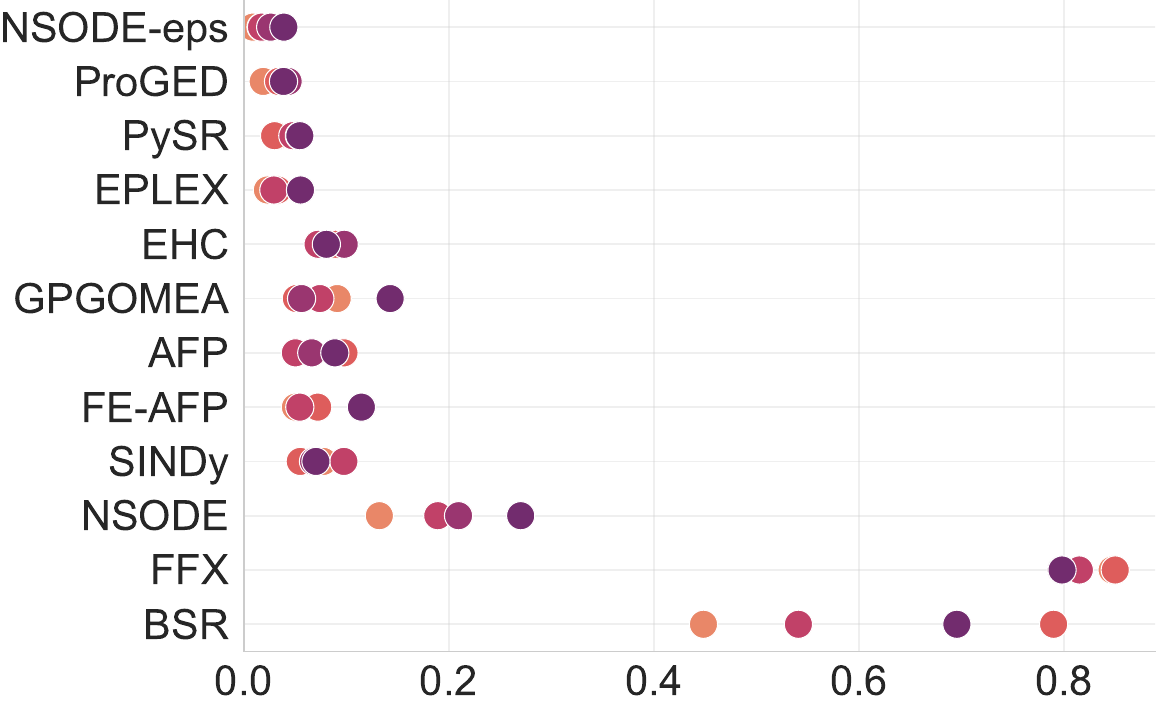}
    \caption{$L_{\infty}$, n=256}
\end{subfigure}
\vspace{0.4cm}
\begin{subfigure}{.33\textwidth}
    \centering
    \includegraphics[width=1\linewidth]{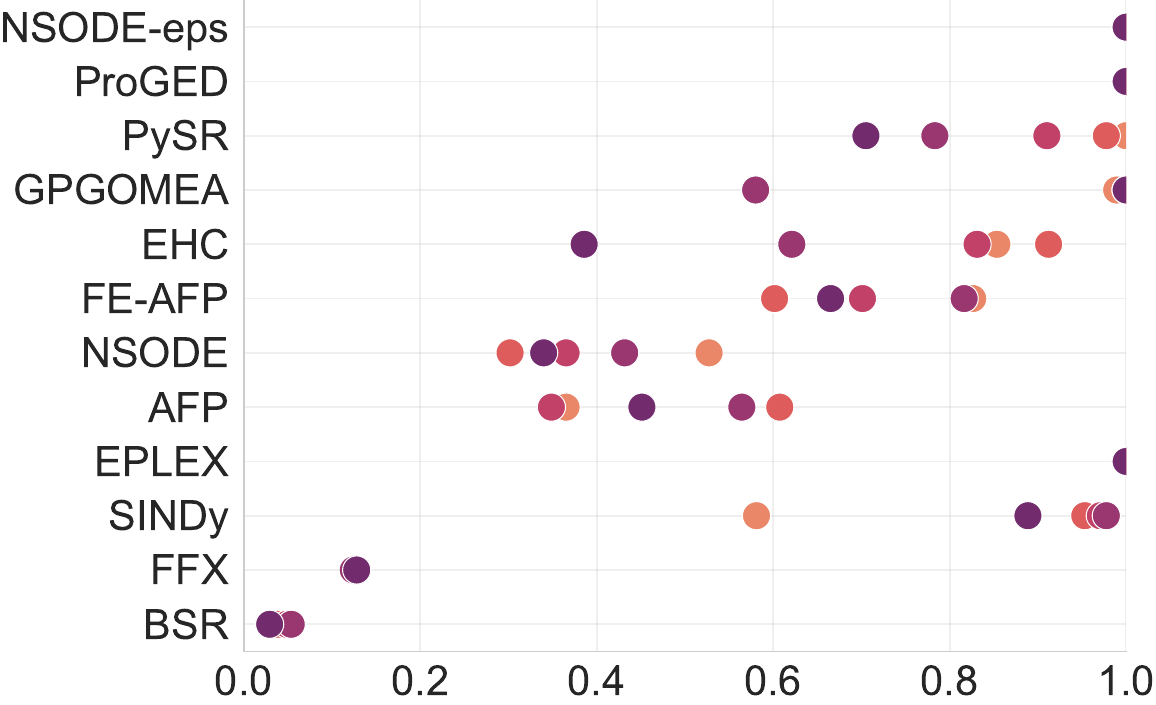}
    \caption{$\nicefrac{\texttt{isclose}}{n}$, n=128}
\end{subfigure}%
\begin{subfigure}{.33\textwidth}
    \centering
    \includegraphics[width=1\linewidth]{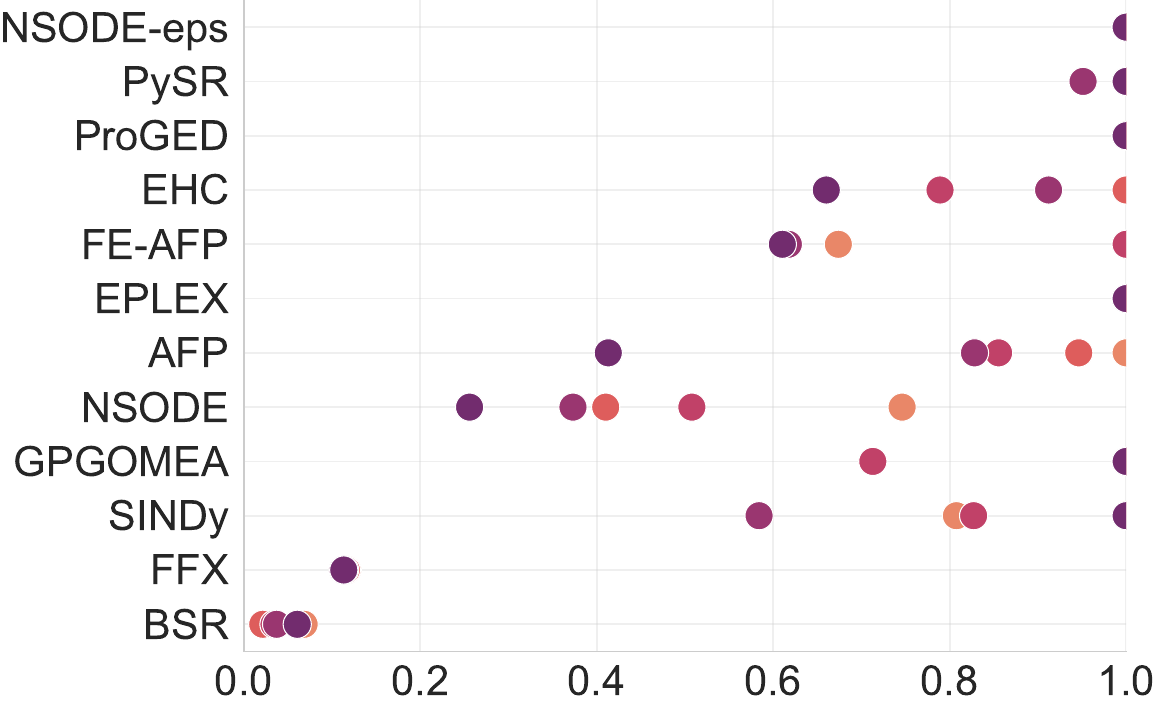}
    \caption{$\nicefrac{\texttt{isclose}}{n}$, n=192}
\end{subfigure}%
\begin{subfigure}{.33\textwidth}
    \centering
    \includegraphics[width=1\linewidth]{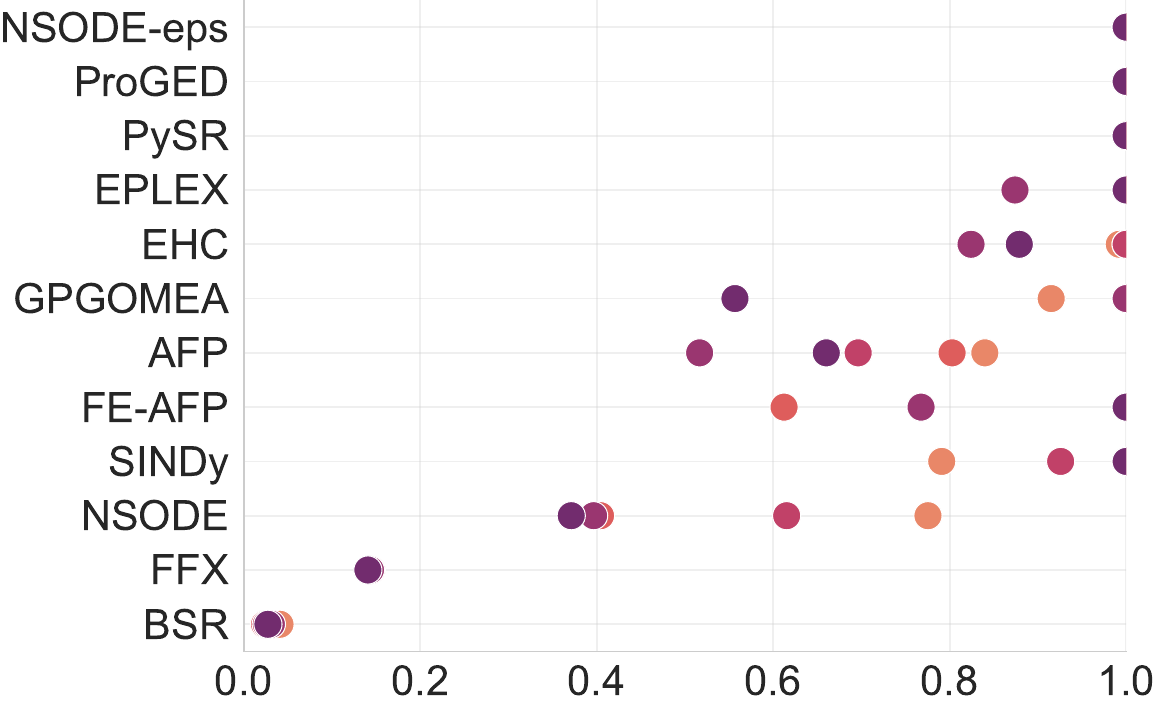}
    \caption{$\nicefrac{\texttt{isclose}}{n}$, n=256}
\end{subfigure}%
\caption{Extrapolation. Median scores on \textbf{Large} for $n$ irregularly sampled time points across different noise levels $\sigma$.}
\label{app:results_large_extra}
\end{figure*}

\clearpage
\section{Example trajectories}
\label{app:example_trajectories}

\begin{figure*}[h!]
\centering
\begin{subfigure}{.33\textwidth}
    \centering
    \includegraphics[width=1\linewidth]{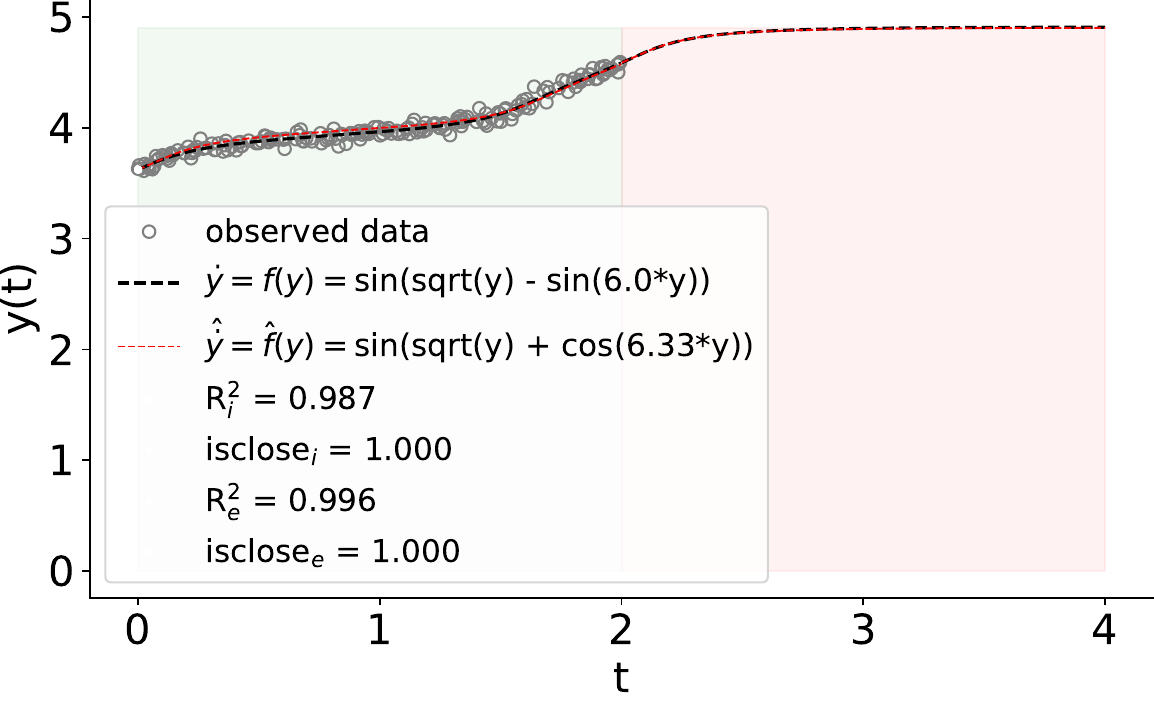}
    \caption{}\label{app:example_a}
\end{subfigure}%
\begin{subfigure}{.33\textwidth}
    \centering
    \includegraphics[width=1\linewidth]{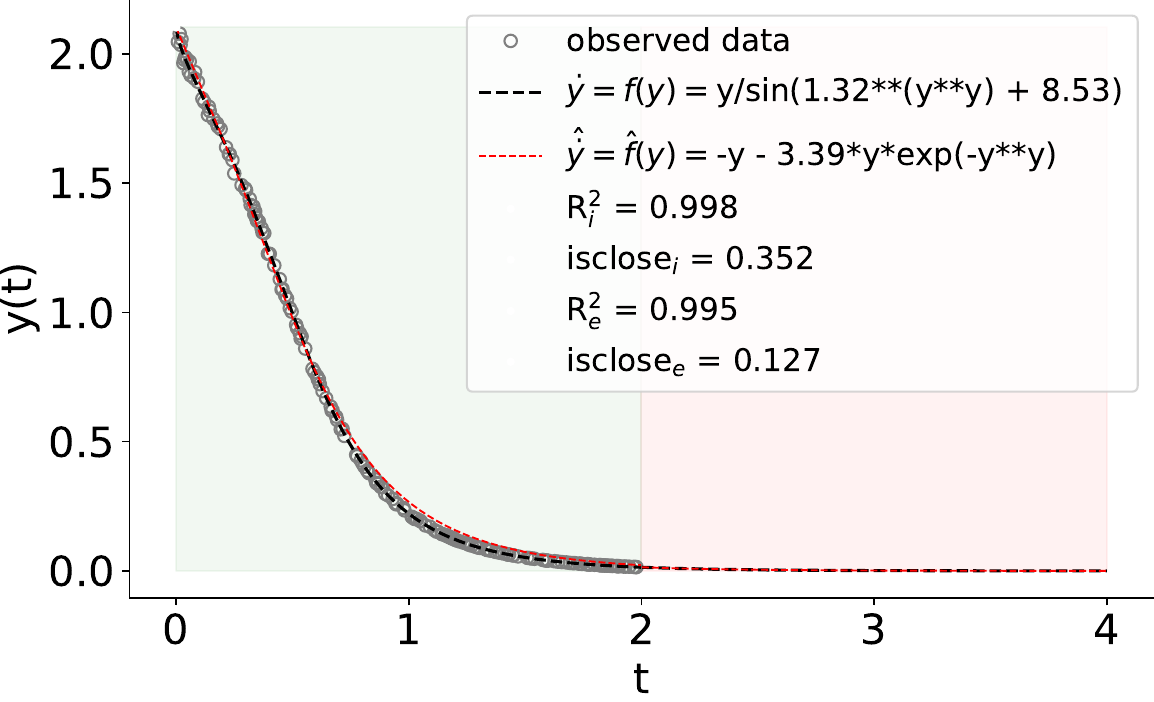}
    \caption{}\label{app:example_b}
\end{subfigure}%
\begin{subfigure}{.33\textwidth}
    \centering
    \includegraphics[width=1\linewidth]{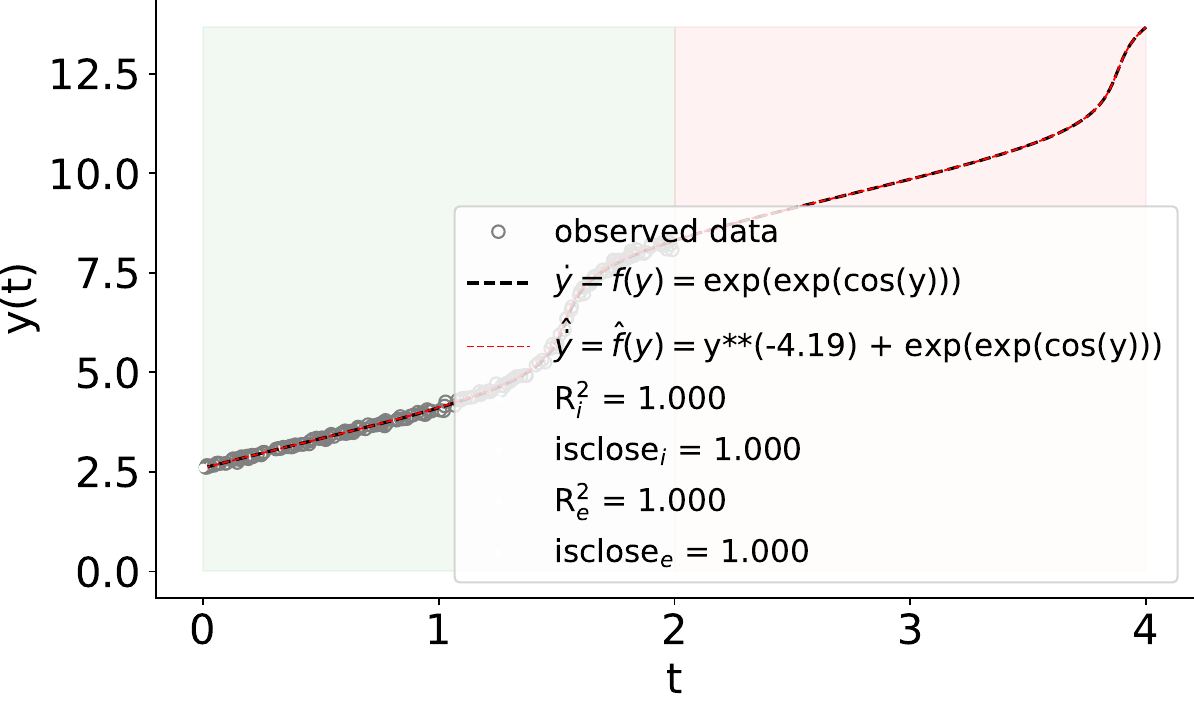}
    \caption{}\label{app:example_c}
\end{subfigure}%
\\
\begin{subfigure}{.33\textwidth}
    \centering
    \includegraphics[width=1\linewidth]{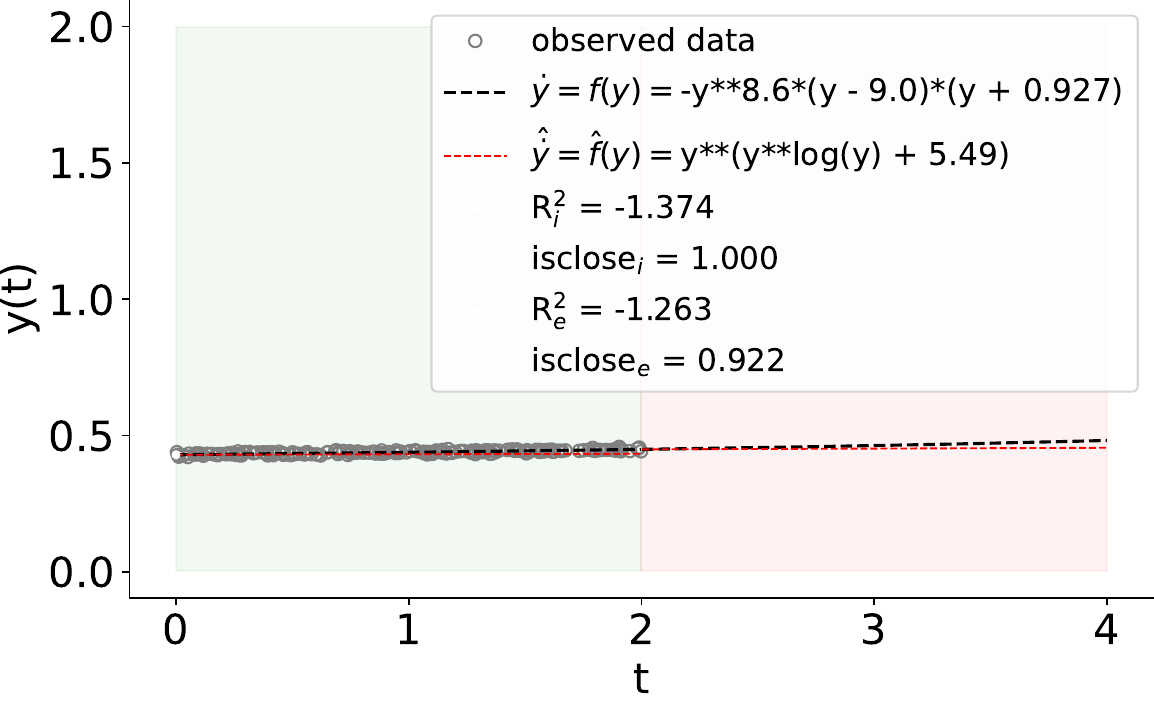}
    \caption{}\label{app:example_d}
\end{subfigure}%
\begin{subfigure}{.33\textwidth}
    \centering
    \includegraphics[width=1\linewidth]{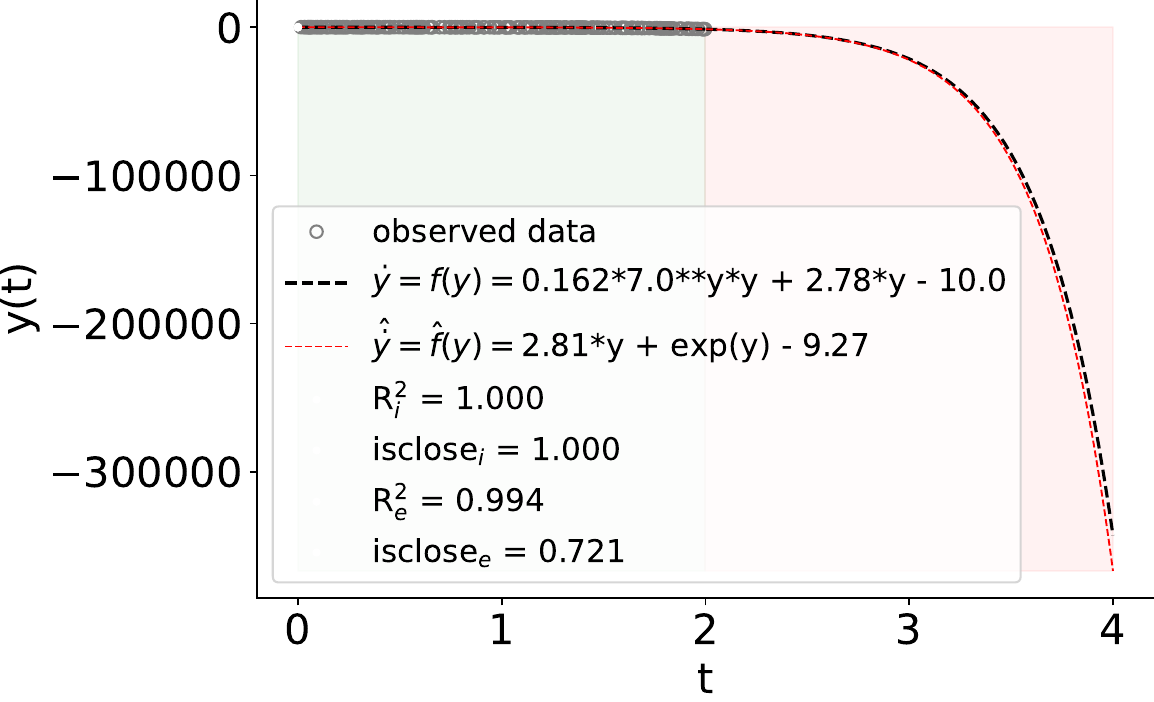}
    \caption{}\label{app:example_e}
\end{subfigure}%
\begin{subfigure}{.33\textwidth}
    \centering
    \includegraphics[width=1\linewidth]{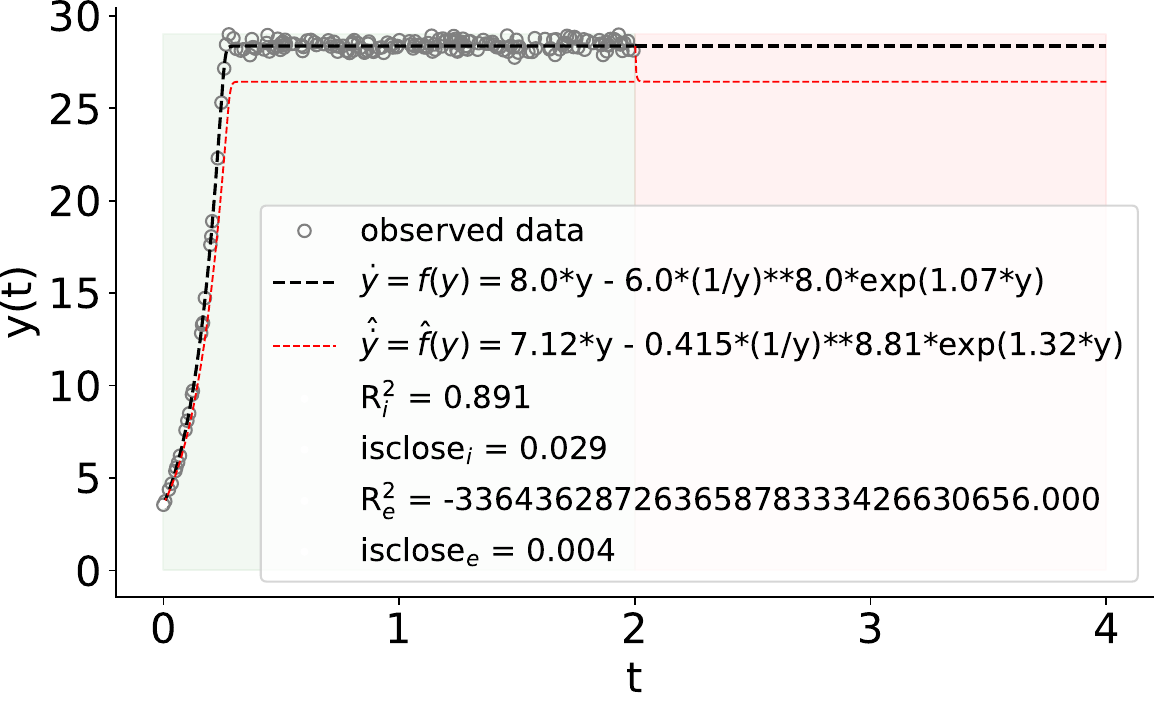}
    \caption{}\label{app:example_f}
\end{subfigure}%
\caption{Different example trajectories and trajectories predicted by NSODE-eps with noise standard deviation of $\sigma=0.01$ and $n=192$ irregularly sampled points. Green and red area correspond to interpolation and extrapolation regimes.}
\label{app:example_predictions}
\end{figure*}

Below we provide a few selected trajectories alongside predictions obtained from NSODE-eps on dataset \textbf{Large} with noise of $\sigma=0.01$ and $n=192$ sampled time points. There are a few aspects to note: firstly, despite being autonomous scalar ODEs whose limit behavior is confined to convergence to an equilibrium or divergence, we can see that this function class can still exhibit rich and highly non-linear behavior before reaching this limit. This is perhaps most pronounced in \cref{app:example_a} and \cref{app:example_c}. Secondly, even though in our evaluation the two accuracy metrics R$^2$ and average \texttt{isclose} appear to be highly correlated we can see that they do not always capture the same phenomenon, compare e.g. in \cref{app:example_b}, \cref{app:example_d} and \cref{app:example_f}.

\section{Open Source Software acknowledgement}
For this research project we heavily relied on available open source software packages which we list in \cref{tab:software_acknowledgement}.

\begin{table*}[h!]
\rowcolors{2}{white}{gray!15}
\centering
\caption{Overview of software packages we used in our work.}\label{tab:software_acknowledgement}
\begin{tabularx}{0.7\linewidth}{cY}
  \toprule \rowcolor{white}
    \textbf{Name} & \textbf{Reference}  \\ \midrule
    Python &  \citep{vanrossum2009python} \\
    PyTorch &  \citep{paszke2019pytorch} \\
    Numpy &  \citep{harris2020numpy} \\
    Pandas &  \citep{reback2020pandas,mckinney-proc-scipy-2010} \\
    Jupyter &  \citep{kluyver2016jupyter} \\
    Matplotlib &  \citep{hunter2007matplotlib} \\
    Scikit-learn &  \citep{pedregosa2011scikit} \\
    Seaborn &  \citep{waskom2021seaborn} \\
    SciPy &  \citep{scipy} \\
    SymPy &  \citep{sympy} \\
    HuggingFace &   \citep{wolf2020transformers} \\
    h5py &  \url{https://www.h5py.org/} \\
    \bottomrule
\end{tabularx}%
\end{table*}

\end{document}